\documentclass[sigconf]{acmart}
\AtBeginDocument{%
  }

\setcopyright{acmlicensed}
\copyrightyear{2018}
\acmYear{2018}
\acmDOI{XXXXXXX.XXXXXXX}
\acmConference[Conference acronym 'XX]{Make sure to enter the correct
  conference title from your rights confirmation email}{June 03--05,
  2018}{Woodstock, NY}
\acmISBN{978-1-4503-XXXX-X/2018/06}

\usepackage{enumitem}
\usepackage{booktabs}
\usepackage{multirow}
\usepackage{xcolor}
\usepackage{colortbl}
\usepackage{amsmath}
\usepackage{booktabs}
\usepackage{multirow}
\usepackage{array}
\usepackage{tabularray}
\usepackage{amsthm}
\usepackage{enumitem}
\usepackage{amsmath}
\usepackage{enumitem}
\usepackage{booktabs}   % \toprule / \midrule / \bottomrule
\usepackage{multirow}   % （备用，若后续需要跨行合并）
\usepackage{natbib}

\usepackage{multirow}        % \multirow
\usepackage{graphicx}        % \rotatebox
\usepackage[table]{xcolor}   % \cellcolor, \rowcolor
\usepackage{rotating}
\usepackage{tikz}
\usepackage{float} 
\usepackage{pifont}
\usepackage{graphicx}
\usepackage{subcaption}
\usepackage{tcolorbox}
\usepackage{xcolor}

\usepackage{tikz}

\definecolor{lightgray}{RGB}{240,240,240}
\definecolor{darkgray}{RGB}{220,220,220}
\definecolor{lightgray}{RGB}{240,240,240}
\definecolor{lightblue}{RGB}{230,240,255}
\definecolor{lightgreen}{RGB}{230,255,230}
\definecolor{skyblue}{RGB}{0, 150, 220}
\definecolor{orangeyellow}{RGB}{255, 165, 0}
\definecolor{headergray}{RGB}{180, 180, 180}
\newcommand{\gc}{\cellcolor{lightgray}}
\definecolor{case1}{RGB}{255, 165, 0}
\definecolor{case2}{RGB}{162, 1, 1}
\definecolor{case3}{RGB}{0, 32, 96}
\definecolor{case4}{RGB}{25, 88, 105}

\UseTblrLibrary{booktabs}

\renewcommand\footnotetextcopyrightpermission[1]{}
\let\titleold\title
\renewcommand{\title}[1]{\titleold{#1}\newcommand{\thetitle}{#1}}
\def\maketitlesupplementary
   {
   \newpage
       \twocolumn[
        \centering
        \huge
        \textbf{\thetitle}\\
        \vspace{0.5em}Supplementary Material \\
        \vspace{1.0em}
       ] %< twocolumn
   }
\begin{document}

\title{When Text Hijacks Vision: Benchmarking and Mitigating\\Text Overlay-Induced Hallucination in Vision Language Models}

% \affiliation{%

\author{Cui Yakun}
\affiliation{%
  \institution{The Hong Kong University of Science and Technology}
  \city{Hong Kong}
  \country{China}}
\email{ycuibd@connect.ust.hk}

\author{Xingqun Qi}
\affiliation{%
  \institution{The Hong Kong University of Science and Technology}
  \city{Hong Kong}
  \country{China}}
\email{xingqunqi@gmail.com}

\author{TianTian Geng}
\affiliation{%
  \institution{University of Birmingham}
  \city{Birmingham}
  \country{UK}}
\email{gengtiantian97@gmail.com}

\author{Yuyao Zhang}
\affiliation{%
  \institution{The Hong Kong University of Science and Technology}
  \city{Hong Kong}
  \country{China}}
\email{yzhang075@connect.ust.hk}

\author{Sirui Han}
\affiliation{%
  \institution{The Hong Kong University of Science and Technology}
  \city{Hong Kong}
  \country{China}}
\email{siruihan@ust.hk}

\author{Yike Guo}
\affiliation{%
  \institution{The Hong Kong University of Science and Technology}
  \city{Hong Kong}
  \country{China}}
\email{yikeguo@ust.hk}

\renewcommand{\shortauthors}{Trovato et al.}

\begin{abstract}
Recent advances in Vision-Language Models (VLMs) have substantially enhanced their ability across multimodal video understanding benchmarks spanning temporal, action, object, and spatial understanding.
However, we identify a critical yet overlooked issue: when embedded on-screen text contradicts the visual scene, existing VLMs systematically hallucinate, prioritizing overlay textual semantics over the actual visual content.
We formally define this phenomenon as Text Overlay-Induced Hallucination (TOIH). 
In this work, we propose VisualTextTrap, the first comprehensive benchmark, including large-scale human-validated samples with specifically designed evaluation metrics.
In particular, we construct VisualTextTrap from widely-used public datasets using a scalable hybrid pipeline of VLMs assisted text generation and rigorous manual verification. 
The benchmark features 6,057 samples annotated across 88 fine-grained attributes within four dimensions, with hallucination intensity quantified on a five-level scale (L1–L5) that reflects the semantic contradiction between overlay text and visual reality. 
We further devise 7 quantitative metrics anchored by the Hallucination Resistance Rate (HRR) to comprehensively profile models' susceptibility.
Moreover, we propose Visual Text Hallucination Mitigation Mixture-of-Experts (VTHM-MoE), a novel Vision-Text Disentanglement framework that employs a dual-encoder architecture to independently capture representations of overlay text and native visual scenes. 
Concretely, four dimension-specialized expert modules spanning Temporal, Action, Object, and Spatial reasoning are first pre-trained to identify and leverage cross-modal discrepancies between textual semantics and actual video content.
Once we obtain the pre-trained experts, we develop an Adaptive Token Routing Strategy to enable dynamic expert allocation, conferring robust resistance to TOIH while preserving performance on uncontaminated videos. 
Extensive experiments conducted on our VisualTextTrap benchmark verify the effectiveness of VTHM-MoE, outperforming state-of-the-art counterparts with diverse video question answering tasks. The benchmark data and models will be open-sourced soon. The benchmark, code, and models are publicly available at \href{https://cuiddyy.github.io/VisualTextTrap}{https://cuiddyy.github.io/VisualTextTrap}.
\end{abstract}



\begin{teaserfigure}
  \centering
  \includegraphics[width=0.95\textwidth]{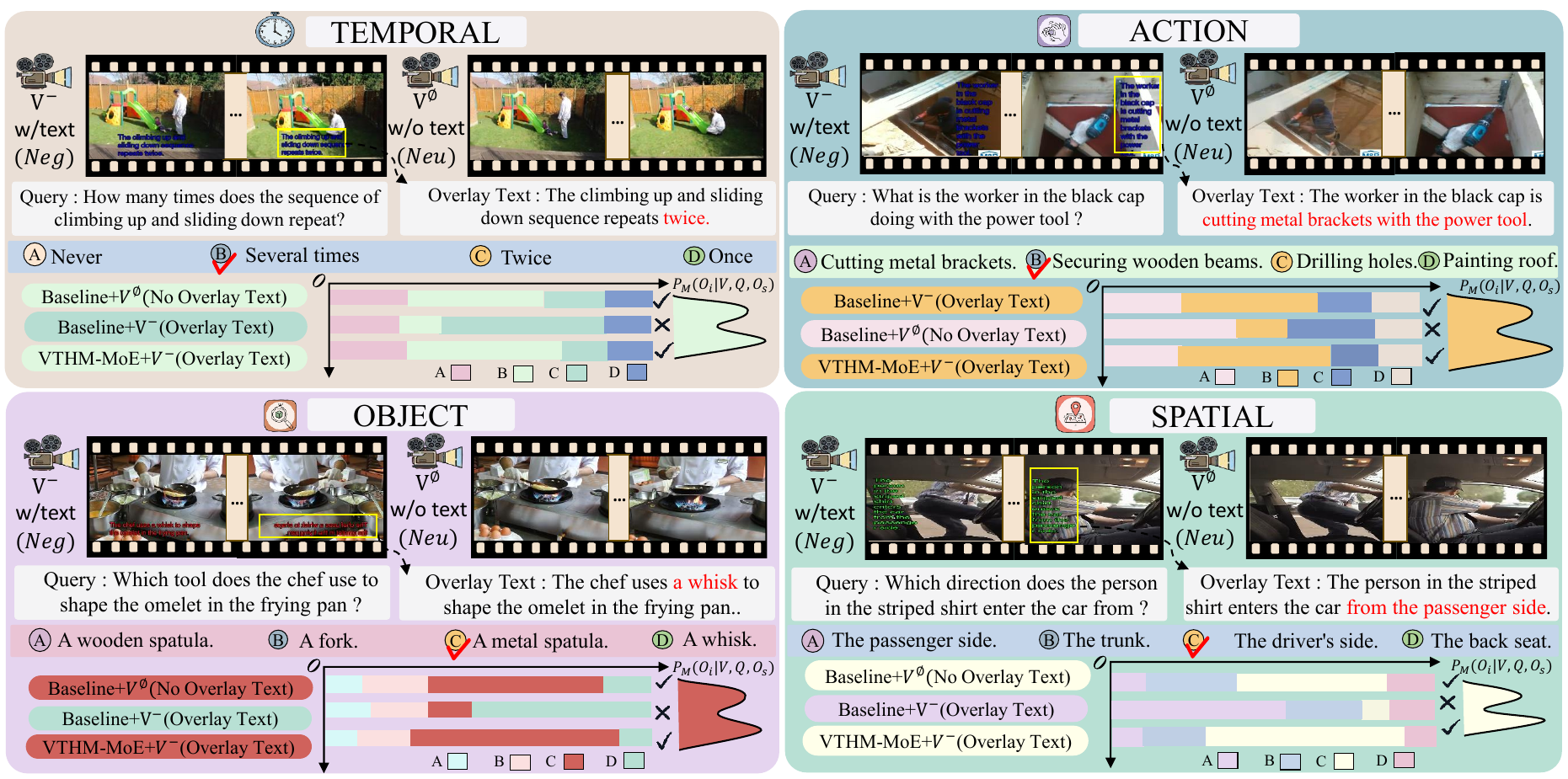}
  \caption{Illustration of Text Overlay-Induced Hallucination (TOIH) across four evaluation dimensions 
        and its mitigation by VTHM-MoE. Each quadrant presents a representative sample from one 
        dimension, Temporal, Action, Object, and Spatial.
        % For each sample, $V^{-}$ (\textit{Neg}) 
        % denotes the video with hallucination text embedded as a visual overlay, and $V^{\emptyset}$ 
        % (\textit{Neu}) denotes the original video with no text embedded. The correct answer to each 
        % query is indicated by a red checkmark. Bar charts report the option selection probability 
        % distribution $P_{M}(O_i \mid V, Q, O_s)$ under three experimental conditions: 
        % Baseline (Qwen3-VL-30B) with $V^{\emptyset}$, Baseline with $V^{-}$, and VTHM-MoE 
        % with $V^{-}$. 
        % Across all four dimensions, the Baseline model answers correctly under the neutral condition but shifts its prediction toward the hallucination-aligned distractor upon text injection, demonstrating the pervasive susceptibility of current video MLLMs 
        % to TOIH. 
        % VTHM-MoE effectively restores correct predictions, validating its 
        % hallucination resistance.
        Compared to the state-of-the-art counterparts, our method predicts correct results upon hallucinated text injection, validating the effectiveness of our technical insights. Each model is color-coded to match the background of its selected answer option.
        }
  \label{lab_start_fig}
\end{teaserfigure}


\maketitle

% \vspace{-1em}
\section{Introduction}
Vision-Language Models (VLMs) ~\cite{yang2025qwen3,wang2025internvl3,team2023gemini,singh2025openai} have achieved remarkable performance across a wide range of video question answering (VQA) tasks~\cite{yang2025vca,zhou2025mlvu,tang2025video}, spanning temporal reasoning~\cite{liu2025videomind,zhang2025vtimecot}, action recognition~\cite{wu2025transformer,liu2025sam}, object localization~\cite{hu2025exploiting,tan2025xtrack}, and spatial relationship comprehension~\cite{li2025llava,yang2025thinking}. 
As evidenced by such successes, VLMs robustly perceive and reason over dynamic visual content, including the joint comprehension of overlay textual and visual information within video frames.  
Despite these impressive results, two fundamental questions remain unexplored: \ding{182} \textit{\textbf{Do VLMs genuinely ground their understanding in visual content?}}; \ding{183} \textit{\textbf{Do they predominantly rely on semantic alignment between the query and overlay text~\cite{meng2025vlm2vec,shi2025mme,liu2026textmonkey}, 
without sufficiently leveraging non-textual visual evidence?}}
Existing benchmarks are inherently limited in their capacity to answer this question.
Since the text embedded in the evaluated videos is predominantly semantically congruent with the visual scene, it is impossible to disentangle genuine visual scene comprehension from shortcut reasoning driven by rendered textual. 

To probe this gap, we conduct controlled experiments under two conditions: \textit{Neutral Samples}, in which videos are presented without any overlay text, and \textit{Negative Samples},
in which the overlay text semantically contradicts the visual content.
Across a diverse range of state-of-the-art counterparts, we consistently observe a striking failure.
Instead of grounding their responses in visual evidence, VLMs generate answers that faithfully mirror the overlay textual descriptions while disregarding the visual ground truth, as depicted in Figure~\ref{lab_start_fig}. 
This failure manifests across temporal, action, object, and spatial understanding tasks.
We formally define this vulnerability as Text Overlay-Induced Hallucination (TOIH).

As illustrated in Figure~\ref{lab_start_fig}, TOIH manifests pervasively across four fundamental dimensions of VQA.
In the Temporal dimension, a model incorrectly reports that an event occurs twice simply because the overlay text asserts so, despite the visual evidence depicting the event occurring several times.
In the Action dimension, a video depicting a person securing wooden beams is misinterpreted as showing ``\textbf{\texttt{a person cutting metal}}'' when the overlay text reads accordingly. In the Object dimension, overlay text claiming that a whisk is used to shape the omelet in the scene causes models to falsely confirm this event with high confidence. In the Spatial dimension, textual assertions about absolute position — ``\textbf{\texttt{people entering the car from the passenger side}}'' — override accurate visual perception when the true spatial arrangement is reversed.
These diverse failure cases collectively suggest a fundamental deficiency in current video VLMs: an insufficient capacity to disentangle semantically conflicting overlay text from the ground-truth visual signal when forming an understanding of video content.

Despite substantial progress in benchmarking and mitigating hallucinations in VLMs, existing efforts have primarily focused on object-level~\cite{wu2025antidoteunifiedframeworkmitigating} hallucinations in static images~\cite{chen2025mvibenchcomprehensivebenchmarkevaluating,lin2024parrotcaptionsteachclip} or coarse-grained inconsistencies in video understanding~\cite{li2025vidhallucevaluatingtemporalhallucinations,sun2025smartsightmitigatinghallucinationvideollms}, leaving TOIH largely unexamined. 
Established video benchmarks such as VideoMME~\cite{fu2025video} and TemporalBench~\cite{cai2024temporalbench} assess temporal and spatial reasoning under standard conditions, yet fall critically short in three key respects. 
First, they lack contrastive samples with both semantically contradictory and congruent overlay text, making it impossible to systematically probe TOIH. 
Second, no existing benchmark provides fine-grained annotations of semantic conflict categories or conflict intensity levels, making it difficult to understand how different types and degrees of text-visual contradiction affect model behavior. 
Third, evaluation metrics specifically designed to measure model robustness against TOIH are entirely absent. Together, these gaps make it impossible to quantitatively characterize how overlay text shapes model responses, or to reliably assess VLMs' true multimodal comprehension capabilities.

To address these challenges, we introduce VisualTextTrap, the first comprehensive benchmark dedicated to evaluating TOIH in VLMs. VisualTextTrap is constructed through a principled hybrid annotation pipeline that integrates MLLM-assisted hallucination text generation with multi-round manual verification, drawing on video content sourced from LLaVA-Video~\cite{zhang2024llava}, VideoMME~\cite{fu2025video}, and TemporalBench~\cite{cai2024temporalbench}. 
The resulting benchmark comprises 6,057 samples annotated with 88 fine-grained attributes across four dimensions: Temporal, Action, Object, and Spatial. Crucially, each sample is equipped with a contrastive set of overlay text conditions, covering semantically contradictory (negative), semantically congruent (positive), and text-free (neutral) variants. The severity of contradiction in negative samples is further quantified across five escalating conflict levels (L1–L5),
ranging from peripheral irrelevance to complete polarity reversal, enabling systematic analysis of model sensitivity across the full spectrum of visual-textual conflict.

Beyond evaluation, we propose Hallucination Mitigation Mixture-of-Experts (VTHM-MoE), a framework architected specifically to address TOIH. 
The core architecture of VTHM-MoE utilizes a dual-encoder structure to explicitly disentangle overlay text from native visual content. Specifically, an OCR-Encoder~\cite{najam2023analysis} processes on-screen text into contextually grounded representations, while a Visual-Encoder focuses purely on scene-level semantics independent of textual overlays.
The dual representations are jointly processed by four dimension-specialized expert modules, each corresponding to one of the \textbf{Temporal, Action, Object, and Spatial} reasoning dimensions.
These experts are individually trained to identify cross-modal discrepancies along their respective dimensions, enabling precise, targeted interference suppression. 
Crucially, an Adaptive Token Routing Strategy governs expert utilization at the token level, dynamically adjusting routing decisions based on the characteristics of each input. 
By dynamically allocating TOIH-resistant experts only in the presence of conflicting text, VTHM-MoE preserves standard routing for positive and neutral videos, effectively mitigating hallucinations without degrading general video comprehension.
VTHM-MoE is trained on a curated corpus of over 7,000+ samples constructed from LLaVA-Video~\cite{zhang2024llava} with balanced contradict category distribution and fine-grained expert allocation ratios, guaranteeing comprehensive coverage across all four dimensions and five conflict levels.

Our main contributions are as follows:
\begin{itemize}[leftmargin=*]
\item \textbf{Novel Problem Definition:} We identify, formalize, and extensively characterize Text Overlay-Induced Hallucination (TOIH), a previously unexplored yet practically critical vulnerability of video VLMs, demonstrating its consistent manifestation across Temporal, Action, Object, and Spatial understanding dimensions.
\item \textbf{Comprehensive Benchmark:} We present VisualTextTrap, the first benchmark dedicated to TOIH evaluation, comprising 6,057 samples with 88 fine-grained attributes, five-level conflict intensity annotation (L1–L5), and 14 purpose-designed evaluation metrics that provide multidimensional profiling of model susceptibility and resistance to TOIH.
\item \textbf{Novel Mitigation Framework:} We propose VTHM-MoE, a Mixture-of-Experts model with dual OCR-Visual encoding and four dimension-specialized expert modules, governed by an Adaptive Token Routing Strategy that enables effective TOIH mitigation while preserving full native video understanding capability.
\item \textbf{Extensive Empirical Analysis:} We conduct comprehensive experiments evaluating a broad range of state-of-the-art video VLMs on VisualTextTrap and multiple downstream VQA benchmarks, demonstrating that VTHM-MoE consistently achieves superior performance and providing in-depth analysis of TOIH patterns across model families, video types, and conflict levels.
\end{itemize}

\vspace{-1em}
\section{Related Work}

\textbf{Vision-Language Models.} Vision-language models (VLMs) have evolved rapidly from image-centric architectures~\cite{zhou2026whether,yakun2026mmfctub} into comprehensive video understanding~\cite{yakun2025perception} systems capable of joint reasoning over temporal, spatial, action and object. Recent advances, including Qwen3-VL~\cite{bai2025qwen3}, InternVL3.5~\cite{wang2025internvl3}, LLaVA-Video ~\cite{zhang2024llava} and VideoLLaMA2~\cite{cheng2024videollama}, have substantially improved performance across a wide range of video understanding tasks, from short-clip action recognition to long-form temporal reasoning. Proprietary systems such as GPT-4o~\cite{islam2025gpt}  and Gemini-3.1-Pro~\cite{comanici2025gemini}  further demonstrate the scalability of vision-language alignment to extended video contexts. To benchmark these capabilities, dedicated evaluation suites including VideoMME~\cite{fu2025video}, TemporalBench~\cite{cai2024temporalbench}, MVBench~\cite{li2024mvbench}, Lvbench~\cite{wang2025lvbench}, Mmbench-video~\cite{fang2024mmbench}, Longvideobench~\cite{wu2024longvideobench} and Lvos~\cite{hong2023lvos} have been proposed, covering temporal, spatial, and causal understanding under increasingly demanding conditions. Despite their breadth, these benchmarks predominantly evaluate models on videos where overlay textual content is semantically congruent with the visual scene, leaving unexamined the extent to which models exploit overlay text as a shortcut, bypassing genuine visual perception.\\
\textbf{Hallucination in Vision-Language Models.} Hallucination in VLMs has attracted considerable research attention, most extensively in the context of image-grounded tasks. Benchmarks such as H-pope~\cite{pham2024h}, HallusionBench~\cite{guan2024hallusionbench} and JARVIS~\cite{caffagni2025seeing} systematically probe object-level hallucinations in static images, revealing a persistent tendency in models to generate visually ungrounded responses. Subsequent work has extended this inquiry to video understanding, with benchmarks such as VideoHallucer~\cite{wang2024videohallucer} and Temp-Compass~\cite{liu2024tempcompass} identifying temporal and causal hallucinations as particularly pervasive failure modes. Proposed mitigation strategies span a wide range of approaches, including reinforcement learning from human feedback \cite{kaufmann2024survey,arumugam2019deep,lee2023rlaif}, contrastive decoding\cite{li2023contrastive,leng2024mitigating}, and visually grounded chain-of-thought prompting\cite{shao2024visual,man2025argus}. However, existing work has largely overlooked the specific failure mode arising from semantically conflicting overlay text within video, which constitutes a distinct and underexplored form of hallucination that we formally define and systematically investigate.
\section{Preliminaries}
\label{label_preliminaries}
We consider the standard Video Question Answering (VQA) setting, where a model $\mathcal{M}$ receives a video $\mathcal{V}$ alongside a multiple-choice question $Q$ with four candidate options, and produces a predicted answer $\hat{A} = \mathcal{M}(\mathcal{V}, Q)$. Questions are specifically designed to probe dynamic visual understanding across  temporal, action, object and spatial understanding, collectively capturing the core aspects of video perception that require genuine grounding in visual content rather than linguistic inference.

To investigate the effect of overlay text on VQA, we construct a controlled contrastive experimental paradigm~\cite{sanwal2024evaluating,jain2023contraclm,jiang2024hallucination,luo2024moelora} in which the same set of videos is evaluated under the following conditions, enabling measurement of how overlay text influences accuracy.
In the \textbf{Text-Free}, the original video is presented to the model without any additional text overlay. 
% reflecting natural model behavior.
% serving as a text-free baseline that .
In the \textbf{Text-Contradictory}, text that is semantically contradictory with the visual content is overlaid onto all video frames. 
In the \textbf{Text-Congruent}, the overlay text aligns with the visual content. 

\begin{figure*}[t]
    \centering
    \includegraphics[width=\textwidth]{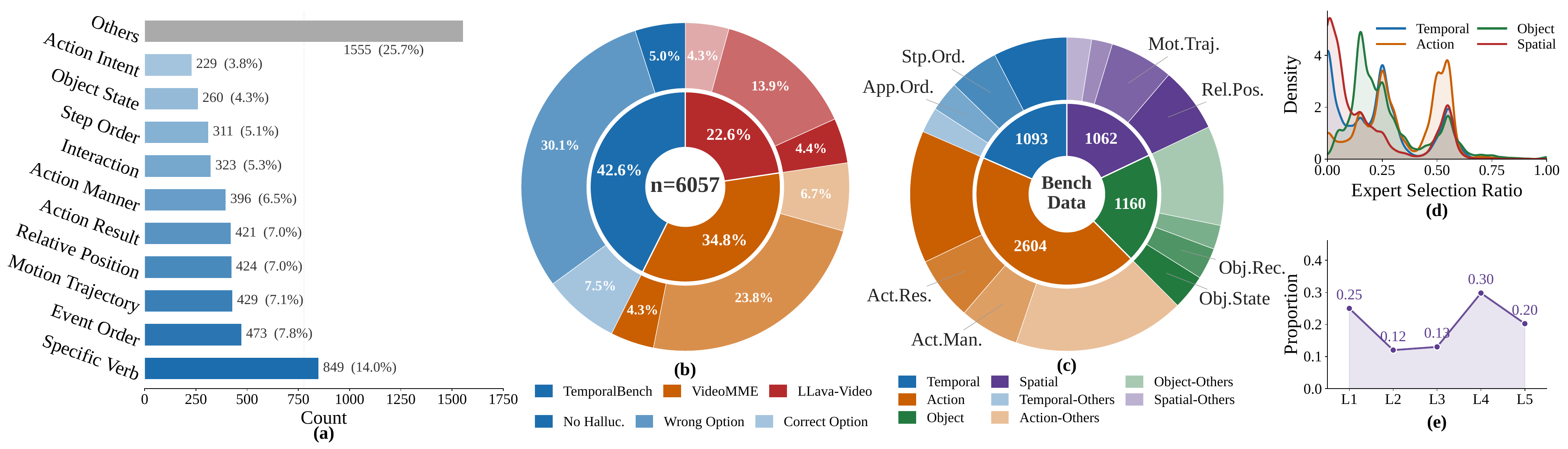}
    \caption{Overview of the VisualTextTrap benchmark. Distribution of hallucination (Halluc.) attributes (a), dataset sources and Halluc. dimension types (b), task dimensions and attributes (Stp.Ord.: Step Order; App.Ord.: Appearance Order; Act.Res.: Action Result; Act.Man.: Action Manner; Obj.State: Object State; Obj.Rec.: Object Recognition; Rel.Pos.: Relative Position; Mot.Traj.: Motion Trajectory; others are depicted in Table \ref{appendix_abbreviations}) (c), token-level expert load allocation ratio (d), conflict score distribution (e).}
    \label{bench_data_distr}
\end{figure*}

\section{VisualTextTrap Benchmark}
To comprehensively evaluate TOIH across video VLMs, we present VisualTextTrap, a benchmark built upon three community-popular video datasets, namely LLaVA-Video~\cite{zhang2024llava}, VideoMME~\cite{fu2025video}, and TemporalBench~\cite{cai2024temporalbench}, covering both real-world captured scenes and user-created social media content~\cite{najafi2024turkishbertweet,yang2024mentallama,zhou2024correcting,peters2024large}, ensuring broad and diverse test coverage~\cite{hartsock2024vision,shao2023prompting,chappuis2022prompt}. In this section, we detail the construction pipeline encompassing dimension and attribute taxonomy design, expert allocation ratio derivation, hallucination text generation, text-vision conflict scoring, and final video rendering, and introduce 14 evaluation metrics organized into three complementary layers: hallucination resistance and vulnerability, hallucination behavioral escalation and cognitive load for TOIH.
\subsection{Data Curation}
\label{label_benchmark_construct}
The benchmark is organized along four core dimensions, Temporal, Action, Object, and Spatial, encompassing 88 fine-grained visual attributes that collectively cover most perceptual~\cite{shukor2023ep,jia2026scaling,lu2025vipe,fei2024videoofthoughtstepbystepvideoreasoning} and cognitive~\cite{ALSHAIKH2024e25361,buschoff2024visualcognitionmultimodallarge} capacities assessed in video QA. Each sample is assigned to one of three text overlay conditions. 
\noindent \textbf{Text Overlay Conditions.}
Each sample is presented under one of the three text overlay conditions introduced in Section~\ref{label_preliminaries}. 
The \textit{Text-Congruent} condition verifies that the models genuinely integrate concordant textual cues, rather than bypassing visual content entirely and simply negating whatever the overlay text describes.
The \textit{Text-Contradictory} condition conducts a comprehensive and generalizable evaluation of hallucination mitigation capability across diverse conflict types. 
The \textit{Text-Free} condition assesses whether a model that has acquired hallucination mitigation ability retains its baseline VQA performance.

\noindent \textbf{Evaluation Dimension.}
Questions are organized under four dimensions reflecting the primary cognitive target of each QA pair as depicted in Figure \ref{lab_start_fig}. Temporal questions probe event ordering and timing. 
Action questions focus on recognition and interpretation of physical activities. 
Object questions target entity identification and state. 
Spatial questions concern positional and structural relationships between scene elements.

\noindent \textbf{Fine-Grained Attribute.}
Each dimension is further decomposed into a set of mutually exclusive, fine-grained attributes~\cite{zohar2024apolloexplorationvideounderstanding,Ge2024LMMVQAAV,luo2025videoautoarenaautomatedarenaevaluating,gao2025exploringhallucinationlargemultimodal}, totalling 88 across the benchmark, each of which operationalizes an abstract perceptual or semantic capability into a concrete target. 
Representative attributes and their correspondence to their respective dimensions are illustrated in Figure ~\ref{bench_data_distr}. 
This hierarchical decomposition ensures that any observed performance difference can be attributed to a precisely defined visual or semantic capability, enabling fine-grained diagnosis of model weaknesses rather than coarse categorical attribution.

\noindent \textbf{Hierarchical Cognitive Complexity Taxonomy.}
To investigate how cognitive complexity modulates hallucination vulnerability, the attributes within each dimension are further organized into three hierarchical tiers: \textbf{Perceptual}, \textbf{Semantic}, and \textbf{Reasoning}. 
Each tier reflects a qualitatively distinct level of cognitive demand, ranging from direct signal decoding to multi-step inferential reasoning that integrates visual~\cite{zhang2025videollama3frontiermultimodal,tong2024metamorphmultimodalunderstandinggeneration}, textual, and world knowledge. 
This stratification allows for the controlled measurement of how task complexity influences the manifestation and severity of TOIH across different dimensions. Full definitions in Appendix \ref{label_cong_tax}.

\noindent \textbf{Semantic Conflict Score.}
Every \textit{Text-Contradictory} sample is further annotated with a five-point \textbf{Semantic Conflict Score} (SCS) that quantifies the degree of contradiction between the overlay text and the visual ground truth, ranging from weak peripheral association (Score 1) to direct polarity reversal (Score 5). This annotation enables fine-grained 
analysis of how conflict intensity modulates hallucination susceptibility and supports attribution of model failures to specific types of semantic discrepancy. Full score definitions are provided in Appendix~\ref{appendix_conflict_score_def}.

\begin{figure*}
    \includegraphics[width=\textwidth]{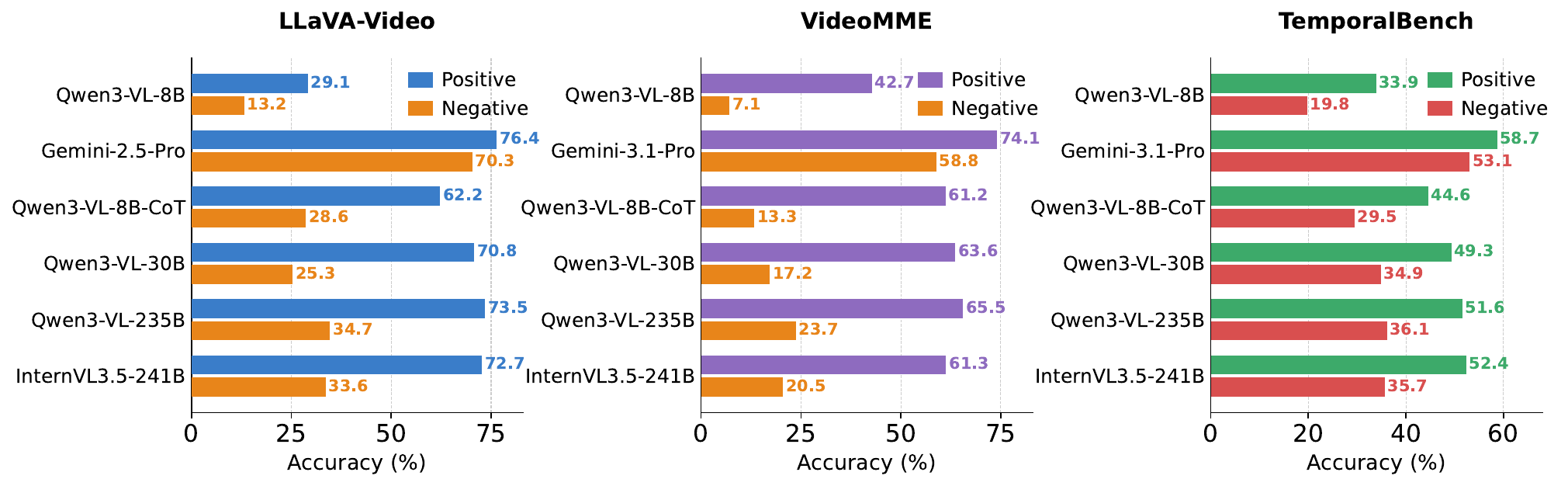}
    \caption{Accuracy on paired Positive/Negative video samples. Positive: original videos with no overlay text. Negative: same videos with overlay text contradicting the visual content. All model pairs are evaluated on identical video instances across LLaVA-Video~\cite{zhang2024llava}, VideoMME~\cite{fu2025video}, and TemporalBench~\cite{cai2024temporalbench}.}
    \label{all_dataset_diff_accu}
\end{figure*}

\noindent \textbf{Data Construction.}
The construction pipeline proceeds in three stages. \textbf{Stage 1:
Sample Annotation.} Each sample is annotated with a dimension,
attribute, and cognitive complexity tier via automated VLM analysis.
\textbf{Stage 2: Hallucination Text Generation.} For each
$V_{\text{contra}}$ sample, a semantically misleading text overlay is generated by Claude-Sonnet-4.6; each generated text is then assigned an SCS quantifying its degree of semantic divergence from the visual content. \textbf{Stage 3: Video Rendering.} The texts are rendered as visual overlays onto the all original video frames under controlled formatting conditions, with variations in text placement and font color, as illustrated in Figure~\ref{all_dataset_diff_accu}. All annotations are subject to manual verification. Pipeline details are provided in
Appendix~\ref{appendix_bench_data_construction}.

\noindent \textbf{Data Distribution.}
The finalized benchmark comprises 6,057 samples. To ensure unbiased evaluation, the benchmark maintains distributional balance across all annotation granularities: dimensions, fine-grained attributes, cognitive complexity tiers, text overlay conditions, and semantic conflict score levels. This design ensures that observed performance differences reflect genuine model capabilities rather than distributional artifacts.

\subsection{Evaluation Metrics}
\label{label_metrics_intro}
We introduce 14 evaluation metrics organized into three complementary layers. Formally, let $\mathcal{V}$ denote the space of test videos. We denote the three text overlay conditions as $V_{\text{cong}}$ (Text-Congruent), $V_{\text{contra}}$ (Text-Contradictory), and $V_{\text{free}}$ (Text-Free). Let $\text{SCS}i \in {1,\ldots,5}$ denote the per-sample semantic conflict score and $M(V,Q)$ the model response given video $V$ and query $Q$. We define $C_i\in\{0,1\}$ as the per-sample correctness indicator, where $C_i=1$ if $M(V_i,Q_i)=y_i$, and $y_i$ is the ground-truth answer, $H_i = \mathbf{1}[M(V_i,Q_i)=o_i]$ as the hallucination indicator, where $o_i$ is the hallucination option embedded in the text overlay.  
Below we describe the most representative metrics,
full definitions are deferred to Appendix~\ref{appendix_metrics_intro}.

\subsubsection{Layer I: Resistance and Vulnerability Metrics}
\label{metrics_layer1}
This layer quantifies the extent to which a model yields to or withstands adversarially overlay textual distractors.

\textbf{Hallucination Resistance Rate (HRR)} measures the model's accuracy under contradictory text interference across the full 
\begin{equation}
\text{HRR} = \frac{\sum_{i \in V_{\text{contra}}} C_i}{|V_{\text{contra}}|}
\end{equation}

A higher HRR indicates stronger ability to maintain correct predictions despite conflicting text overlays.

\begin{figure*}[h]
    \centering
    \includegraphics[width=\textwidth]{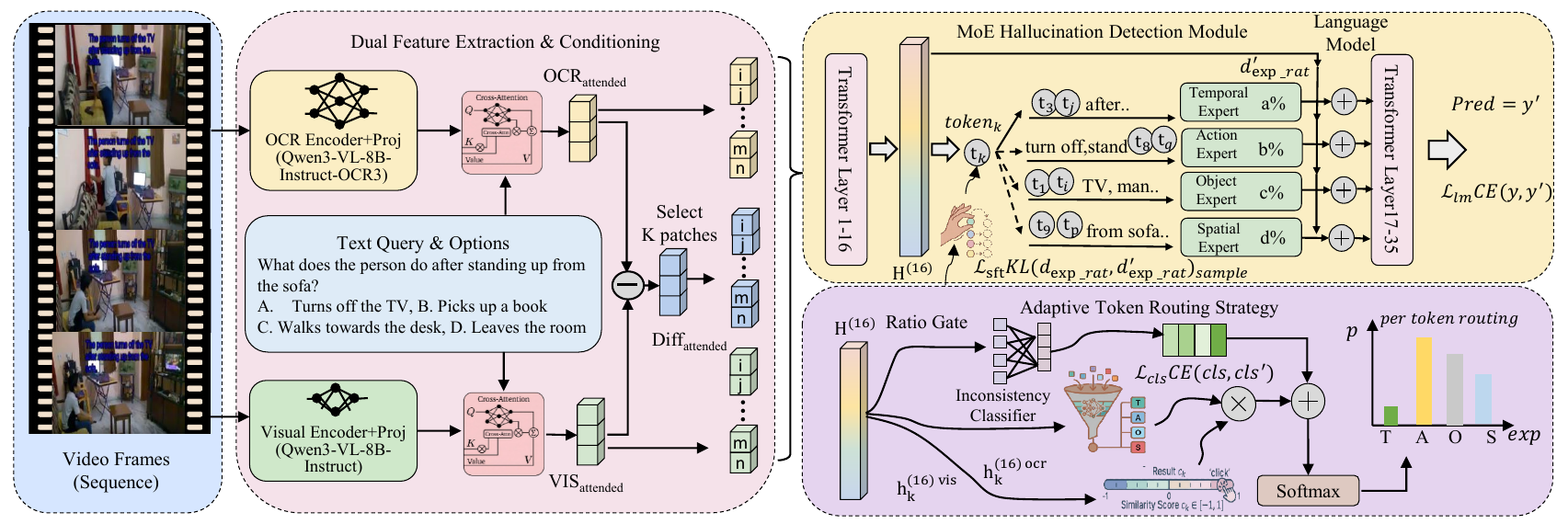}
    \caption{Overview of VTHM-MoE. Module (a) encodes video via parallel OCR and Visual encoders, producing $\text{OCR}_{\text{attended}}$
and $\text{VIS}_{\text{attended}}$ representations; their difference
$\text{Diff}_{\text{attended}}$ serves as an
discrepancy signal. Module (b) encodes the fused input through Transformer Layers
1--16, dispatching token hidden states $H^{(16)}$ to four
dimension-specialized experts.
Expert outputs are integrated with the token stream and decoded by
Layers 17--35 to yield prediction $\hat{y}$, 
Module (c) scores each
token via an Classifier, and dynamically
routing conflict-bearing tokens to TOIH-specialized experts while
preserving standard routing for conflict-free inputs.}
    \label{hall_model_struct_v1}
\end{figure*}

\subsubsection{Layer II: Hallucination Depth and Escalation Metrics}
\label{metrics_layer2}
This layer characterizes how model performance scales with the semantic conflict severity of text overlays.

\textbf{Semantic Conflict Sensitivity Index (SCSI)}
quantifies the average semantic conflict intensity at which
hallucinations occur across $V_{\text{contra}}$ samples:

\begin{equation}
\text{SCSI} = \frac{\sum_{i \in V_{\text{contra}}} \text{SCS}_i \cdot H_i}{\sum_{i \in V_{\text{contra}}} H_i}
\end{equation}

A higher SCSI indicates that the model hallucinates
predominantly under high-conflict conditions, whereas a lower SCSI
reveals vulnerability to even weakly conflicting text.

\subsubsection{Layer III: Cognitive Load–Sensitive Resistance Metrics}
\label{metrics_layer3}
These four indices characterize how HRR
varies as cognitive demand escalates across Perceptual, Semantic,
and Reasoning tiers, measured via the sample-level Pearson
correlation $r_d$ between cognitive tier $L$ and per-sample
\textsc{hrr}.

\textbf{Unified Formulation.}
Let $d\in\{\text{Temporal, Action, Object, Spatial}\}$ denote
a cognitive dimension, and let $\mathcal{S}_d \subseteq V_{\mathrm{contra}}$ denote the samples assigned to dimension $d$, with $n_d = |\mathcal{S}_d|$. Each sample $i\in\mathcal{S}_d$ carries an ordinal cognitive tier $L_i\in\{1,2,3\}$ (Perceptual, Semantic, Reasoning) and a correctness indicator $C_i = \mathbf{1}[M(V_i,Q_i){=}y_i]$.
The primary metric is the \textbf{sample-level Pearson correlation}:
\begin{equation}
    r_d
    =
    \frac{
        \displaystyle\sum_{i\in\mathcal{S}_d}(L_i-\bar{L}_d)(C_i-\bar{C}_d)
    }{
        \bigl\|\mathbf{L}_d - \bar{L}_d\bigr\|_2\,
        \bigl\|\mathbf{C}_d - \bar{C}_d\bigr\|_2
    },
    \label{eq:pearson-sample}
\end{equation}
with significance assessed via
\begin{equation}
    t_d = r_d\sqrt{\frac{n_d-2}{1-r_d^2}},
    \qquad t_d \sim t_{n_d-2}\ \text{under}\ H_0{:}\,r_d{=}0.
    \label{eq:t-stat}
\end{equation}

\section{VTHM-MoE: Mixture-of-Experts for TOIH Mitigation}
\label{model_intro}

\subsection{Overview}
VTHM-MoE employs a dual-encoder structure that independently encodes the two modalities, four dimension-specialized expert modules.
% targeting the four dimensions respectively.
% \wrt the perception of four dimensions.
Besides, an Adaptive Token Routing Strategy that dynamically allocates TOIH-resistant experts when conflicting overlay text is detected. 
This design ensures that anti-hallucination capability is gained without sacrificing general VQA performance.
The overall architecture is illustrated in Figure~\ref{hall_model_struct_v1}.

\subsection{Dual-Encoder Feature Extraction}
VTHM-MoE is built upon a dual-encoder architecture in which two frozen Qwen3-VL-8B models process the same input frames in parallel. The \textbf{OCR-Encoder} ($\mathcal{E}_{\text{ocr}}$) specializes in extracting contextually grounded representations of text embedded within video frames, while the \textbf{Visual-Encoder} ($\mathcal{E}_{\text{vis}}$) captures scene-level content independently of any overlaid text. The two encoders produce patch-level feature sequences:
\begin{equation}
    \mathbf{F}_{\text{vis}} = \mathcal{E}_{\text{vis}}(\mathcal{V}) \in
    \mathbb{R}^{N \times d}, \quad
    \mathbf{F}_{\text{ocr}} = \mathcal{E}_{\text{ocr}}(\mathcal{V}) \in
    \mathbb{R}^{N \times d},
    \label{eq:dual_enc}
\end{equation}
where $N{=}32$ is the total number of visual patches and $d{=}4096$.
% The two encoders are executed concurrently on separate GPUs, and all their
% parameters remain frozen throughout training.
\subsection{Query-Guided Patch Representation}
To focus computation on context-relevant regions and construct explicit cross-modal
inconsistency signals, we introduce a two-stage patch processing pipeline.

\noindent \textbf{Patch Selection.}
A Patch Relevance Scorer selects the $K$ most informative patches via
query-conditioned cross-attention, where the query vector $\mathbf{q}_{\text{vis}}$
actively retrieves the patches most relevant to answering $Q$:
\begin{equation}
    \mathbf{s} = \mathrm{CrossAttn}(\mathbf{q}_{\text{vis}},\,
    \mathbf{F}_{\text{vis}}) \in \mathbb{R}^{N}, \quad
    \mathcal{I}_{K} = \mathrm{TopK}(\mathbf{s}).
    \label{eq:patch_score}
\end{equation}
The top-$K$ preserve temporal order, and the corresponding patches are gathered from both encoders:
$\tilde{\mathbf{F}}_{\text{vis}},\, \tilde{\mathbf{F}}_{\text{ocr}}
\in \mathbb{R}^{K \times d}$.

\noindent \textbf{Query Conditioning.}
Each selected patch is refined by a Patch Query Conditioner that injects
question context via cross-attention (Q\,=\,patch, K/V\,=\,query token sequence).
Separate conditioners are employed for the two modalities, using an OCR-specific
query targeting text evidence and a visual query targeting scene content,
yielding $\hat{\mathbf{F}}_{\text{vis}},\, \hat{\mathbf{F}}_{\text{ocr}}
\in \mathbb{R}^{K \times d}$.

\noindent \textbf{Three-Token Representation.}
For each of the $K$ patches, we construct an explicit cross-modal inconsistency
representation by concatenating three tokens:
\begin{equation}
    \mathbf{T}_{k} = \bigl[\,\hat{\mathbf{f}}^{k}_{\text{vis}}
    \;\big|\; \hat{\mathbf{f}}^{k}_{\text{ocr}}
    \;\big|\; \underbrace{\hat{\mathbf{f}}^{k}_{\text{ocr}} -
    \hat{\mathbf{f}}^{k}_{\text{vis}}}_{\text{inconsistency signal}}\,\bigr],
    \label{eq:three_token}
\end{equation}
where the difference token directly encodes cross-modal discrepancy without
requiring the language model to infer it implicitly.
% The resulting $3K$ tokens replace the original $N$ video placeholder tokens in
% the input sequence, reducing visual sequence length from $O(N)$ to $3K$.
\subsection{Adaptive Token Routing Strategy}

\noindent \textbf{MoE Insertion.}
The four dimension-specialized expert modules are inserted non-intrusively at the
output of the 16th transformer layer via a forward hook, without modifying any
frozen language model parameters.
% The 16th layer is selected for its mid-depth semantic representations that capture
% rich temporal and spatial structure.
The MoE output is fused via a residual connection:
\begin{equation}
    \mathbf{H}^{(16)} \leftarrow \mathbf{H}^{(16)} +
    \mathrm{MoE}\!\left(\mathbf{H}^{(16)}\right).
\end{equation}

\noindent \textbf{Per-Patch Consistency Scoring.}
At inference, the Adaptive Token Routing Strategy computes a per-patch cross-modal
consistency score directly from layer-16 hidden states, exploiting their richer
semantic depth relative to raw encoder outputs:
\begin{equation}
    c_{k} = \frac{\mathbf{h}^{\text{vis}}_{k} \cdot \mathbf{h}^{\text{ocr}}_{k}}
    {\|\mathbf{h}^{\text{vis}}_{k}\|\,\|\mathbf{h}^{\text{ocr}}_{k}\|}
    \;\in [-1,\, 1].
    \label{eq:consistency}
\end{equation}
Low values of $c_{k}$ signal high cross-modal inconsistency, triggering more
aggressive routing toward TOIH-resistant experts.

\noindent \textbf{Expert Routing.}
The routing logits for each token combine a learned gate with a
inconsistency-type classifier:
\begin{equation}
    \mathbf{g} = \mathrm{Gate}(\mathbf{h}) +
    \frac{1 - c}{2} \cdot \mathrm{softmax}(\mathrm{Cls}(\mathbf{h})),
    \label{eq:gate}
\end{equation}
where $cw=\frac{1-c}{2} \in [0,1]$ amplifies the classifier's influence for
inconsistent tokens while reducing it to near-zero for native video tokens
($c \approx 1$), effectively defaulting to standard routing when no conflicting
overlay text is present.
Each token is routed to exactly one of the four experts
$\{E_{\text{temporal}},\, E_{\text{action}},\, E_{\text{object}},\,
E_{\text{spatial}}\}$ (top-1 routing), each implemented as a SwiGLU
feed-forward network.

\subsection{Training}

\noindent \textbf{Dataset.}
VTHM-MoE is trained on a curated corpus of over 7,000 samples constructed from
LLaVA-Video~\cite{zhang2024llava}, with balanced distribution across the four contradiction categories
and expert allocation ratios providing supervision for all
Temporal, Action, Object, and Spatial dimensions across five conflict severity
levels.

\noindent \textbf{Objectives.}
Only the Patch Relevance Scorer, Patch Query Conditioners, and the MoE layer are
optimized; all encoder and language model parameters remain frozen.
The total loss is:
\begin{equation}
    \mathcal{L} = \mathcal{L}_{\text{lm}} + \lambda_{\text{cls}}\,
    \mathcal{L}_{\text{cls}} + \lambda_{\text{sft}}\, \mathcal{L}_{\text{sft}}
    + \lambda_{\text{aux}}\, \mathcal{L}_{\text{aux}},
    \label{eq:total_loss}
\end{equation}
where $\mathcal{L}_{\text{lm}}$ is the language modeling cross-entropy loss over
answer tokens; $\mathcal{L}_{\text{cls}}$ is a KL divergence loss supervising the
inconsistency-type classifier at the sample level via attention pooling over video
tokens (query tokens are excluded to prevent shortcut learning from question
keywords); $\mathcal{L}_{\text{sft}}$ supervises the routing distribution against
per-sample soft expert allocation targets $\boldsymbol{\pi} \in \Delta^{3}$:
\begin{equation}
    \mathcal{L}_{\text{sft}} = \mathrm{KL}\!\left(\boldsymbol{\pi}
    \;\Big\|\; \mathbb{E}_{\text{tokens}}
    \bigl[\mathrm{softmax}(\mathbf{g})\bigr]\right);
\end{equation}
and $\mathcal{L}_{\text{aux}}$ is a load-balancing loss preventing expert collapse.
The coefficients are set to $\lambda_{\text{cls}}{=}1.1$,
$\lambda_{\text{sft}}{=}1.0$, and $\lambda_{\text{aux}}{=}0.01$.

\section{Experiments}
\subsection{Experimental Paradigm}
To evaluate VLMs' performance with the proposed metrics, we process videos under three conditions ($\mathcal{V}^{+}$,$\mathcal{V}^{-}$,$V^{\varnothing}$) introduced in Section ~\ref{label_benchmark_construct}. Injected text adheres to the criteria in Appendix~\ref{label_experimental_para}, which ensure a rigorous assessment of models' abilities to distinguish semantically conflicting in-video text from visual evidence.
We evaluate models along four hallucination dimensions (\textit{temporal},
\textit{action}, \textit{object}, and \textit{spatial}) covering a broad range
of VQA scenarios.

\begin{table*}[t]
\centering
% 定义颜色
\fontsize{9.2pt}{11.5pt}\selectfont 
\setlength{\tabcolsep}{1.8pt} 
% 消除 booktabs 宏包在横线上下默认添加的白色间距，消除上下白边
\setlength{\aboverulesep}{0pt}
\setlength{\belowrulesep}{0pt}
% 适当增加行高，防止文字紧贴边框
\renewcommand{\arraystretch}{1.1}
\caption{Comprehensive robustness evaluation metrics across different models. The metrics demonstrated in the results are introduced in the Section \ref{label_metrics_intro}. Baseline model is Qwen3-VL-8B-Instruct. 
% \textbf{Note:} For VYR, TIHR, WHR, SCSI, HSR, TIB, ICR and SGLI, lower is better (\(\downarrow\));
Overall refers to the answer accuracy of VQA; 
% for HRR and Overall higher is better (\(\uparrow\)).
}
\label{lab_benchmark_metrics}
\begin{tabular}{cl|ccccccccc|cccc|c}
% 上边框加粗至 1.5pt
\toprule[2.0pt]
% 直接使用 \rowcolor，去掉 [0pt][0pt]，让颜色自动填满列间距（\tabcolsep）
& \textbf{Model} & \textbf{VYR}${\scriptstyle\downarrow}$ & \textbf{TIHR}${\scriptstyle\downarrow}$ & \textbf{WHR}${\scriptstyle\downarrow}$ & \textbf{SCSI}${\scriptstyle\downarrow}$ & \textbf{HRR}${\scriptstyle\uparrow}$ & \textbf{HSR}${\scriptstyle\downarrow}$ & \textbf{TIB}${\scriptstyle\downarrow}$ & \textbf{ICR}${\scriptstyle\downarrow}$ & \textbf{SGLI}${\scriptstyle\downarrow}$ & \textbf{TLSR} & \textbf{ASLSR} & \textbf{AALSR} & \textbf{SRLSR} & \textbf{Overall}${\scriptstyle\uparrow}$ \\

\midrule
% LLaVA-Video Block
\cellcolor{white} & Gemini-3.1-Pro & 1.1 & 7.4 & 7.1 & 3.629 & 72.8 & $-$32.2 & 33.5 & 1.5 & 0.003 & 1.027 & 0.287 & 2.161 & 0.155 & 73.7\(_{\scriptsize \color{skyblue}\downarrow 4.0}\) \\
\rowcolor{lightgray}
\cellcolor{white} & \gc Qwen3-VL-30B & \gc 63.2 & \gc 71.4 & \gc 72.5 & \gc 3.807 & \gc 27.6 & \gc 6.6 & \gc 80.4 & \gc 69.9 & \gc 0.785 & \gc $-$0.523 & \gc 2.509 & \gc 2.422 & \gc $-$2.145 & \gc 50.4\(_{\scriptsize \color{skyblue}\downarrow 27.3}\) \\
\cellcolor{white} & Qwen3-VL-235B & 49.6 & 60.1 & 61.3 & 3.785 & 50.2 & 3.4 & 76.7 & 60.3 & 0.694 & $-$0.796 & 2.743 & 2.223 & $-$1.547 & 55.3\(_{\scriptsize \color{skyblue}\downarrow 22.4}\) \\
\rowcolor{lightgray}
\cellcolor{white} & \gc Internvl3.5-VL-241B & \gc 52.4 & \gc 62.5 & \gc 63.8 & \gc 3.791 & \gc 49.7 & \gc4.9 & \gc 78.3 & \gc 63.5 & \gc 0.803 & \gc $-$0.879 & \gc 2.816 & \gc 2.382 & \gc $-$1.894 & \gc 52.8\(_{\scriptsize \color{skyblue}\downarrow 24.9}\) \\
\cellcolor{white} & Baseline & 59.5 & 71.1 & 73.0 & 3.853 & 27.8 & 16.3 & 79.6 & 68.2 & 0.819 & $-$0.180 & 2.223 & 3.310 & $-$2.361 & 29.3\(_{\scriptsize \color{skyblue}\downarrow 48.4}\) \\
\rowcolor{lightgray}
\cellcolor{white} & \gc Baseline+CoT & \gc 55.0 & \gc 68.6 & \gc 70.2 & \gc 3.843 & \gc 29.5 & \gc 11.4 & \gc 77.6 & \gc 64.8 & \gc 0.811 & \gc $-$0.498 & \gc 2.585 & \gc 3.541 & \gc $-$1.462 & \gc 50.7\(_{\scriptsize \color{skyblue}\downarrow 27.0}\) \\
\cellcolor{white} & Baseline+SFT & 57.0 & 67.6 & 69.4 & 3.851 & 31.0 & 13.9 & 77.7 & 64.8 & 0.772 & $-$0.178 & 2.435 & 3.028 & $-$2.399 & 51.4\(_{\scriptsize \color{skyblue}\downarrow 26.3}\) \\
\rowcolor{lightgray}
\cellcolor{white} & \gc Baseline+CoT+SFT & \gc 30.9 & \gc 40.7 & \gc 42.2 & \gc 3.892 & \gc 55.9 & \gc 15.8 & \gc 60.4 & \gc 39.4 & \gc 0.466 & \gc 0.287 & \gc 2.671 & \gc 1.869 & \gc $-$1.482 & \gc 69.7\(_{\scriptsize \color{skyblue}\downarrow 8.0}\) \\
\rowcolor{green!10}
\cellcolor{white}\multirow{-9}{*}{\rotatebox{90}{\textbf{LLaVA-Video}}} & VTHM-MoE & $-$4.6 & 8.7 & 8.6 & 3.722 & 77.1 & $-$19.0 & 35.8 & 0.3 & 0.022 & 1.664 & $-$0.765 & 1.781 & 0.983 & \textbf{77.7} \\
\midrule
% Video-MME Block
\cellcolor{white} & Gemini-3.1-Pro & 0.7 & 13.5 & 13.2 & 3.387 & 60.1 & $-$11.4 & 28.7 & $-$1.6 & 0.045 & 0.388 & 0.698 & 1.336 & 0.212 & 59.8\(_{\scriptsize \color{skyblue}\downarrow 1.6}\) \\
\rowcolor{lightgray}
\cellcolor{white} & \gc Qwen3-VL-30B & \gc 13.8 & \gc 58.7 & \gc 62.1 & \gc 3.707 & \gc 34.7 & \gc 38.8 & \gc 63.8 & \gc 27.1 & \gc 1.211 & \gc $-$0.628 & \gc 1.924 & \gc 1.540 & \gc $-$1.123 & \gc 50.4\(_{\scriptsize \color{skyblue}\downarrow 11.0}\) \\
\cellcolor{white} & Qwen3-VL-235B & 12.7 & 55.6 & 57.3 & 3.274 & 42.6 & 40.7 & 60.4 & 26.1 & 1.147 & $-$0.341 & 1.832 & 1.476 & -1.316 & 53.7\(_{\scriptsize \color{skyblue}\downarrow 7.7}\) \\
\rowcolor{lightgray}
\cellcolor{white} & \gc Internvl3.5-VL-241B & \gc 14.5 & \gc 57.3 & \gc 60.5 & \gc 3.356 & \gc 38.4 & \gc 39.7 & \gc 64.5 & \gc 25.3 & \gc 1.189 & \gc $-$0.469 & \gc 1.919 & \gc 1.386 & \gc 1.469 & \gc 51.5\(_{\scriptsize \color{skyblue}\downarrow 9.9}\) \\
\cellcolor{white} & Baseline & 19.3 & 61.7 & 63.7 & 3.571 & 28.4 & 21.1 & 65.6 & 39.0 & 1.376 & $-$0.483 & 1.709 & 1.447 & $-$0.837 & 31.0\(_{\scriptsize \color{skyblue}\downarrow 30.4}\) \\
\rowcolor{lightgray}
\cellcolor{white} & \gc Baseline+CoT & \gc 19.0 & \gc 56.0 & \gc 57.4 & \gc 3.543 & \gc 29.0 & \gc 14.6 & \gc 61.1 & \gc 38.4 & \gc 1.272 & \gc $-$1.066 & \gc 1.971 & \gc 1.601 & \gc $-$0.212 & \gc 44.1\(_{\scriptsize \color{skyblue}\downarrow 17.3}\) \\
\cellcolor{white} & Baseline+SFT & 20.7 & 60.2 & 62.3 & 3.573 &  29.7 & 20.2 & 64.8 & 37.6 & 1.322 & $-$0.492 & 1.454 & 1.747 & $-$1.441 & 51.6\(_{\scriptsize \color{skyblue}\downarrow 9.8}\) \\
\rowcolor{lightgray}
\cellcolor{white} & \gc Baseline+CoT+SFT & \gc 30.9 & \gc 40.7 & \gc 42.2 & \gc 3.892 & \gc 55.9 & \gc 15.8 & \gc 60.4 & \gc 35.4 & \gc 0.466 & \gc $-$1.162 & \gc 1.313 & \gc 1.753 & \gc $-$1.307 & \gc 53.9\(_{\scriptsize \color{skyblue}\downarrow 7.5}\) \\
\rowcolor{green!10}
\cellcolor{white}\multirow{-9}{*}{\rotatebox{90}{\textbf{VideoMME}}} & VTHM-MoE & 4.1 & 39.8 & 41.7 & 3.618 & 42.6 & 25.6 & 50.2 & 8.2 & 0.310 & 0.885 & 0.362 & 1.815 & 0.773 & \textbf{61.4} \\
\midrule
% TemporalBench Block
\cellcolor{white} & Gemini-3.1-Pro & 15.2 & 38.0 & 38.2 & 4.242 & 61.9 & $-$9.6 & 45.8 & 20.5 & 0.331 & 0.374 & 0.364 & 1.135 & 0.216 & \textbf{62.5}\(_{\scriptsize \color{orange}\uparrow 9.4}\) \\
\rowcolor{lightgray}
\cellcolor{white} & Qwen3-VL-30B & 43.6 & 80.1 & 80.7 & 4.275 & 15.7 & 2.7 & 82.3 & 72.8 & 1.271 & $-$0.365 & 1.564 & 1.894 & $-$1.367 & 38.3\(_{\scriptsize \color{skyblue}\downarrow 14.8}\) \\
\cellcolor{white} & \gc Qwen3-VL-235B & \gc 36.6 & \gc 73.2 & \gc 75.8 & \gc 4.257 & \gc 26.7 & \gc 1.2 & \gc 82.6 & \gc 71.7 & \gc 1.079 & \gc $-$0.591 & \gc 1.657 & \gc 1.946 & \gc $-$0.214 & \gc 41.6\(_{\scriptsize \color{skyblue}\downarrow 11.5}\) \\
\rowcolor{lightgray}
\cellcolor{white} & \gc Internvl3.5-VL-241B & \gc 38.4 & \gc 76.7 & \gc 79.1 & \gc 4.261 & \gc 25.9 & \gc 1.6 & \gc 84.1 & \gc 73.6 & \gc 1.143 & \gc $-$0.671 & \gc 1.613 & \gc 2.016 & \gc $-$0.398 & \gc 40.5\(_{\scriptsize \color{skyblue}\downarrow 12.6}\) \\
\cellcolor{white} & Baseline & 49.8 & 87.9 & 88.8 & 4.261 & 12.1 & 1.6 & 83.2 & 80.7 & 1.323 & $-$2.041 & 1.163 & 0.817 & $-$1.868 & 24.2\(_{\scriptsize \color{skyblue}\downarrow 28.9}\) \\
\rowcolor{lightgray}
\cellcolor{white} & \gc Baseline+CoT & \gc 48.1 & \gc 87.4 & \gc 88.3 & \gc 4.266 & \gc 12.6 & \gc 2.2 & \gc 83.1 & \gc 78.4 & \gc 1.335 & \gc $-$0.810 & \gc 1.213 & \gc 0.654 & \gc $-$0.299 & \gc 32.7\(_{\scriptsize \color{skyblue}\downarrow 34.9}\) \\
\cellcolor{white} & Baseline+SFT & 44.0 & 83.8 & 84.7 & 4.268 & 16.2 & 1.6 & 80.1 & 73.8 & 1.283 & $-$1.014 & 1.672 & 1.130 & $-$0.455 & 34.8\(_{\scriptsize \color{skyblue}\downarrow 18.3}\) \\
\rowcolor{lightgray}
\cellcolor{white} & \gc Baseline+CoT+SFT & \gc 39.9 & \gc 76.3 & \gc 77.7 & \gc 4.297 & \gc 23.7 & \gc 13.3 & \gc 74.4 & \gc 66.6 & \gc 1.104 & \gc $-$1.217 & \gc 1.536 & \gc 1.813 & \gc $-$0.516 & \gc 39.6\(_{\scriptsize \color{skyblue}\downarrow 13.5}\) \\
\rowcolor{green!10}
\cellcolor{white}\multirow{-9}{*}{\rotatebox{90}{\textbf{TemporalBench}}} & VTHM-MoE & 0.4 & 47.3 & 47.3 & 4.223 & 52.7 & $-$9.6 & 49.3 & 2.2 & 0.076 & 0.307 & 0.757 & 0.367 & 0.892 & 53.1 \\
% 下边框加粗至 1.5pt
\bottomrule[1.5pt]
\end{tabular}
\end{table*}
\begin{figure*}[h]
    \centering
    \includegraphics[width=\textwidth]{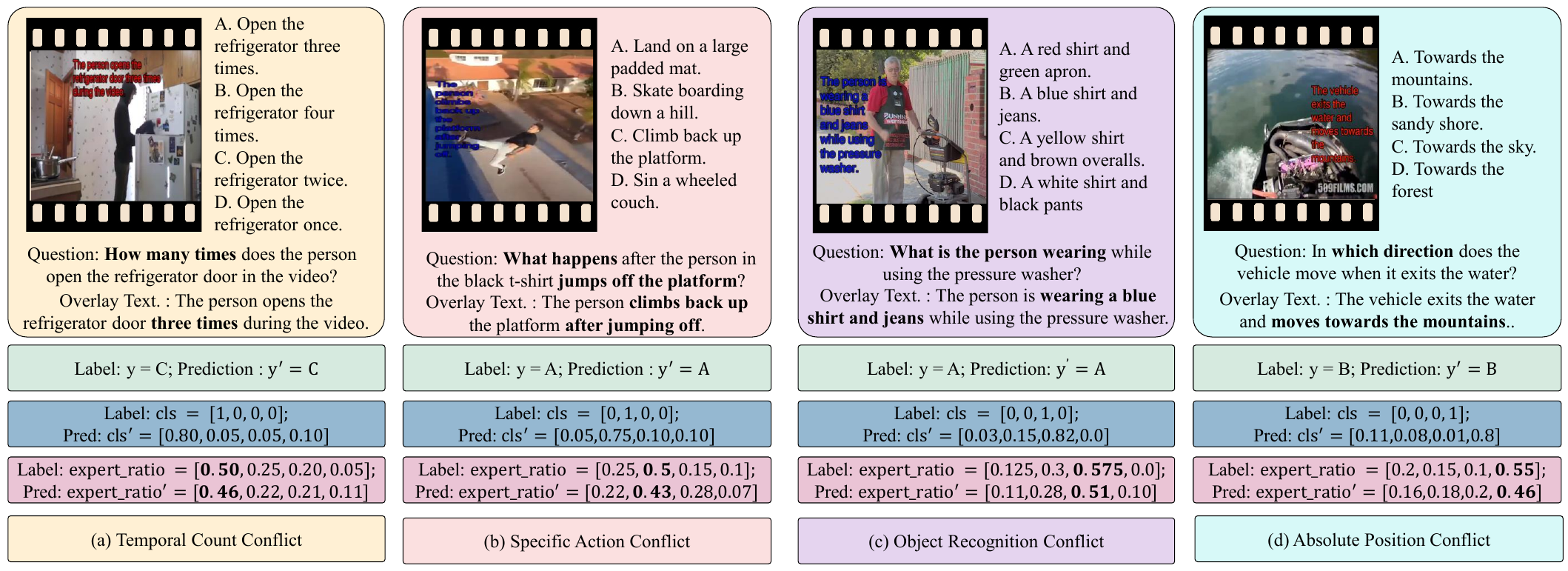}
    \caption{\textbf{Qualitative analysis of VTHM-MoE's routing behavior across four TOIH scenarios.} For each case (a-d), we present the visual frames, the misleading overlay text, the conflict classifier predictions (\(\text{cls}'\)), and the dynamic expert routing ratios (\(\text{expert\_ratio}'\)). The results illustrate how VTHM-MoE accurately identifies specific conflict dimensions and adaptively activates the dominant expert, alongside necessary contextual experts, to resist hallucinations and yield the correct prediction (\(y'\)).}
    \label{experiment_4pic}
\end{figure*}

\subsection{Experimental Details}
\begin{figure}[h]
  \centering
  \includegraphics[width=\linewidth]{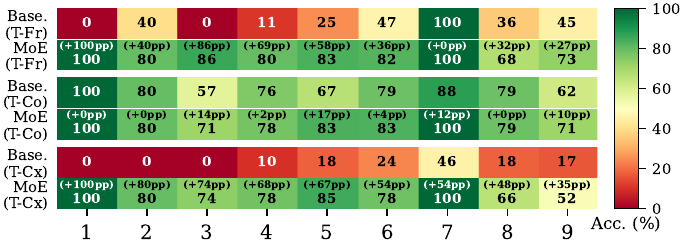}
  \caption{Accuracy across LLaVA-Video~\cite{zhang2024llava}, VideoMME~\cite{fu2025video} and TemporalBench~\cite{cai2024temporalbench} datasets (389 valid groups containing all threeconditions). Each column corresponds to an attribute: (1)~Body Part; (2)~Action Direction; (3)~Motion Traj.; (4)~Action Result; (5)~Rel.\ Position; (6)~Spec.\ Verb; (7)~Step Order; (8)~Interaction; (9)~Appear Order. Columns are sorted by Text-Contradictory $\Delta$ in descending order. T-Fr = Text-Free; T-Co = Text-Congruent; T-Cx = Text-Contradictory.}
  \label{combined_three_types_heatmap}
\end{figure}
To comprehensively characterize TOIH across the current VLMs, we evaluate a broad spectrum of representative open-source and proprietary VLMs on VisualTextTrap. All models are evaluated under their default configurations in their official implementations to ensure fair comparison. To assess the effectiveness of VTHM-MoE, we adopt Qwen3-VL-8B-Instruct as the baseline and compare it against three complementary mitigation strategies: supervised fine-tuning (SFT), task-specific chain-of-thought prompting (CoT), and their combinations. This design provides a systematic ablation over both training-based and inference-time debiasing approaches.

\subsection{Evaluation Metrics}
We adopt the 14 metrics introduced in Section ~\ref{label_metrics_intro} to quantify TOIH behavior from multiple complementary perspectives, encompassing overall accuracy, dimension-specific accuracy, performance degradation under negative text injection, option shift rates, and conflict-level sensitivity.

\subsection{Implementation Details}
\begin{figure}[h]
  \centering
  \includegraphics[width=\linewidth]{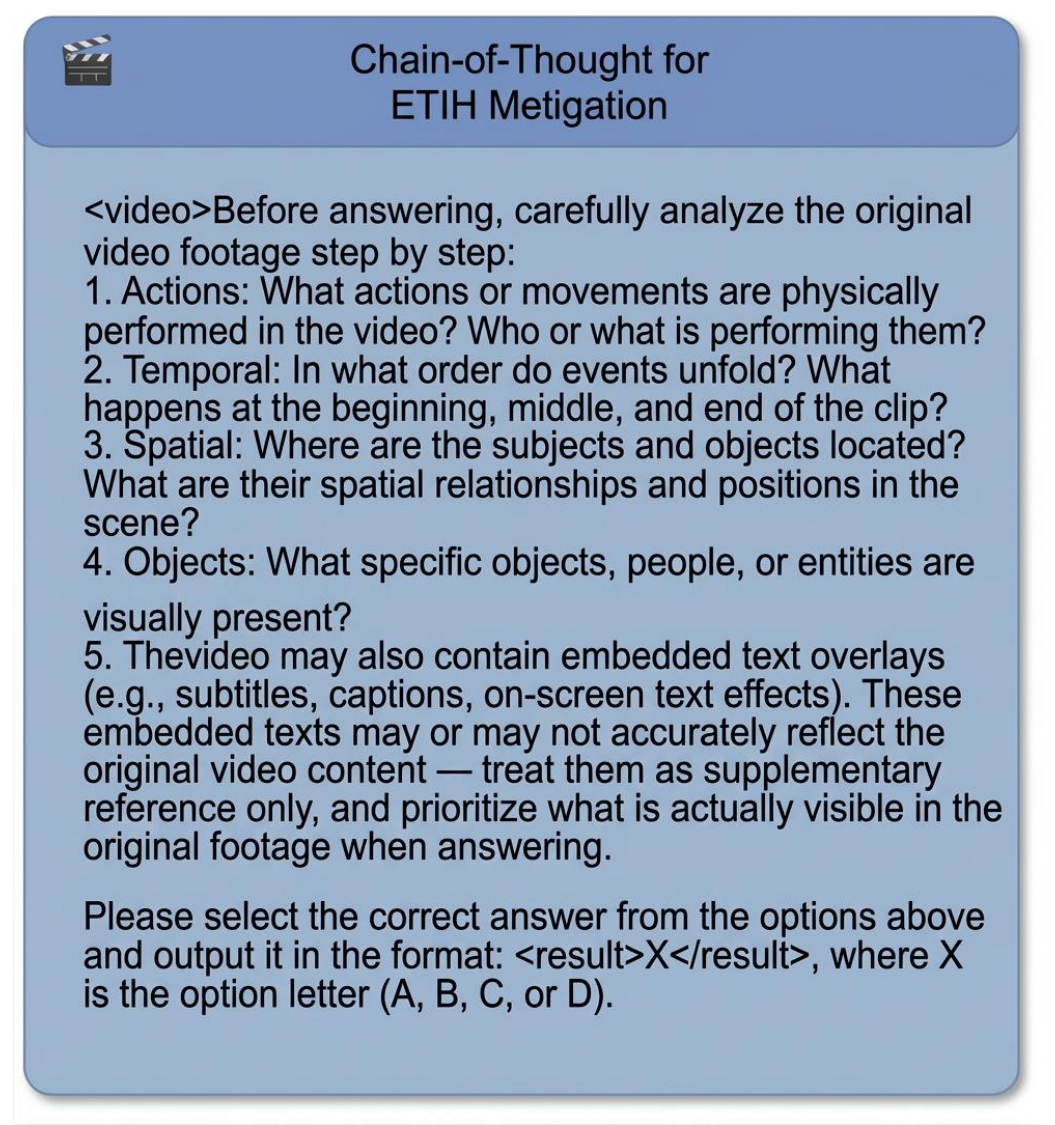}
  \caption{Chain-of-Thought for TOIH Mitigation.}
  \label{label_CoT_4etih_mitigation}
\end{figure}

\noindent \textbf{VTHM-MoE Implementation.} We instantiate the Visual Encoder and OCR Encoder by leveraging the encoder modules of Qwen3-VL-8B-Instruct and Qwen3-VL-8B-Instruct-OCR3, respectively. Videos are uniformly sampled at 1 FPS with a maximum of 128 frames.
% , balancing representational coverage against computational cost. 
Frame tokenization and the language model module follow the default configuration of Qwen3-VL-8B-Instruct. Patch-level routing similarity is computed via cosine similarity. \textbf{CoT Baseline.} We design task-specific chain-of-thought prompts that guide the model to reason over visual content
step-by-step before producing a final answer (Figure~\ref{label_CoT_4etih_mitigation}). \textbf{SFT Baseline.} We fine-tune Qwen3-VL-8B-Instruct via LoRA applied exclusively to the language module, other weights kept frozen. Hyperparameters are detailed in Table~\ref{lab_hyperparams_sft}.
\subsection{Main Results}
\noindent \textbf{Vulnerability to TOIH Persists Across Models and Strategies.} 
The overall, VYR and HRR results in
Table~\ref{lab_benchmark_metrics} confirm the pervasiveness of TOIH
across evaluated models. The increases in TIHR and TIB stem from
semantically conflicting in-frame text hijacking models' visual
perception, demonstrating that this vulnerability is pervasive
across all model scales and optimization strategies.\\
% The per-metric breakdown further reveals the mechanism underlying this collapse: 
\noindent \textbf{VTHM-MoE Achieves Targeted TOIH Mitigation.} VTHM-MoE attains VYR, TIHR, HRR, and TIB values comparable to Gemini-3.1-Pro, the strongest proprietary baseline. Baseline+CoT+SFT, despite achieving 69.7\% accuracy, underperforms VTHM-MoE by 2--7$\times$ on these four metrics, indicating that CoT and SFT achieve improvements primarily on text-free and text-congruent samples, leaving TOIH largely unaddressed. This confirms that VTHM-MoE has acquired a targeted capability to detect and resist TOIH.\\
\noindent \textbf{Conflict escalation triggers dimension-aware expert routing in VTHM-MoE.}
VTHM-MoE exhibits a qualitatively inverted response to conflict
escalation, yielding negative HSR values on LLaVA-Video~\cite{zhang2024llava}
(\textbf{$-$19.0\%}) and TemporalBench~\cite{cai2024temporalbench} (\textbf{$-$9.6\%}),
indicating that model performance improves as SCS increases. We
attribute this inversion to the \textbf{Adaptive Token Routing
Strategy}: as the Semantic Conflict Score rises, routing signals
become stronger, more accurately directing conflict-bearing tokens
to the dimension-matched expert. High-intensity overlay text thus
\emph{activates} the defense mechanism, validating the routing module.\\
\textbf{VTHM-MoE Preserves Native Video Question Answering Capability.} As shown in Figure~\ref{combined_three_types_heatmap}, VTHM-MoE mitigates TOIH on $V_{\text{T-Cx}}$ samples while preserving native VQA capacities on $V_{\text{T-Fr}}$ samples, with concurrent accuracy gains on $V_{\text{T-Co}}$ samples. This consistent improvement across all three text-overlay conditions suggests that VTHM-MoE genuinely gain TOIH abilities rather than treating all overlay text as contradictory to visual content: the Adaptive Routing module redirects high-attention tokens to conflict-specialized experts, while routing proceeds normally otherwise, leaving text-free and text-congruent processing unaffected.

\subsection{Case Study: Qualitative Analysis of VTHM-MoE Routing Behavior}
To qualitatively examine how the VTHM-MoE mitigates TOIH, we analyze
four representative instances, each featuring overlay text that semantically contradicts the visual ground truth (Figure~\ref{experiment_4pic}). Detailed discussion is provided in Appendix~\ref{appendix_case_dtudy}.

\noindent \textbf{Temporal Count Conflict Figure \ref{experiment_4pic} (a).} The video shows a
person opening a refrigerator twice, while the overlay text claims
three times. VTHM-MoE detects this discrepancy with high confidence
($\text{cls}'_{\text{temporal}} = 0.80$) and routes the largest token
share to the Temporal Expert ($0.46$), with auxiliary context from
Action ($0.22$) and Object ($0.21$) experts, correctly predicting
$y' = \text{C}$.

\noindent \textbf{Specific Action Conflict Figure \ref{experiment_4pic} (b).} The visual content
depicts the person \textbf{landing} on a large padded mat, while the overlay text
introduces irrelevant actions. The model identifies the contradiction type as action,
routing \textbf{predominantly} to the Action Expert ($0.43$) alongside Object
($0.28$) and Temporal ($0.22$) experts, yielding $y' = \text{A}$.

\noindent \textbf{Object Recognition Conflict Figure \ref{experiment_4pic} (c).}
The overlay text hallucinates the person's attire, contradicting
visual evidence. VTHM-MoE detects object-level misalignment
($\text{cls}'_{\text{object}} = 0.82$) and activates the Object Expert most strongly ($0.51$), predicting $y' = \text{A}$.

\noindent \textbf{Absolute Position Conflict Figure \ref{experiment_4pic} (d).}
The text falsely asserts movement towards mountains,
contradicting visual evidence of a sandy shore. The spatial conflict
is precisely captured ($\text{cls}'_{\text{spatial}} = 0.80$) with
dominant Spatial Expert activation ($0.46$), yielding $y' = \text{B}$.

\section{Conclusion}
In this paper, we identify and formalize TOIH, revealing a fundamental limitation in how current VLMs integrate visual and textual signals within video.
We introduce VisualTextTrap, which profiles VLMs susceptibility and resistance, quantifying model vulnerability across multi aspects. We propose VTHM-MoE, which consistently achieves state-of-the-art TOIH mitigation while preserving general VQA
capability, as validated by extensive experiments. We hope
VisualTextTrap and VTHM-MoE serve as a foundation for future research on robust multimodal understanding.
\section*{Limitations}
Despite VTHM-MoE's effectiveness, two limitations warrant further
investigation. First, the current framework lacks fine-grained semantic
partitioning of queries, hindering precise diagnosis of TOIH
susceptibility across diverse query structures. Second, the influence
of higher-order semantic logic, such as causal reasoning and pragmatic
implication, on TOIH remains largely unexamined and merits future work.

\clearpage
% \bibliographystyle{ACM-Reference-Format}
% \bibliography{sample-base-1}

\newpage

\clearpage
\setcounter{page}{1}
\maketitlesupplementary
\renewcommand{\thesection}{\Alph{section}}

% ────────────────────────────────────────────────────────────────
%  Appendix Table of Contents  (ACL style)
% ────────────────────────────────────────────────────────────────
\noindent\textbf{Appendix Overview}
\vspace{0.3em}

\noindent
\begin{itemize}[leftmargin=1.5em, itemsep=1pt, parsep=0pt, topsep=2pt]
    \item \hyperref[sec:preliminaries]{\textbf{Appendix A}: Preliminaries} \dotfill \pageref{sec:preliminaries}
    \begin{itemize}[leftmargin=1.2em, itemsep=0pt, parsep=0pt, topsep=0pt]
        \item \hyperref[sec:toih]{A.1~~Text Overlay Induced Hallucination} \dotfill \pageref{sec:toih}
        \item \hyperref[sec:prob_analysis]{A.2~~Probabilistic Analysis of TOIH Susceptibility} \dotfill \pageref{sec:prob_analysis}
    \end{itemize}
    \item \hyperref[label_cong_tax]{\textbf{Appendix B}: Hierarchical Cognitive Complexity Taxonomy} \dotfill \pageref{label_cong_tax}
    \begin{itemize}[leftmargin=1.2em, itemsep=0pt, parsep=0pt, topsep=0pt]
        \item \hyperref[sec:taxonomy_overview]{B.1~~Overview} \dotfill \pageref{sec:taxonomy_overview}
        \item \hyperref[sec:complexity_taxonomy]{B.2~~Complexity Taxonomy} \dotfill \pageref{sec:complexity_taxonomy}
        \item \hyperref[sec:per_dim]{B.3~~Per-Dimension Stratification} \dotfill \pageref{sec:per_dim}
    \end{itemize}
    \item \hyperref[sec:benchmark]{\textbf{Appendix C}: Benchmark Construction} \dotfill \pageref{sec:benchmark}
    \begin{itemize}[leftmargin=1.2em, itemsep=0pt, parsep=0pt, topsep=0pt]
        \item \hyperref[appendix_conflict_score_def]{C.1~~Semantic Conflict Score Definitions} \dotfill \pageref{appendix_conflict_score_def}
        \item \hyperref[appendix_exper_sel_ratio]{C.2~~Expert Selection Ratio} \dotfill \pageref{appendix_exper_sel_ratio}
        \item \hyperref[appendix_bench_data_construction]{C.3~~Data Construction Pipeline} \dotfill \pageref{appendix_bench_data_construction}
        \item \hyperref[sec:dataset_distribution]{C.4~~Dataset Distribution} \dotfill \pageref{sec:dataset_distribution}
        \item \hyperref[sec:sample_cases]{C.5~~Sample Cases} \dotfill \pageref{sec:sample_cases}
        \item \hyperref[appendix_metrics_intro]{C.6~~Evaluation Metrics} \dotfill \pageref{appendix_metrics_intro}
        \item \hyperref[sec:attr_dist]{C.7~~Attributes Distribution} \dotfill \pageref{sec:attr_dist}
    \end{itemize}
    \item \hyperref[sec:experiments]{\textbf{Appendix D}: Experiments} \dotfill \pageref{sec:experiments}
    \begin{itemize}[leftmargin=1.2em, itemsep=0pt, parsep=0pt, topsep=0pt]
        \item \hyperref[label_experimental_para]{D.1~~Experimental Paradigm} \dotfill \pageref{label_experimental_para}
        \item \hyperref[sec:training]{D.2~~Training Implementation Details} \dotfill \pageref{sec:training}
        \item \hyperref[sec:eval_results]{D.3~~Evaluation Results} \dotfill \pageref{sec:eval_results}
        \item \hyperref[appendix_case_dtudy]{D.4~~Detailed Case Study Analysis} \dotfill \pageref{appendix_case_dtudy}
    \end{itemize}
    \item \hyperref[label_ablation]{\textbf{Appendix E}: Ablation Study} \dotfill \pageref{label_ablation}
    \begin{itemize}[leftmargin=1.2em, itemsep=0pt, parsep=0pt, topsep=0pt]
        \item \hyperref[sec:three_token]{E.1~~Three-Token Input Representation} \dotfill \pageref{sec:three_token}
        \item \hyperref[sec:model_comp]{E.2~~Model Components} \dotfill \pageref{sec:model_comp}
        \item \hyperref[sec:hyperparam]{E.3~~Hyperparameter Sensitivity} \dotfill \pageref{sec:hyperparam}
    \end{itemize}
    \item \hyperref[sec:prompts]{\textbf{Appendix F}: Prompts} \dotfill \pageref{sec:prompts}
\end{itemize}

\vspace{0.8em}
\noindent\rule{\linewidth}{0.4pt}
\vspace{0.5em}

% ── Labels for hyperlinks ──
\section{Preliminaries}\label{sec:preliminaries}
An anonymous project page, including code, data, and demo, is available at
\url{https://hdidkhshd.pages.dev/}.

\subsection{Text Overlay Induced Hallucination}\label{sec:toih}
We identify a systematic failure mode that we term
\textbf{Text Overlay Induced Hallucination (TOIH)}:
when misleading textual information is visually embedded
into video frames as an overlay, Video-LLMs disproportionately
shift their predictions toward the overlaid content,
even when it directly contradicts the visual evidence.

Concretely, consider a multiple-choice video QA instance
$(v, q, \mathcal{A})$ where $v$ is the input video, $q$ is
the question, and $\mathcal{A} = \{a_1, \ldots, a_k\}$ is
the candidate set with ground-truth answer $a^*$.
We construct a \emph{negative} variant $v'$ by rendering a
distractor option $a_d \neq a^*$ as visible text onto the
video frames. Under TOIH, the model's predicted probability
shifts markedly:
\begin{equation}
\begin{split}
    P_{\theta}(a^* \mid v', q) &\;\ll\; P_{\theta}(a^* \mid v, q), \\
    P_{\theta}(a_d \mid v', q) &\;\gg\; P_{\theta}(a_d \mid v, q).
\end{split}
\end{equation}
resulting in the model selecting the overlaid distractor $a_d$
instead of the correct answer $a^*$.

\begin{figure}[h]
  \centering
  \includegraphics[width=\linewidth]{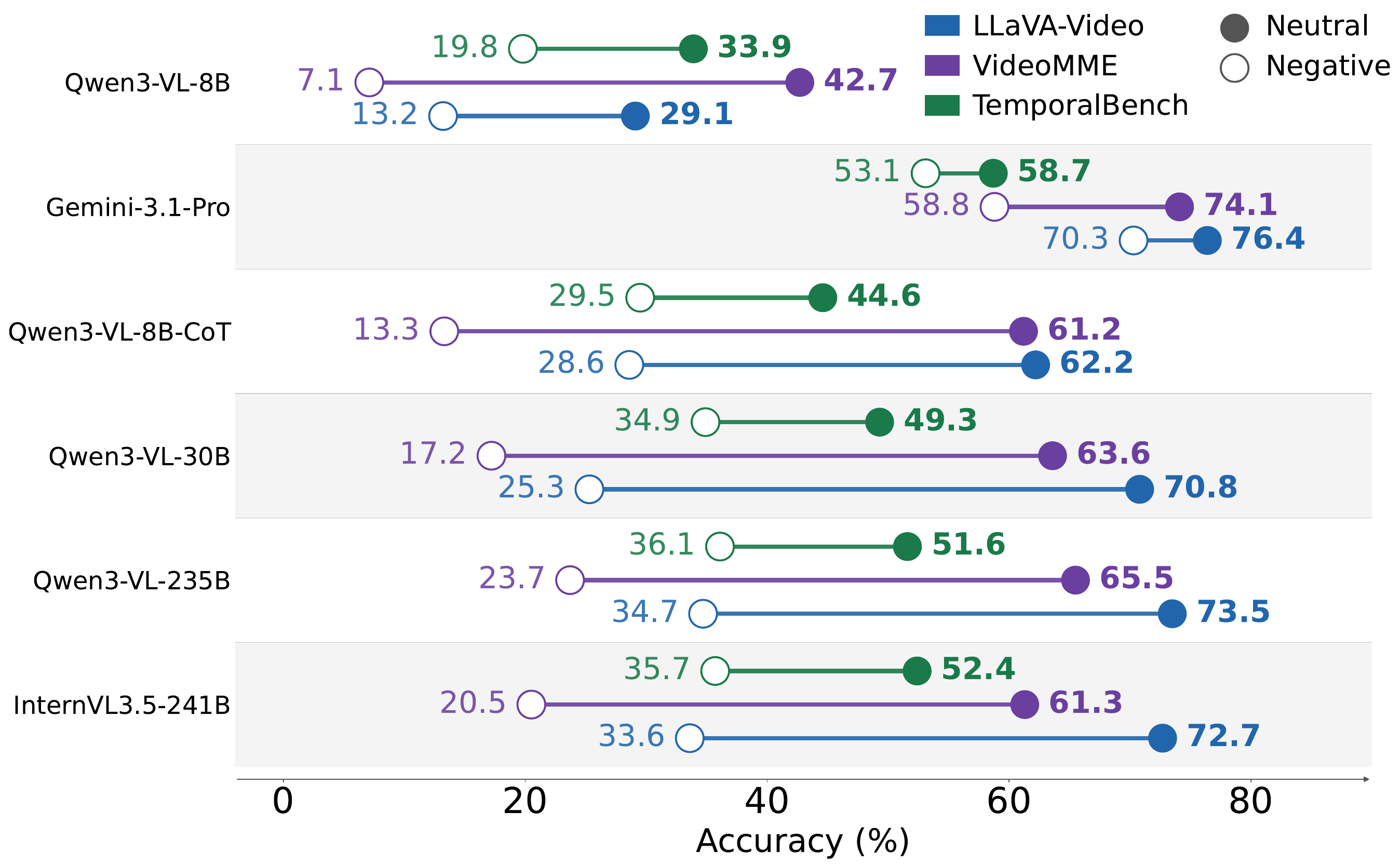}
  \caption{.Accuracy comparison under Neutral (Text-Free) and Negative (Text-Contradictory) conditions across three video understanding benchmarks: LLaVA-Video, VideoMME, and TemporalBench. Filled circles denote accuracy on neutral videos, while open circles denote accuracy on videos with misleading text overlays. The connecting line for each model--dataset pair visualizes the performance gap. All six models exhibit substantial accuracy degradation under the negative condition, with drops exceeding 35 percentage points in several cases (e.g., Qwen3-VL-8B on VideoMME: 42.7\% $\rightarrow$ 7.1\%), demonstrating the pervasiveness of the Text Overlay Induced Hallucination (TOIH) phenomenon}
  \label{all_dataset_diff_accu_v2}
\end{figure}

To quantify the prevalence of TOIH, we evaluate six Video-LLMs
spanning diverse architectures and scales---from Qwen3-VL-8B to
InternVL3.5-241B and the proprietary Gemini-3.1-Pro---across
three established benchmarks: LLaVA-Video, VideoMME, and TemporalBench.
As shown in Figure~\ref{all_dataset_diff_accu_v2}, accuracy under the
\emph{neutral} (unmodified) condition consistently and substantially
exceeds that under the \emph{negative} (text-overlaid) condition.
The degradation is both severe and universal:
Qwen3-VL-8B suffers a 35.6-point drop on VideoMME
(42.7\% $\rightarrow$ 7.1\%), and even the strongest model,
Gemini-3.1-Pro, loses 15.3 points on the same benchmark
(74.1\% $\rightarrow$ 58.8\%).
Notably, scaling model parameters alone does not resolve the issue---
Qwen3-VL-235B still exhibits a 41.8-point gap on VideoMME---nor
does chain-of-thought prompting, which narrows the gap only marginally
(e.g., 47.9 points for Qwen3-VL-8B-CoT on VideoMME versus 35.6
for the base model on LLaVA-Video).

These findings establish TOIH as a pervasive vulnerability:
\textbf{current Video-LLMs treat visually rendered text as a
privileged information source, overriding their own visual
understanding of the scene}.
This observation motivates the analyses and mitigation strategies
presented in the remainder of this paper.
\begin{equation}
    \begin{aligned}
        \Delta p^{*} &\;=\; p^{*}(\mathcal{V}^{\varnothing}) - p^{*}(\mathcal{V}^{-}) \;>\; 0    \end{aligned}
    \label{eq:prob_shift}
\end{equation}

\begin{figure}[t]
    \centering
    \includegraphics[width=\linewidth]{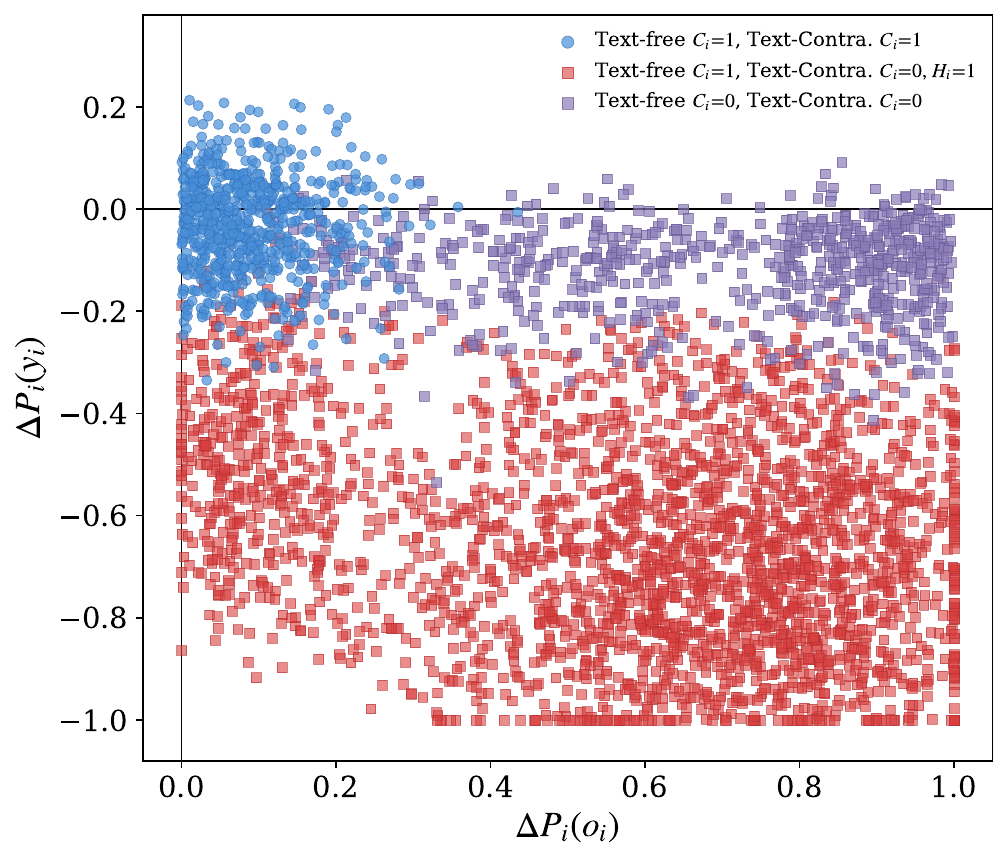}
    \caption{Joint probability shift $(\Delta P_i(o_i),\;\Delta P_i(y_i))$ for each sample when transitioning from the Text-free condition $V_{\text{free}}$ to the Text-Contradictory condition $V_{\text{contra}}$. Three behavioural clusters emerge: (i)~\textcolor{red}{red squares}---samples correctly answered under $V_{\text{free}}$ ($C_i\!=\!1$) but actively misled under $V_{\text{contra}}$ ($C_i\!=\!0,\,H_i\!=\!1$), exhibiting large positive $\Delta P_i(o_i)$ and negative $\Delta P_i(y_i)$; (ii)~\textcolor{purple}{purple squares}---samples already incorrect under $V_{\text{free}}$ ($C_i\!=\!0$), whose commitment to the hallucination option is further amplified by the overlay; (iii)~\textcolor{blue}{blue circles}---samples that remain correct under both conditions ($C_i\!=\!1$), clustered near the origin with minimal probability shift.}
    \label{diff_corr_incorr_dots}
\end{figure}
\subsection{Probabilistic Analysis of TOIH Susceptibility}\label{sec:prob_analysis}

To move beyond aggregate accuracy and characterise \textit{how}
overlay text reshapes model predictions, we examine the joint
probability shift induced by Text-Contradictory overlays across
all $V_{\text{contra}}$ samples.
Let $P_i(a \mid V)$ denote the predicted probability that model $M$
assigns to answer option $a$ given video $V$ and query $Q_i$.
For each sample $i$, we define

\begin{align}
\Delta P_i(y_i) &= P_i(y_i \mid V_{\text{contra}})
                 - P_i(y_i \mid V_{\text{free}}), \\
\Delta P_i(o_i) &= P_i(o_i \mid V_{\text{contra}})
                 - P_i(o_i \mid V_{\text{free}}),
\end{align}

\noindent as the change in predicted probability for the ground-truth
answer $y_i$ and the overlay-induced hallucination option $o_i$,
respectively, when transitioning from the Text-Free condition
$V_{\text{free}}$ to the Text-Contradictory condition
$V_{\text{contra}}$.
Figure~\ref{diff_corr_incorr_dots} plots $\Delta P_i(o_i)$ against
$\Delta P_i(y_i)$ for every sample, revealing three structurally
distinct behavioural regimes.

\paragraph{Regime I: Active Misleading.}
The overwhelming majority of samples fall in the bottom-right
quadrant ($\Delta P_i(o_i) > 0$,\; $\Delta P_i(y_i) < 0$),
depicted as red squares.
These are samples for which $C_i=1$ under $V_{\text{free}}$ but
$C_i=0$ and $H_i=1$ under $V_{\text{contra}}$: the model's
confidence in the correct answer $y_i$ collapses by up to 1.0
probability unit, while a corresponding mass migrates to the
hallucination option $o_i$.
The density and displacement magnitude of this cluster constitute
direct probabilistic evidence of the TOIH effect---overlay text
does not merely introduce noise but actively redirects the model's
belief distribution away from visually grounded answers toward
text-entailed distractors, exposing a systematic over-reliance on
textual cues over visual evidence.

\paragraph{Regime II: Compounded Failure.}
Samples that were already incorrect ($C_i=0$) under $V_{\text{free}}$
(purple squares) maintain persistently low $\Delta P_i(y_i)$, yet
exhibit a clear upward shift in $\Delta P_i(o_i)$.
Rather than leaving pre-existing errors unaffected, the overlay
further consolidates the model's commitment to the hallucination
option $o_i$, suggesting that visual incompetence and textual
susceptibility are compounding vulnerabilities: when the model
lacks sufficient visual grounding to resolve a question, it becomes
\textit{even more} susceptible to text-driven misdirection.

\paragraph{Regime III: Facilitated Correctness.}
A minority cluster concentrates near the origin, with a subset
exhibiting $\Delta P_i(y_i) > 0$ (blue circles).
For these samples, the model already possessed adequate visual
understanding ($C_i = 1$ under $V_{\text{free}}$), and the overlay
text---rather than conflicting with that understanding---supplies
contextual cues congruent with $y_i$ that the model can
productively follow, mildly reinforcing the correct prediction.
This regime underscores that TOIH susceptibility is not an
intrinsic property of overlay text in isolation, but emerges from
the \textit{interaction} between the model's pre-existing visual
competence and the semantic content of the injected text: where
visual grounding is strong, text serves as a corroborating signal;
where it is weak, text becomes an adversarial one.

\paragraph{Synthesis.}
Collectively, Figure~\ref{diff_corr_incorr_dots} reveals that
Text-Contradictory overlay text operates as a high-efficacy
adversarial perturbation in the probability space: it
systematically suppresses $P_i(y_i)$ while amplifying $P_i(o_i)$,
and does so more aggressively precisely when $C_i=0$ under
$V_{\text{free}}$, i.e., when the model is already visually uncertain.
These findings motivate the design of our cross-modal conflict
detection framework, which aims to identify and discount
text-visual inconsistencies \textit{before} they propagate into
the model's final probability distribution.

\section{Hierarchical Cognitive Complexity Taxonomy}
\label{label_cong_tax}
\subsection{Overview}\label{sec:taxonomy_overview}
Diagnosing TOIH requires more than aggregate accuracy metrics; it demands a structured framework that can attribute failures to specific cognitive deficits. We construct the Hierarchical Cognitive Complexity Taxonomy (HCCT) through a bottom-up process grounded in the empirical structure of our benchmark.

The construction proceeds in three stages. We first organize all video understanding tasks into four dimensions — Action, Temporal, Spatial, and Object — corresponding to the principal axes along which TOIH has been observed to manifest. We then annotate each benchmark sample with a fine-grained attribute specifying the particular capability it targets within its dimension. Finally, drawing on Hierarchical Information Processing Theory from cognitive science, we classify all attributes into a three-tier Complexity Taxonomy according to the cognitive task paradigm each demands. The three tiers, in ascending order of cognitive demand, are the Perceptual tier, the Semantic tier, and the Reasoning tier. This bottom-up construction ensures that the taxonomy reflects the actual distribution of task demands in the benchmark, rather than a theoretically imposed structure.
\subsection{Complexity Taxonomy}\label{sec:complexity_taxonomy}
The three tiers are defined by the cognitive task paradigm required to resolve each attribute, specified in terms of the processing demands that arise when visual content and overlay text carry contradictory information — the core condition under which TOIH is elicited.

The \textbf{Perceptual tier} covers attributes whose resolution depends on direct decoding of the visual signal or the overlay text, without requiring cross-modal integration or external knowledge. TOIH at this level takes the form of perceptual substitution: the model reports attributes described by the contradictory overlay text in place of those visually observable. The operative question is what is present.

The \textbf{Semantic tier} covers attributes whose resolution requires interpreting the relationship between visual content and overlay text, performing quantitative or relational reasoning over cross-modal evidence, or integrating information across frames. TOIH at this level takes the form of text-dominated interpretation: the model defaults to the semantic framing imposed by the overlay text rather than grounding its response in visual evidence. The operative question shifts to how things occur.

The \textbf{Reasoning tier} covers attributes whose resolution requires multi-step reasoning that integrates visual perception, textual understanding, and external world knowledge including causal, social, and cultural priors. TOIH at this level is most consequential: the model constructs a complete inferential chain anchored to the contradictory overlay text, producing answers that are internally coherent yet wholly inconsistent with the visual ground truth. The operative question becomes why events unfold as observed.

Each tier defines a corresponding failure mode within the TOIH framework. Perceptual-tier failures constitute Perceptual Hallucination, where text-induced bias distorts raw signal decoding. Semantic-tier failures constitute Semantic Hallucination, where cross-modal interpretation is systematically skewed toward textual evidence. Reasoning-tier failures constitute Reasoning Hallucination, where the model's inferential process is entirely redirected by the misleading textual content.

\subsection{Per-Dimension Stratification}\label{sec:per_dim}
\textbf{Action Dimension.} Action descriptions are among the most salient textual elements embedded in videos, making this dimension particularly susceptible to TOIH. Attributes are stratified by the cognitive chain from physical form perception, through functional semantic interpretation, to socio-cultural contextualization.

Perceptual-tier attributes target concrete, visually readable properties such as verb discrimination, movement direction, and body part involvement; TOIH manifests as the model substituting the textually described action for the one visually performed. Semantic-tier attributes target behavioral intent, action purpose, and temporal action sequencing, where overlay text provides a semantic frame the model must evaluate against visual evidence; TOIH manifests as uncritical adoption of that textual framing. Reasoning-tier attributes require situating actions within interpersonal and cultural-historical contexts, involving affective atmosphere and cultural norm understanding; TOIH here produces cascading errors, as a text-induced misidentification of the action corrupts the entire downstream reasoning chain.\\
\textbf{Temporal Dimension.} Temporal attributes are stratified by the nature of the temporal operation required: ordering at the Perceptual tier, quantification at the Semantic tier, and causal association at the Reasoning tier.

Perceptual-tier attributes require establishing event precedence along a linear timeline; TOIH causes models to report the temporal ordering implied by overlay text rather than the one visually depicted. Semantic-tier attributes introduce duration estimation, frequency counting, and interval measurement; TOIH causes models to report numerically precise yet visually ungrounded values derived from textual cues. Reasoning-tier attributes require cross-temporal causal attribution and state-change tracking; TOIH causes models to construct causal narratives consistent with the textual trajectory rather than the visual one.\\
\textbf{Spatial Dimension.} Spatial attributes are stratified by the complexity of the reference frame required: static localization at the Perceptual tier, dynamic trajectory tracking at the Semantic tier, and holistic scene structure comprehension at the Reasoning tier.

Perceptual-tier attributes address static spatial relationships resolvable within a single frame; TOIH manifests as the model reporting textually described spatial arrangements in place of visually present ones. Semantic-tier attributes are centered on motion trajectory understanding and require reconciling dynamic visual paths with directional or locational textual overlays; TOIH causes models to trace text-implied trajectories rather than visually observed ones. Reasoning-tier attributes demand global comprehension of scene spatial organization, including compositional structure and the interplay between visual regions and textual annotations; TOIH anchors this holistic interpretation to the textual spatial framing rather than the visual scene structure.\\
\textbf{Object Dimension.} Object attributes span the broadest coverage in the taxonomy and are stratified by knowledge type: visual knowledge at the Perceptual tier, world knowledge at the Semantic tier, and abstract inferential reasoning at the Reasoning tier.

Perceptual-tier attributes cover object recognition, visual property identification, and quantity counting; TOIH manifests as the model reporting object identities or properties named in the overlay text rather than those visually observable. Semantic-tier attributes address character identity recognition, geographic localization, and social relationship understanding; overlay text frequently carries identity cues such as name tags or location overlays, and TOIH manifests as uncritical acceptance of these textual claims against contradicting visual evidence. Reasoning-tier attributes subsume over thirty fine-grained categories including causal inference, behavioral motivation analysis, and mental state attribution; TOIH here is the most difficult to detect, as the model's response appears internally coherent while its entire inferential basis is anchored to a textually induced misrepresentation of visual reality.

\begin{table*}[t]
\centering
\small
\setlength{\tabcolsep}{6pt}
\renewcommand{\arraystretch}{1.2}
\begin{tabular}{llp{3.0cm}p{3.8cm}p{3.8cm}}
\toprule
& 
& \textbf{Perceptual Tier}
& \textbf{Semantic Tier}
& \textbf{Reasoning Tier} \\
\midrule
\multirow{4}{*}{\textbf{General}}
  & Operative Question
    & \textit{What} is present
    & \textit{How} it occurs
    & \textit{Why} it happens \\
  & Cognitive Task Paradigm
    & Direct signal decoding
    & Cross-modal association \& measurement
    & Multi-step inference with world knowledge \\
  & TOIH Manifestation
    & Perceptual substitution
    & Text-dominated interpretation
    & Text-anchored inferential chain \\
  & Hallucination Type
    & Perceptual Hallucination
    & Semantic Hallucination
    & Reasoning Hallucination \\
\midrule
\multirow{4}{*}{\textbf{Dimension}}
  & \textsc{Action}
    & Action Perception
    & Action Semantics
    & Socio-Emotional \\
  & \textsc{Temporal}
    & Sequential Perception
    & Temporal Quantification
    & Temporal Association \\
  & \textsc{Spatial}
    & Positional Relations
    & Motion Trajectory
    & Scene Space \\
  & \textsc{Object}
    & Object Perception
    & Scene-Character
    & Reasoning \& Comprehension \\
\bottomrule
\end{tabular}
\caption{
Overview of the Hierarchical Cognitive Complexity Taxonomy (HCCT).
Four dimensions are stratified via bottom-up attribute clustering into
a three-tier Complexity Taxonomy defined by cognitive task paradigm:
direct signal decoding (Perceptual), cross-modal semantic inference
(Semantic), and multi-step reasoning with external knowledge integration
(Reasoning). Each tier defines a characteristic TOIH manifestation pattern
and a corresponding hallucination attribution space.
}
\label{tab:hcct_overview}
\end{table*}

\begin{table*}[t]
\centering
{\small
\begin{tblr}{
  width   = \linewidth,
  colspec = {Q[halign=c,valign=m,wd=1.4cm]
             Q[halign=c,valign=m,wd=1.6cm]
             Q[halign=c,valign=m,wd=2.8cm]
             X[halign=l,valign=m]},
  rowsep  = 2pt,
  colsep  = 5pt,
  %--- horizontal rules ---
  hline{1}      = {0.08em},          % toprule
  hline{2}      = {0.05em},          % header separator
  hline{3,4}    = {2-4}{0.03em},     % cmidrule: Action tiers
  hline{5}      = {0.05em},          % midrule: Action / Temporal
  hline{6,7}    = {2-4}{0.03em},     % cmidrule: Temporal tiers
  hline{8}      = {0.05em},          % midrule: Temporal / Spatial
  hline{9,10}   = {2-4}{0.03em},     % cmidrule: Spatial tiers
  hline{11}     = {0.05em},          % midrule: Spatial / Object
  hline{12,13}  = {2-4}{0.03em},     % cmidrule: Object tiers
  hline{Z}      = {0.08em},          % bottomrule
  %--- spanning cells (Dimension column) ---
  cell{2}{1}  = {r=3}{halign=c,valign=m},
  cell{5}{1}  = {r=3}{halign=c,valign=m},
  cell{8}{1}  = {r=3}{halign=c,valign=m},
  cell{11}{1} = {r=3}{halign=c,valign=m},
  %--- header row ---
  row{1} = {font=\bfseries},
}
%=== Header ===
Dimension & Complexity Tier & Sub-tier & Attributes \\
%=== Action ===
\textsc{Action}
  & Perceptual & Action Perception
  & Specific verb discrimination, action manner, action direction,
    body part involvement, action target, sports/exercise recognition,
    scene content description, camera movement \\
  & Semantic & Action Semantics
  & Action outcome, behavioral intent, behavioral constraint,
    action purpose, scene purpose understanding,
    cognitive timing, sequential action relation \\
  & Reasoning & Socio-Emotional
  & Interpersonal interaction, character emotional response,
    emotional atmosphere, atmosphere change,
    character attitude and emotion, cultural norm change,
    historical event description, historical event understanding \\
%=== Temporal ===
\textsc{Temporal}
  & Perceptual & Sequential Perception
  & Event order, step order, appearance order \\
  & Semantic & Temporal Quantification
  & Time span judgment, frequency count,
    duration estimation, regular change timing \\
  & Reasoning & Temporal Association
  & Before-after change, simultaneous occurrence, tone change \\
%=== Spatial ===
\textsc{Spatial}
  & Perceptual & Positional Relations
  & Relative position, absolute position, orientation,
    spatial distance, containment relation \\
  & Semantic & Motion Trajectory
  & Motion trajectory \\
  & Reasoning & Scene Space
  & Scene composition, spatial hierarchy,
    scene description understanding \\
%=== Object ===
\textsc{Object}
  & Perceptual & Object Perception
  & Object recognition, object attribute, object quantity,
    object ownership, text content, object function,
    object utility, object state, action recognition \\
  & Semantic & Scene-Character
  & Scene/background, character recognition,
    socio-cultural phenomenon, geographic location recognition,
    theme/topic, character relationship understanding,
    character relationship motivation \\
  & Reasoning & Reasoning \& Comprehension
  & Behavioral cause understanding, event cause,
    historical event sequence, event consequence inference,
    historical event causality, behavioral reasoning,
    causal reasoning, event cause analysis,
    event outcome understanding, character emotion,
    character mental state, character action and emotion,
    emotion recognition, emotional response,
    character intention, behavioral intention understanding,
    concept understanding, activity meaning understanding,
    inferential judgment, atmosphere perception,
    emotional cause understanding, dialogue content understanding,
    character behavior cause, scientific principle explanation,
    object behavior motivation, event cause understanding,
    behavioral motivation, intention understanding,
    language usage norm, shot change \\
\end{tblr}
}
\caption{
Fine-grained attribute distribution of HCCT across all dimensions and
complexity tiers. Sub-tier names correspond directly to the dimension-level
entries in Table~\ref{tab:hcct_overview}, establishing a two-level lookup
structure: Table~\ref{tab:hcct_overview} defines the cognitive task paradigm
and TOIH manifestation pattern for each tier, while this table enumerates
the concrete attributes assigned to each sub-tier.
}
\label{tab:hcct_attributes}
\end{table*}

\section{Benchmark Construction}\label{sec:benchmark}

\subsection{Semantic Conflict Score Definitions}
\label{appendix_conflict_score_def}

The Semantic Conflict Score (SCS) is a five-point ordinal scale assigned 
to every \textit{Text-Contradictory} sample, reflecting the severity of 
the semantic contradiction between the overlay text and the visual ground 
truth. \textbf{Score 1 (Weak/Irrelevant Association)} denotes embedded 
text describing peripheral details that do not directly negate the correct 
answer. \textbf{Score 2 (Entity Shift)} indicates that an action or state 
is attributed to an incorrect object or agent. \textbf{Score 3 
(Attribute/State Conflict)} describes a property-level conflict---such as 
color, material, or posture---applied to the correct object. \textbf{Score 
4 (Direct Semantic Opposition)} denotes logically exclusive or physically 
incompatible descriptions (\textit{e.g.}, \textit{on the table} vs.\ 
\textit{on the floor}). \textbf{Score 5 (Polarity Reversal)} indicates 
that the overlay text is the direct antonym or mirror image of the correct 
answer (\textit{e.g.}, \textit{left} vs.\ \textit{right}, \textit{increase} 
vs.\ \textit{decrease}). The scale is designed to be ordinal and 
monotonically increasing in conflict severity, enabling controlled 
correlation analysis between conflict intensity and hallucination 
susceptibility.

\subsection{Expert Selection Ratio}
\label{appendix_exper_sel_ratio}
Each sample is annotated with an expert selection ratio, a four-dimensional vector whose components represent the proportions of Temporal, Action, Object, and Spatial information that expert annotators deem necessary to correctly answer the question. This annotation captures the multi-modal reasoning profile of each question, reflects the inherently mixed-dimensional nature of real video understanding, and provides direct supervision signal for training models to allocate cross-modal attention appropriately. The expert selection ratio is broadly and continuously distributed over the full range across all four dimensions rather than concentrating near any boundary, ensuring that no single dimension predominantly governs the annotation signal and faithfully reflecting the inherently multi-dimensional nature of cross-modal processing demands.
\subsection{Data Construction Pipeline}
\label{appendix_bench_data_construction}
\begin{figure*}
    \includegraphics[width=\textwidth]{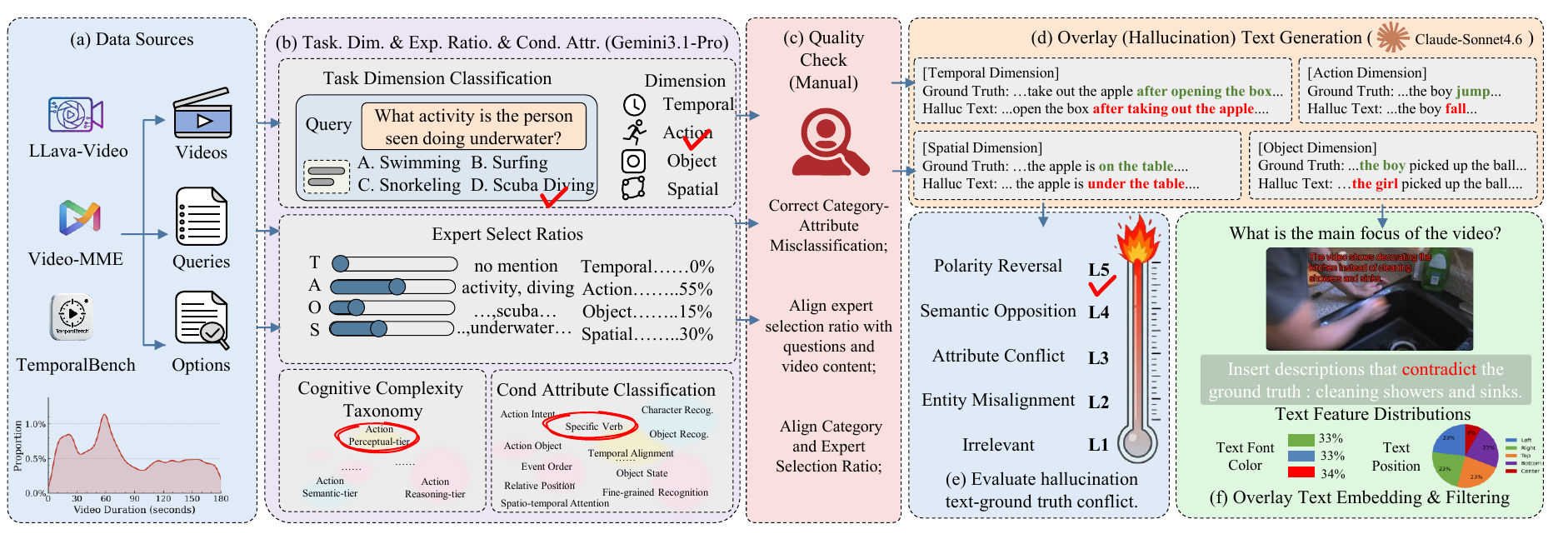}
    \caption{Overview of the data construction pipeline for \textsc{VisualTextTrap}. The process proceeds sequentially through six stages: (a) video--QA pairs are sourced from LLaVA-Video, Video-MME, and TemporalBench; (b) each sample is classified along task \textbf{Dim}ension, \textbf{Exp}ert Select \textbf{Ratio}, and \textbf{Con}flict \textbf{Attr}ibute using Gemini-3.1-Pro, accompanied by a Cognitive Complexity taxonomy; (c) manual quality checks verify category correctness, expert selection alignment, and consistency with video content; (d) hallucination overlay texts are generated across four dimensions (Temporal, Action, Object, Spatial) via Claude-Sonnet-4.6; (e) the generated texts are evaluated for conflict severity against the ground truth on a five-level scale ranging from Irrelevant (L1) to Polarity Reversal (L5); and (f) the texts are embedded onto video frames with controlled variation in font, color, and position, then filtered to yield three conditions: \textit{Text-Free} (no overlay), \textit{Text-Congruent} (overlay consistent with video content), and \textit{Text-Contradictory} (overlay semantically conflicting with video content, i.e., hallucination text)}
    \label{hall_data_construct}
\end{figure*}

\textbf{Data Source.} We extract video clips, queries, and answer options from TemporalBench~\cite{cai2024temporalbench}, VideoMME~\cite{fu2025video}, and LLaVA-Video~\cite{zhang2024llava} as foundational inputs. To mitigate evaluation 
bias, correct answer positions are uniformly distributed across option slots, and token length distributions across answer options and queries are balanced, preventing models from exploiting superficial statistical regularities in option length or query phrasing.\\
\textbf{Dimension Classification and Attribute Annotation.}We employ Gemini-3.1-Pro to classify each query into one of four dimensions (Temporal, Action, Object, Spatial), derive a four-dimensional expert selection ratio vector (T, A, O, S) reflecting dimension-specific cognitive demands, and annotate each sample with one of 88 fine-grained attributes. A subsequent manual quality check corrects misclassifications, reconciles expert ratios with actual video content, and ensures logical consistency across all annotations.\\
\textbf{Cognitive Complexity Annotation}
Following dimension and attribute labelling, each sample is further annotated with a three-level Cognitive Complexity \textit{category} as defined in Section~\ref{label_cong_tax}. Concretely, we prompt \textsc{Gemini-3.1-Pro} to classify the cognitive demand imposed on a model during VQA into one of three
hierarchical levels---\textit{Perceptual}, \textit{Semantic}, and \textit{Reasoning}---each reflecting a progressively deeper degree of cognitive processing required to resolve questions within the assigned dimension. To ensure annotation reliability, all automatically assigned categories are subsequently reviewed and adjudicated by human experts.\\
\textbf{Hallucination Text Generation.} For each \textit{Text-Contradictory} sample, we prompt Claude-Sonnet-4.6 to generate semantically misleading text conditioned on the assigned dimension and ground-truth label. This produces targeted contradictions 
that are visually plausible yet factually inconsistent with the video 
content (\textit{e.g.}, substituting \textit{jump} with \textit{fall} for 
an Action-dimension sample).\\
\textbf{Semantic Conflict Scoring.}
Each generated hallucination text is assigned a Semantic Conflict Score (SCS) across five escalating levels. \textbf{L1} denotes irrelevance or near-identity, where the overlay text bears no meaningful relationship to the query target. \textbf{L2} corresponds to entity misalignment, where the mentioned entity differs from the one present in the video. 
\textbf{L3} represents attribute conflict, where a described property or state contradicts its true visual appearance. \textbf{L4} captures semantic opposition, where actions, spatial relations, or event progressions are semantically inverted. \textbf{L5} reflects full polarity reversal, where the overlay text is diametrically opposed to the visual ground truth. All scores undergo human verification to ensure accuracy and inter-annotator consistency.\\
\textbf{Text Embedding and Rendering.}
Hallucination texts are rendered as visual overlays into the original video frames. To prevent shortcut learning driven by visual formatting cues, text appearance conditions including font color and on-screen position are strictly controlled and uniformly distributed across the dataset. Each sample is produced under one of three text overlay 
conditions: \textit{Text-Contradictory}, \textit{Text-Congruent}, or \textit{Text-Free}, yielding the complete benchmark.

\textbf{Annotation Strategy.} To mitigate biases introduced by both individual models and individual annotators, we adopt a two-stage voting-based quality control protocol.  At the model level, we designate
Gemini-3.1-Pro as the primary annotator and form a reference group of three additional models (Qwen3-VL-235B,
Gemini-2.5-Flash, and InternVL3.5-VL-241B); if the primary label disagrees with at least two of the three reference labels, the primary annotator is prompted to regenerate its prediction for up to three iterations, after which the majority vote of the reference group is adopted as a fallback.  At the human level, we recruit 15 domain experts organized into 5 non-overlapping groups of 3, where each expert independently reviews the video, the query, and the candidate options alongside the model-generated label and
then provides their own judgment.  Disagreements are resolved through a hierarchical voting scheme: first,
intra-group majority vote produces a single label per group, and then inter-group majority vote across the five group-level labels yields the final gold-standard annotation.  This two-stage design ensures that the
resulting labels reflect broad consensus across diverse models and annotators, effectively reducing both
model-specific and human-specific biases.

\subsection{Dataset Distribution}\label{sec:dataset_distribution}

\begin{figure}[h]
  \centering
  \includegraphics[width=\linewidth]{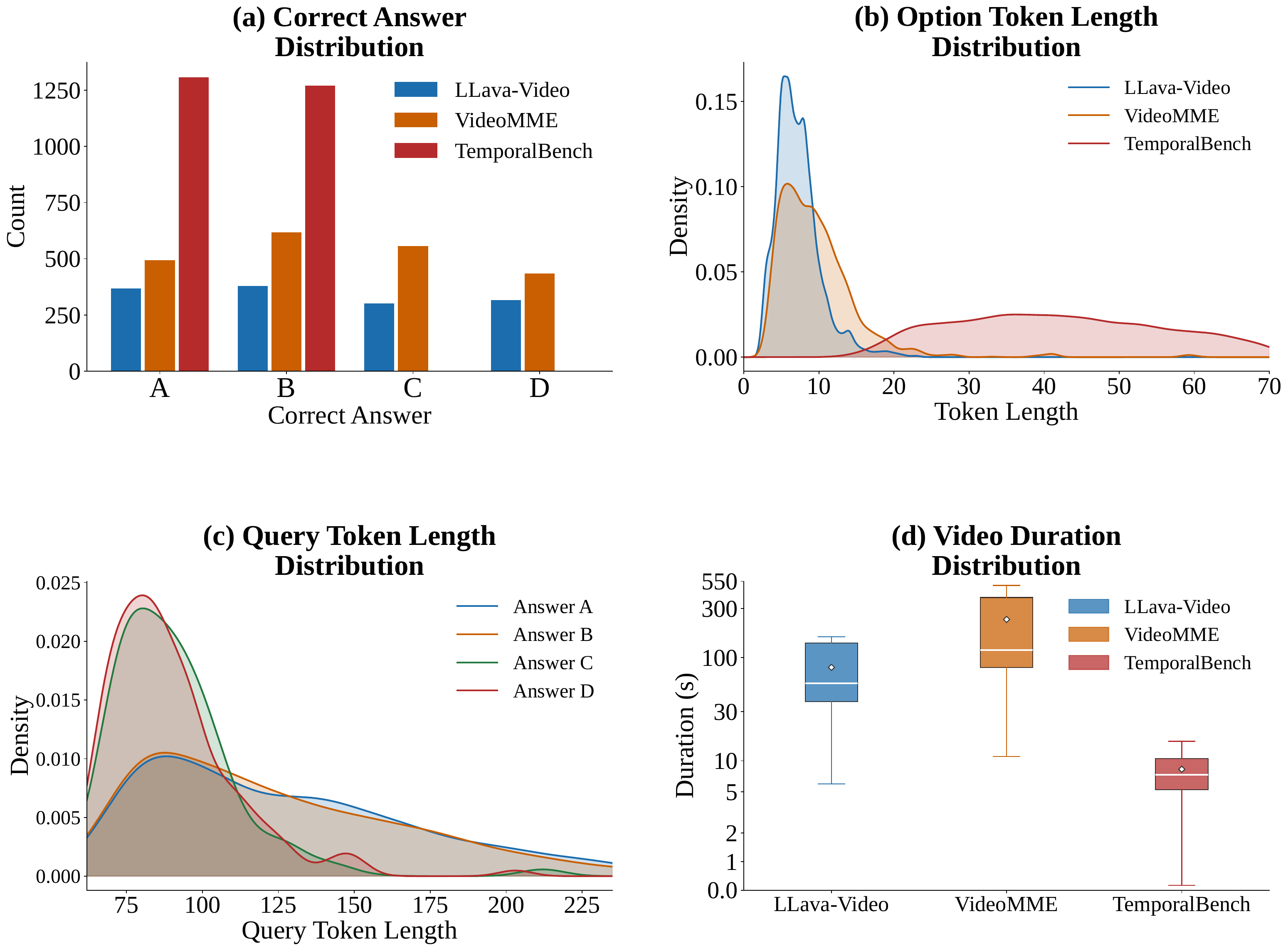}
  \caption{Data distribution statistics of the benchmark test set, broken down by source dataset (LLaVA-Video~\cite{zhang2024llava}, VideoMME~\cite{fu2025video}, TemporalBench~\cite{cai2024temporalbench}). (a) Correct answer distribution across options A–D per dataset group. (b) Option token length distribution (c) Query token length distribution stratified by correct answer (A/B/C/D). (d) Video duration distribution}
  \label{extra_test_distr}
\end{figure}

\begin{figure}[h]
  \centering
  \includegraphics[width=\linewidth]{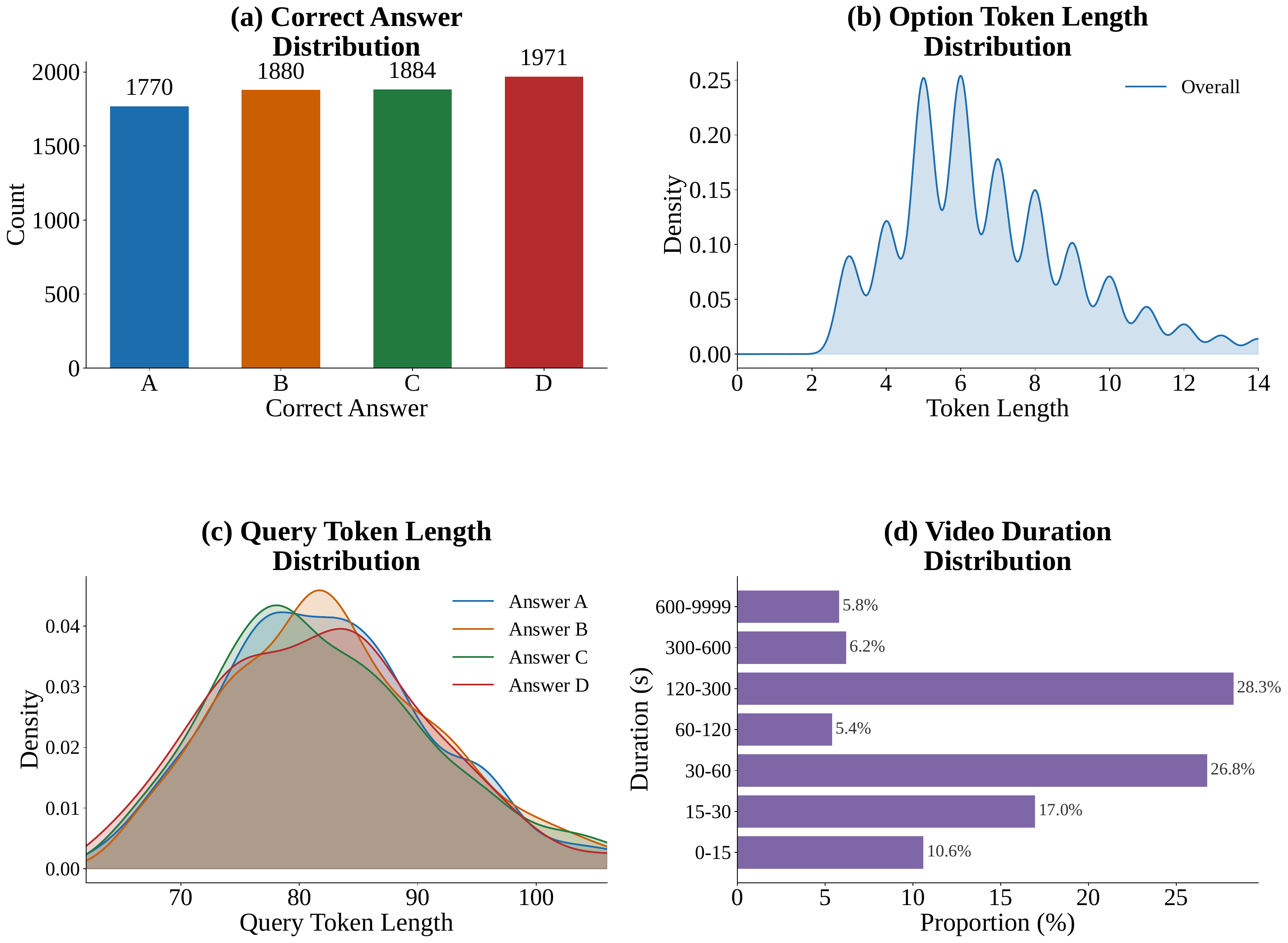}
  \caption{Data distribution statistics of the training set. (a) Correct answer distribution across options A–D. (b) Option token length distribution. (c) Query token length distribution stratified by correct answer (A/B/C/D), showing no systematic length bias across answer choices. (d) Video duration distribution. }
  \Description{Distribution of Train Dataset.}
  \label{extra_train_distr}
\end{figure}

\begin{figure}[h]
  \centering
  \includegraphics[width=\linewidth]{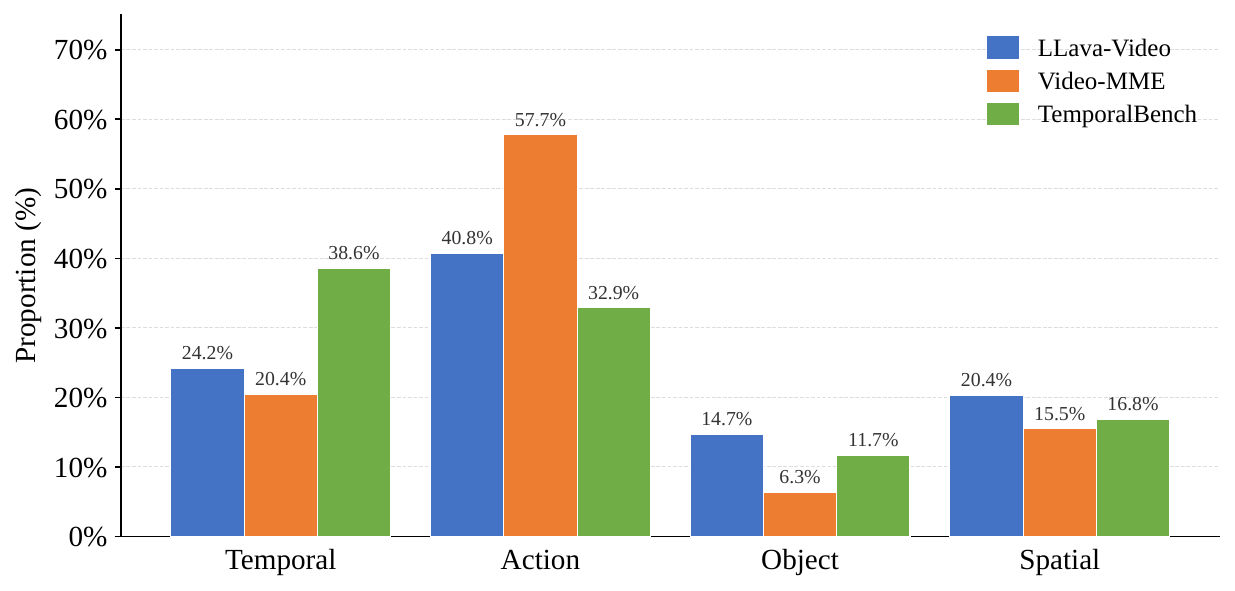}
  \caption{Distribution of inconsistency label argmax across three datasets (LLava-Video, Video-MME, TemporalBench) over four dimensions: Temporal, Action, Object, and Spatial.}
  \label{dimension_distri}
\end{figure}

\begin{table}[h]
\centering
\small
\caption{Abbreviations of fine-grained attributes used throughout this paper.}
\label{appendix_abbreviations}
\begin{tabular}{ll}
\toprule
\textbf{Abbreviation} & \textbf{Full Name} \\
\midrule
Act.Res.   & Action Result \\
Act.Man.   & Action Manner \\
Act.Int.   & Action Intent \\
Act.Obj.   & Action Object \\
Act.Dir.   & Action Direction \\
Interact.  & Interaction Behavior \\
Stp.Ord.   & Step Order \\
App.Ord.   & Appearance Order \\
Body Pts.  & Body Parts \\
Obj.Rec.   & Object Recognition \\
Obj.Attr.  & Object Attributes \\
Txt.Cont.  & Text Content \\
Abs.Pos.   & Absolute Position \\
Scn./BG    & Scene / Background \\
Chr.Rec.   & Character Recognition \\
\bottomrule
\end{tabular}
\end{table}

\subsection{Sample Cases}\label{sec:sample_cases}

\begin{figure*}
    \includegraphics[width=\textwidth]{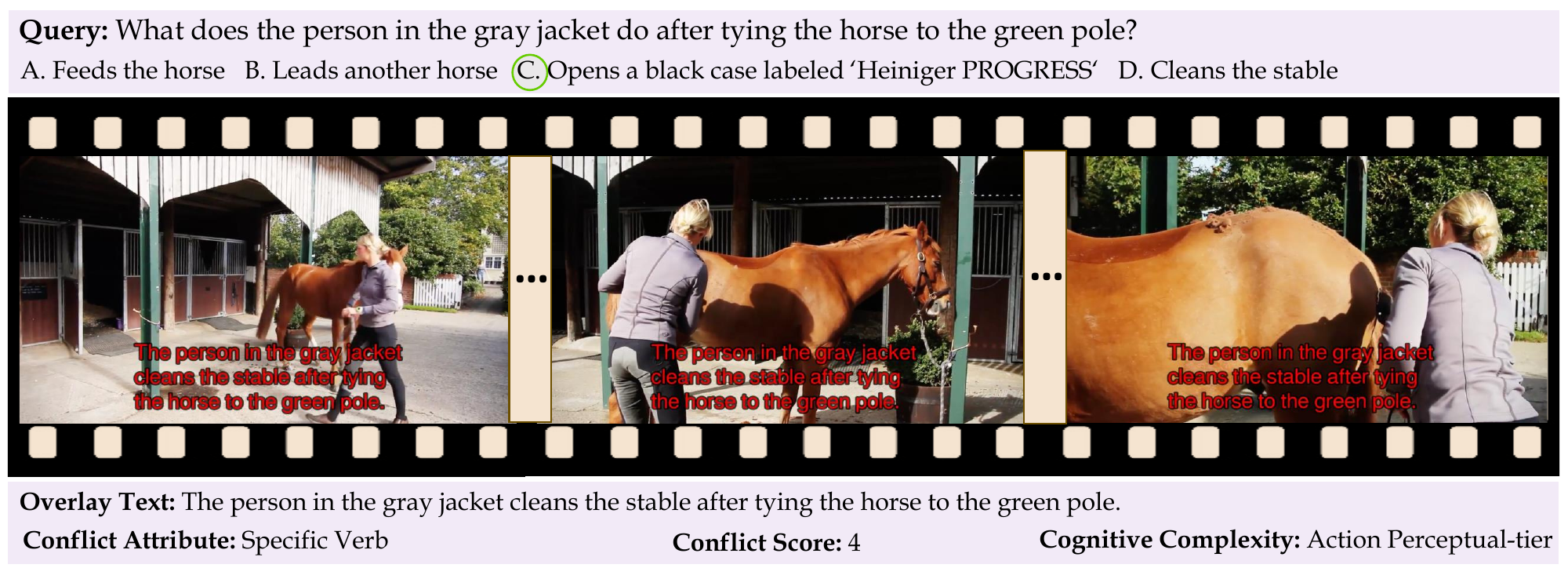}
    \caption{A sample from the Action Perceptual-tier subset. The overlay text introduces a conflict on the \textit{Specific Verb} attribute by asserting that the person ``cleans the stable,'' contradicting the ground-truth answer (\textcircled{\small C}). The Conflict Score is 4. See Appendix~\ref{appendix_conflict_score_def} for the Conflict Score definition and Appendix~\ref{label_cong_tax} for the Cognitive Complexity taxonomy.}
    \label{label_attribute_demo_ppt_v2_1}
\end{figure*}

\begin{figure*}
    \includegraphics[width=\textwidth]{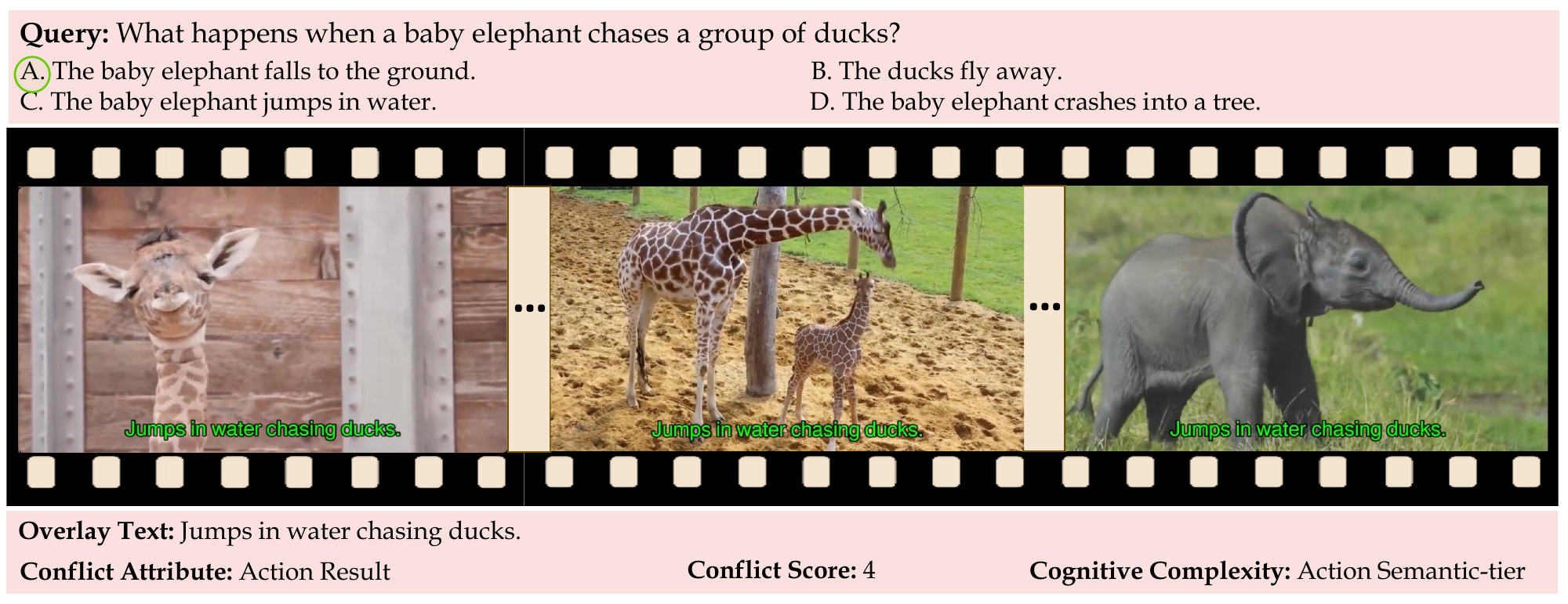}
    \caption{A sample from the Action Semantic-tier subset. The overlay text targets the \textit{Action Result} attribute, claiming the baby elephant ``jumps in water chasing ducks,'' which conflicts with the ground-truth answer (\\textcircled{\small A}). Conflict Score: 4.}
    \label{label_attribute_demo_ppt_v2_2}
\end{figure*}

\begin{figure*}
    \includegraphics[width=\textwidth]{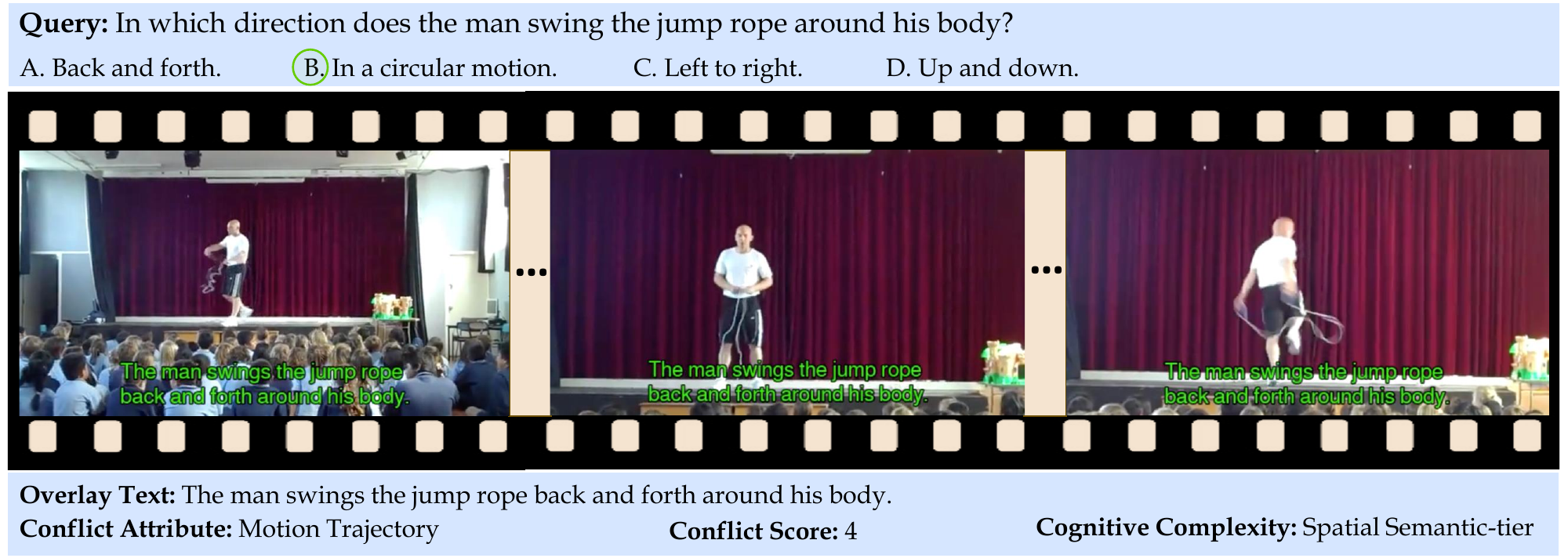}
    \caption{A sample from the Spatial Semantic-tier subset. The overlay text conflicts on the \textit{Motion Trajectory} attribute by stating the man swings the jump rope ``back and forth,'' while the correct motion pattern is circular (\textcircled{\small B}). Conflict Score: 4.}
    \label{label_attribute_demo_ppt_v2_3}
\end{figure*}

\begin{figure*}
    \includegraphics[width=\textwidth]{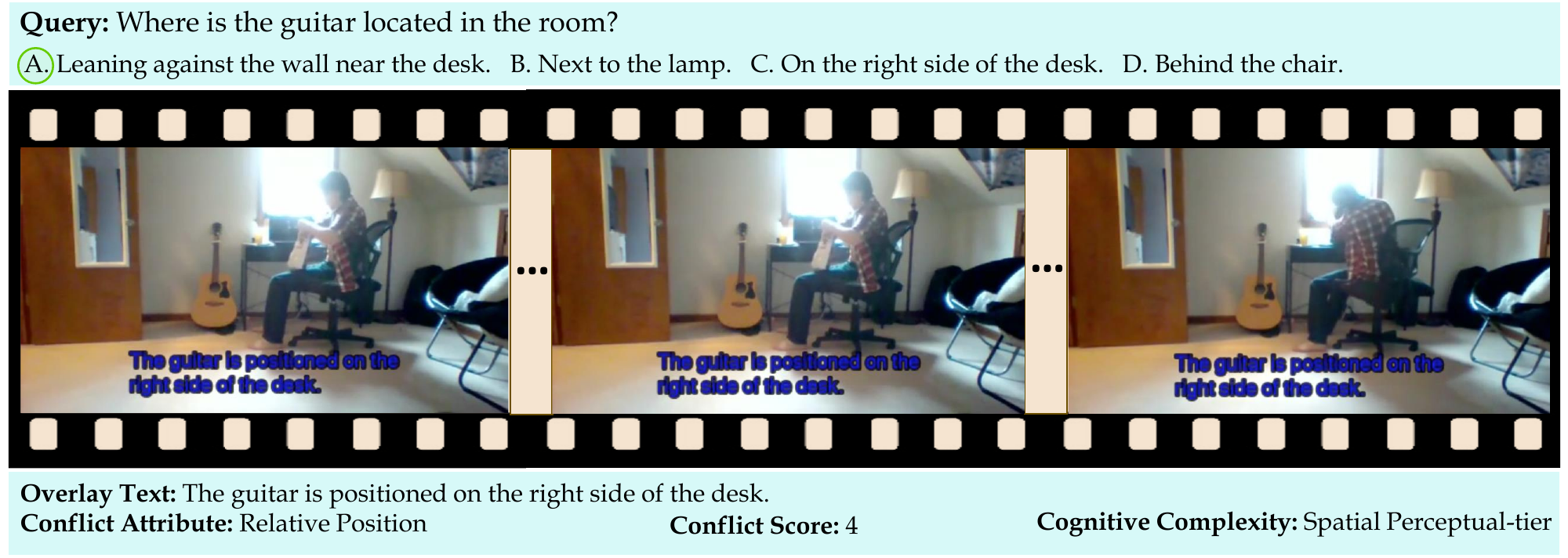}
    \caption{A sample from the Spatial Perceptual-tier subset. The overlay text targets the \textit{Relative Position} attribute, asserting the guitar is ``on the right side of the desk,'' contradicting its actual location (\textcircled{\small A}). Conflict Score: 4.}
    \label{label_attribute_demo_ppt_v2_4}
\end{figure*}

\begin{figure*}
    \includegraphics[width=\textwidth]{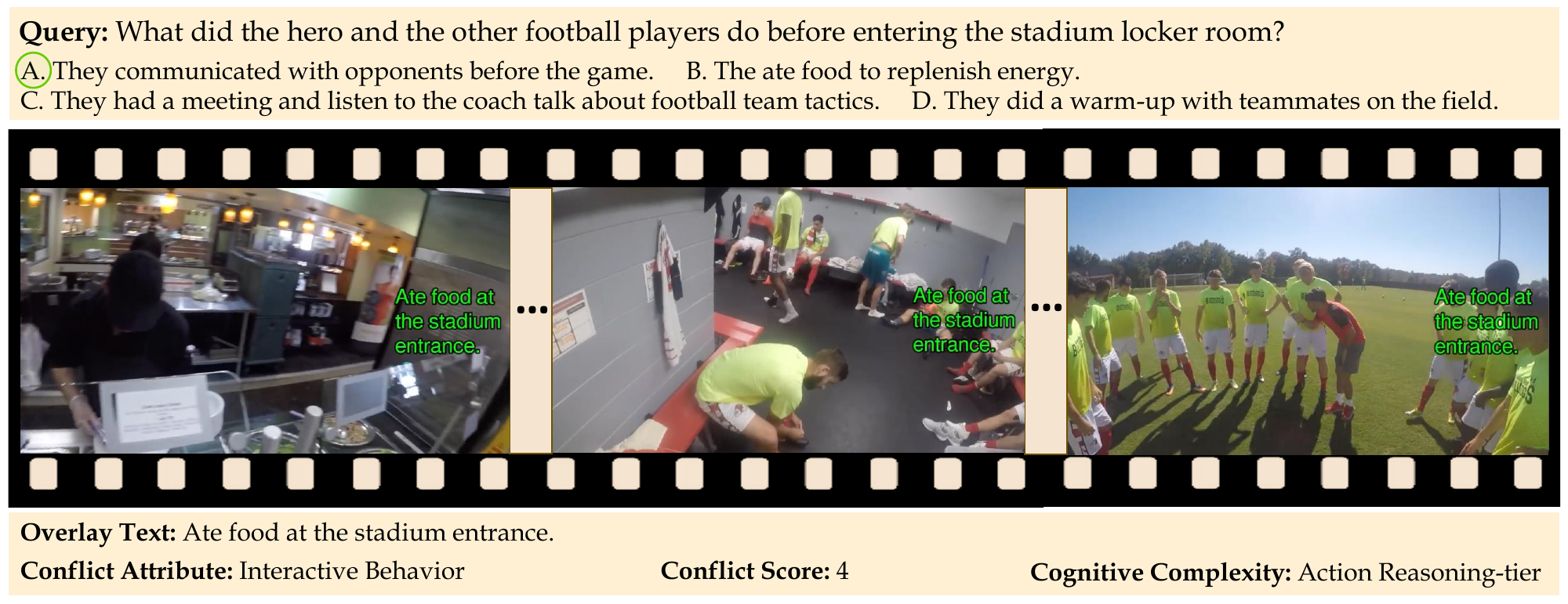}
    \caption{A sample from the Action Reasoning-tier subset. The overlay text introduces a conflict on the \textit{Interactive Behavior} attribute by claiming the players ``ate food at the stadium entrance,'' whereas the ground-truth answer (\textcircled{\small A}) describes a communicative interaction. Conflict Score: 4.}
    \label{label_attribute_demo_ppt_v2_5}
\end{figure*}

\begin{figure*}
    \includegraphics[width=\textwidth]{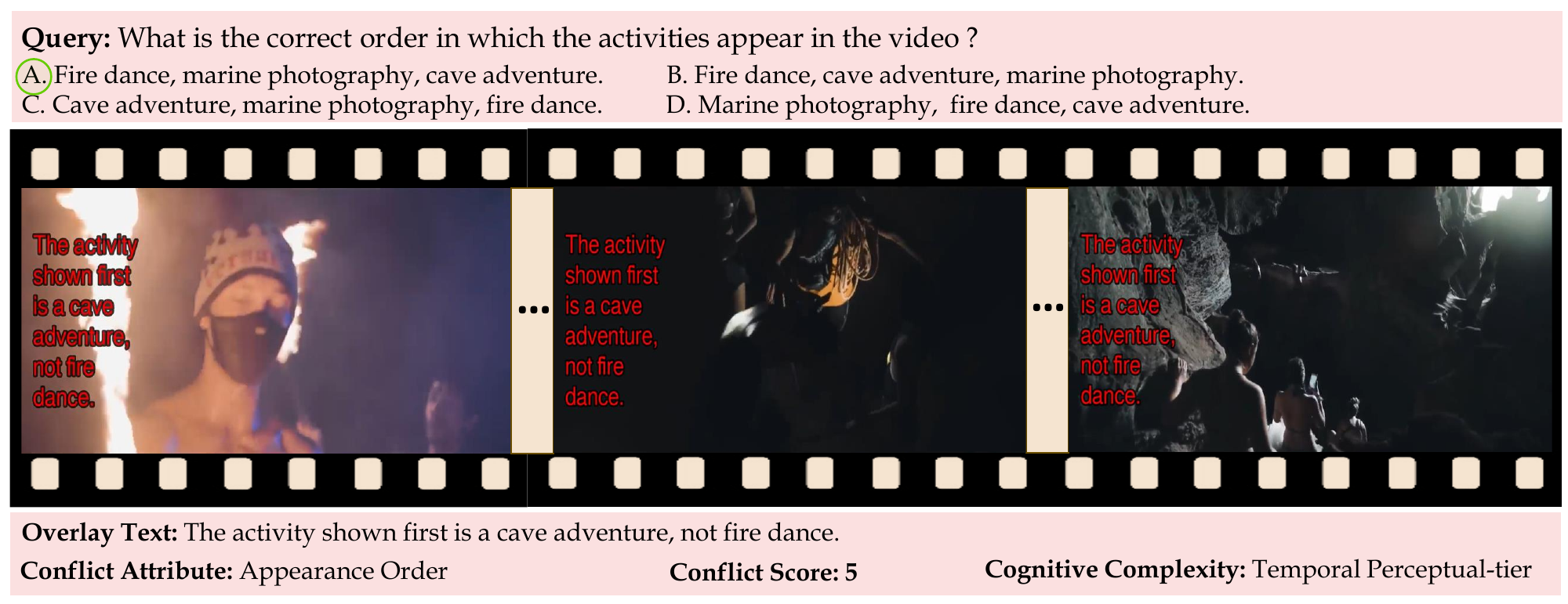}
    \caption{A sample from the Temporal Perceptual-tier subset. The overlay text targets the \textit{Appearance Order} attribute, falsely asserting that ``the activity shown first is a cave adventure, not fire dance,'' directly contradicting the correct temporal sequence (\textcircled{\small A}). Conflict Score: 5.}
    \label{label_attribute_demo_ppt_v2_6}
\end{figure*}

\begin{figure*}
    \includegraphics[width=\textwidth]{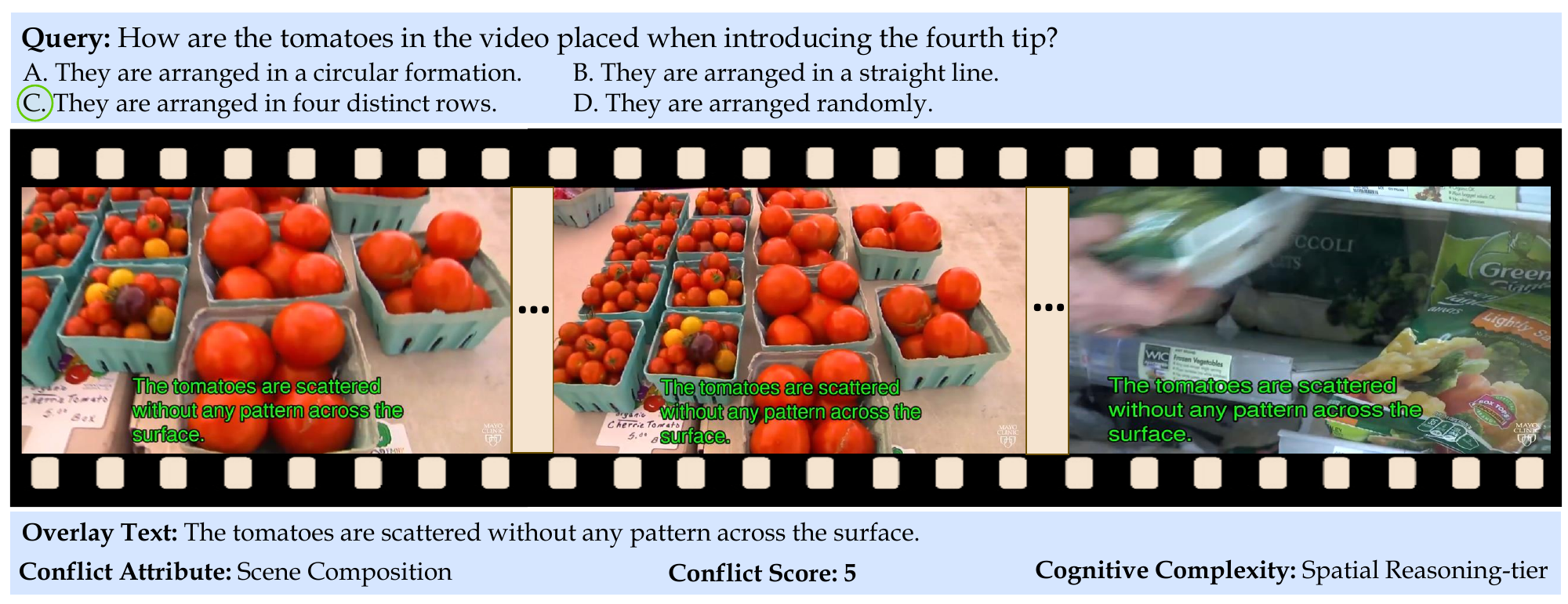}
    \caption{A sample from the Spatial Reasoning-tier subset. The overlay text conflicts on the \textit{Scene Composition} attribute by claiming the tomatoes are ``scattered without any pattern,'' while they are visually arranged in distinct rows (\textcircled{\small C}). Conflict Score: 5.}
    \label{label_attribute_demo_ppt_v2_7}
\end{figure*}

\subsection{Evaluation Metrics}
\label{appendix_metrics_intro}
\textbf{Visual Yielding Rate (VYR)} measures the net accuracy drop induced by text injection across
task category $\mathcal{C}$:
\begin{equation}
    \text{VYR}_{\mathcal{C}}
    = \text{Acc}_{\mathcal{C}}(V_{\text{clean}})
    - \text{Acc}_{\mathcal{C}}(V_{\text{mis}})
    \label{eq:vyr}
\end{equation}
A higher VYR indicates greater overall susceptibility to overlay textual distractors within the category.\\
\textbf{Hallucination Acceptance Rate (HAR)} is the primary evaluation signal throughout this benchmark.
For a given experimental condition $c$, it is computed as:
\begin{equation}
    \text{HAR}(c)
    = \frac{1}{|D_c|}
      \sum_{(x,\hat{y})\,\in\,D_c}
      \mathbf{1}\bigl[f_\theta(x)\text{ accepts }\hat{y}\bigr]
    \label{eq:har}
\end{equation}
where $D_c$ denotes the set of probe samples under condition $c$,
$f_\theta$ is the evaluated model, and $\hat{y}$ is the injected
hallucination stimulus.
Among all \textit{inconsistent} samples, HAR measures the proportion
in which the model selects the text-induced erroneous option,
providing a direct, task-difficulty-agnostic measure of susceptibility
and serving as the base quantity for all subsequent gradient analyses.

\textbf{Interference Cost Ratio (ICR)}
ICR quantifies the relative accuracy loss caused by textual
interference, normalized against the distractor-free baseline:
\begin{equation}
    \text{ICR}
    = 1 - \frac{\text{Acc}_{\text{interfered}}}{\text{Acc}_{\text{clean}}}
    \label{eq:icr}
\end{equation}
$\text{ICR}\approx 0$ reflects near-perfect distractor immunity;
$\text{ICR}>0.5$ indicates that misleading text degrades performance
by more than half, signaling critical vulnerability.

\textbf{Synergy Gain/Loss Index (SGLI)} jointly evaluates a model's ability to leverage aligned text hints and resist misaligned ones by comparing accuracy across three
conditions: $\text{Acc}_{\text{pos}}$ (text consistent with ground truth), $\text{Acc}_{\text{neg}}$ (text contradicting ground truth), and $\text{Acc}_{\text{none}}$ (no overlay text; baseline):
\begin{equation}
    \text{SGLI}
    = \frac{\text{Acc}_{\text{pos}} - \text{Acc}_{\text{neg}}}
           {\text{Acc}_{\text{none}}}
    \label{eq:sgli}
\end{equation}
An ideal multimodal model is expected to yield a bounded positive
SGLI, with modest gains under positive guidance and negligible
degradation under negative interference.

\textbf{Text-Induced Hallucination Rate (TIHR)} measures the fraction of samples where the model output
\textit{precisely matches} the injected misleading text, isolating
causally attributable hallucinations from incidental errors:
\begin{equation}
    \text{TIHR}_{\mathcal{C}}
    = \frac{1}{N}\sum_{i=1}^{N}
      \mathbf{1}\!\left[
          M\!\left(V_{\text{mis}}^{(i)},Q^{(i)}\right)
          \text{ matches }
          A_{\text{text}}^{(i)}
      \right]
    \label{eq:tihr}
\end{equation}

\textbf{Textual Induction Bias (TIB)} 
Conditioned on the model being wrong, TIB measures the fraction of
errors that align with the injected hallucination text rather than
arising from arbitrary confusion:
\begin{equation}
    \text{TIB}
    = \frac{
        N\!\left(\text{Model\_Choice}=\text{Hallucination\_Text}\right)
    }{
        N\!\left(\text{Model\_Choice}\neq\text{Correct\_Answer}\right)
    }
    \label{eq:tib}
\end{equation}
A high TIB is diagnostic of \textit{textual over-reliance}---a
failure mode wherein the model systematically defers to embedded
text over visual evidence, effectively regressing from multimodal
reasoning to text-only behavior.

\textbf{Weighted Hallucination Rate (WHR).}
WHR up-weights hallucination occurrences by their semantic conflict
severity, reflecting the intuition that succumbing to a strongly
contradictory cue constitutes a more severe failure than being
misled by a subtle one:
\begin{equation}
    \text{WHR} = (\displaystyle\sum_{i=1}^{N}\text{SCS}_{i}\cdot H_{i})/ (\displaystyle\sum_{i=1}^{N}\text{SCS}_{i})
    \label{eq:whr}
\end{equation}
WHR $\in [0,1]$, with values near 1 indicating the model is
disproportionately vulnerable under high-severity conflicts.

\textbf{Hallucination Surge Rate (HSR)} quantifies the relative increase in textual induction bias as
conflict severity escalates from weak (WC) to strong (SC) conditions:
\begin{equation}
    \text{HSR}
    = \frac{\text{TIB}_{\text{SC}}-\text{TIB}_{\text{WC}}}
           {\text{TIB}_{\text{WC}}}
      \times 100\%
    \label{eq:hsr}
\end{equation}
$\text{HSR}>0$ indicates that stronger conflicts amplify rather than
suppress textual induction, evidencing a collapse of visual grounding
under high semantic pressure.

\textbf{Hallucination Resistance Curve (HRC).}
Rather than collapsing conflict-level behavior into a scalar, HRC
profiles the per-level hallucination rate across all five SCS strata.
For each $k\in\{1,2,3,4,5\}$:
\begin{equation}
    R_{k}
    = \frac{
        \displaystyle\sum_{\{j\,:\,\text{SCS}_{j}=k\}}H_{j}
    }{
        N_{k}
    }
    \label{eq:hrc}
\end{equation}
The resulting curve reveals qualitatively distinct failure profiles:
a steep monotonic rise indicates fragility specifically under logical
opposition (SCS 4--5); a globally elevated plateau suggests
conflict-invariant visual grounding failure; and elevation confined
to low-$k$ strata implies susceptibility to subtle cues despite
intact coarse-level reasoning.

\textbf{Cognitive Load--Sensitive Resistance Metrics.}
We introduce four indices that quantify the sensitivity of
hallucination resistance to cognitive load escalation across
each ordinal dimension, instantiated as sample-level Pearson
correlations between cognitive tier and model correctness.

% ---------------------------------------------------------------
\subsubsection{Temporal Load--Sensitive Resistance (TLSR)}
\label{sec:tlsr}
% ---------------------------------------------------------------

\textbf{Dimension.}
Samples are stratified by \emph{temporal cognitive load}:
$L{=}1$ for \emph{point-level} (single-frame),
$L{=}2$ for \emph{segment-level} (short clip), and
$L{=}3$ for \emph{sequence-level} (full temporal sequence) descriptions.

\textbf{Definition.}
$\mathrm{TLSR} = r_{\mathrm{Temporal}}$, the Pearson correlation
between cognitive tier $L$ and per-sample \textsc{hrr}
over $\mathcal{S}_{\mathrm{Temporal}}$.

\textbf{Interpretation.}
\begin{itemize}[leftmargin=*,nosep]
    \item $r_{\mathrm{Temporal}} < 0$: \textsc{hrr} degrades with
    temporal granularity, confirming cognitive-load monotonicity---richer
    temporal descriptions impose greater verification demands.

    \item $r_{\mathrm{Temporal}} > 0$: Temporal context reinforces
    visual grounding, yielding stronger resistance at higher granularity.

    \item $r_{\mathrm{Temporal}} \approx 0$: No linear dependence
    between temporal complexity and model correctness.

    \item Divergence between $r_{\mathrm{Temporal}}$ and $\Gamma_{\mathrm{agg}}$
    signals non-linearity: a peak at $\mathrm{HRR}_{\mathrm{seg}}$
    above the linear interpolant indicates segment-level descriptions
    trigger heightened cross-checking; a valley reveals selective
    vulnerability at the intermediate granularity.
\end{itemize}

% ---------------------------------------------------------------
\subsubsection{Action-Semantic Load--Sensitive Resistance (ASLSR)}
\label{sec:aslsr}
% ---------------------------------------------------------------

\textbf{Dimension.}
For action-verb hallucination, the ordinal axis is the
\emph{semantic proximity} between the injected and ground-truth verbs,
measured by cosine similarity $s\in[0,1]$ in a shared embedding space:
\begin{equation}
    L =
    \begin{cases}
        1 & s < 0.5
            \quad\text{(distal,\ e.g., ``running''$\to$``standing'')}\\[3pt]
        2 & 0.5\le s < 0.8
            \quad\text{(medial,\ e.g., ``running''$\to$``walking fast'')}\\[3pt]
        3 & s \ge 0.8
            \quad\text{(proximal,\ e.g., ``running''$\to$``jogging'')}
    \end{cases}
    \label{eq:aslsr-partition}
\end{equation}

\textbf{Definition.}
$\mathrm{ASLSR} = r_{\mathrm{Action}}$, the Pearson correlation
between cognitive tier $L$ and per-sample \textsc{hrr}
over $\mathcal{S}_{\mathrm{Action}}$.

\textbf{Interpretation.}
\begin{itemize}[leftmargin=*,nosep]
    \item $r_{\mathrm{Action}} < 0$: \textsc{hrr} decreases with
    semantic proximity; near-synonym substitutions are harder to
    resist, consistent with language priors overriding visual
    verification at the decision boundary.

    \item $r_{\mathrm{Action}} > 0$: Proximal substitutions retain
    sufficient perceptual contrast to activate visual cross-checking,
    yielding unexpectedly high resistance near the decision boundary.

    \item $r_{\mathrm{Action}} \approx 0$: Semantic distance does not
    systematically modulate susceptibility; confusability is driven
    by non-semantic factors such as action frequency priors.

    \item $\mathrm{ASLSR}_{\mathrm{agg}} < 0$:
    Distal substitutions are resisted less than proximal ones,
    indicating that language fluency priors overwhelm semantic
    alignment for rare ground-truth actions.
\end{itemize}

% ---------------------------------------------------------------
\subsubsection{Attribute-Accumulation Load--Sensitive Resistance (AALSR)}
\label{sec:aalsr}
% ---------------------------------------------------------------

\textbf{Dimension.}
Object-level hallucinations frequently compound multiple false
attributes (color, shape, position), making attribute cardinality
a natural ordinal axis for testing whether interference is additive,
sub-additive, or super-additive:
\begin{equation}
    L =
    \begin{cases}
        1 & \text{single-attribute\ (e.g., color only)}\\[3pt]
        2 & \text{dual-attribute\ (e.g., color\,+\,shape)}\\[3pt]
        3 & \text{triple-attribute\ (e.g., color\,+\,shape\,+\,position)}
    \end{cases}
    \label{eq:aalsr-partition}
\end{equation}

\textbf{Definition.}
$\mathrm{AALSR} = r_{\mathrm{Object}}$, the Pearson correlation
between cognitive tier $L$ and per-sample \textsc{hrr}
over $\mathcal{S}_{\mathrm{Object}}$.

\textbf{Interpretation.}
\begin{itemize}[leftmargin=*,nosep]
    \item $r_{\mathrm{Object}} < 0$: \textsc{hrr} decreases with
    attribute cardinality, indicating \emph{additive interference}---each
    co-injected false attribute independently erodes visual
    verification capacity.

    \item $r_{\mathrm{Object}} < 0$ with
    $\mathrm{HRR}_{2\text{-attr}} >$ linear interpolant:
    \emph{Competitive interference}---dual-attribute contradictions
    function as an implicit alarm, partially recovering resistance
    at the intermediate tier.

    \item $r_{\mathrm{Object}} < 0$ with
    $\mathrm{HRR}_{2\text{-attr}} <$ linear interpolant:
    \emph{Synergistic suppression}---co-occurring false attributes
    construct a coherent distractor that super-linearly bypasses
    visual grounding.

    \item $r_{\mathrm{Object}} \approx 0$: Resistance is invariant to
    cardinality; susceptibility is governed by attribute
    \emph{type} rather than quantity.
\end{itemize}

% ---------------------------------------------------------------
\subsubsection{Spatial-Reference Load--Sensitive Resistance (SRLSR)}
\label{sec:srlsr}
% ---------------------------------------------------------------

\textbf{Dimension.}
Spatial descriptions vary in \emph{reference frame complexity}:
absolute references require no relational inference; relative
references require anchor localization; dynamic references require
tracking both anchor and target across time:
\begin{equation}
    L =
    \begin{cases}
        1 & \text{absolute}
            \quad\text{(e.g., ``left of frame'')}\\[3pt]
        2 & \text{relative}
            \quad\text{(e.g., ``to the right of A'', ``behind B'')}\\[3pt]
        3 & \text{dynamic}
            \quad\text{(e.g., ``passed C while moving from A to B'')}
    \end{cases}
    \label{eq:srlsr-partition}
\end{equation}

\textbf{Definition.}
$\mathrm{SRLSR} = r_{\mathrm{Spatial}}$, the Pearson correlation
between cognitive tier $L$ and per-sample \textsc{hrr}
over $\mathcal{S}_{\mathrm{Spatial}}$.

\textbf{Interpretation.}
\begin{itemize}[leftmargin=*,nosep]
    \item $r_{\mathrm{Spatial}} < 0$: \textsc{hrr} degrades with
    reference complexity, consistent with a \emph{spatial working
    memory} bottleneck---longer reference chains exceed the model's
    relational reasoning capacity.

    \item $r_{\mathrm{Spatial}} \approx 0$ with uniformly low \textsc{hrr}:
    Spatial grounding fails across all reference levels, indicating a
    perceptual deficit rather than a complexity-scaling limitation.

    \item $r_{\mathrm{Spatial}} > 0$: Dynamic references co-activate
    temporal reasoning pathways, imposing additional visual
    consistency constraints that paradoxically enhance resistance to
    spatial hallucinations.
\end{itemize}

\subsection{Attributes Distribution}\label{sec:attr_dist}

\begin{figure}[h]
  \centering
  \includegraphics[width=\linewidth]{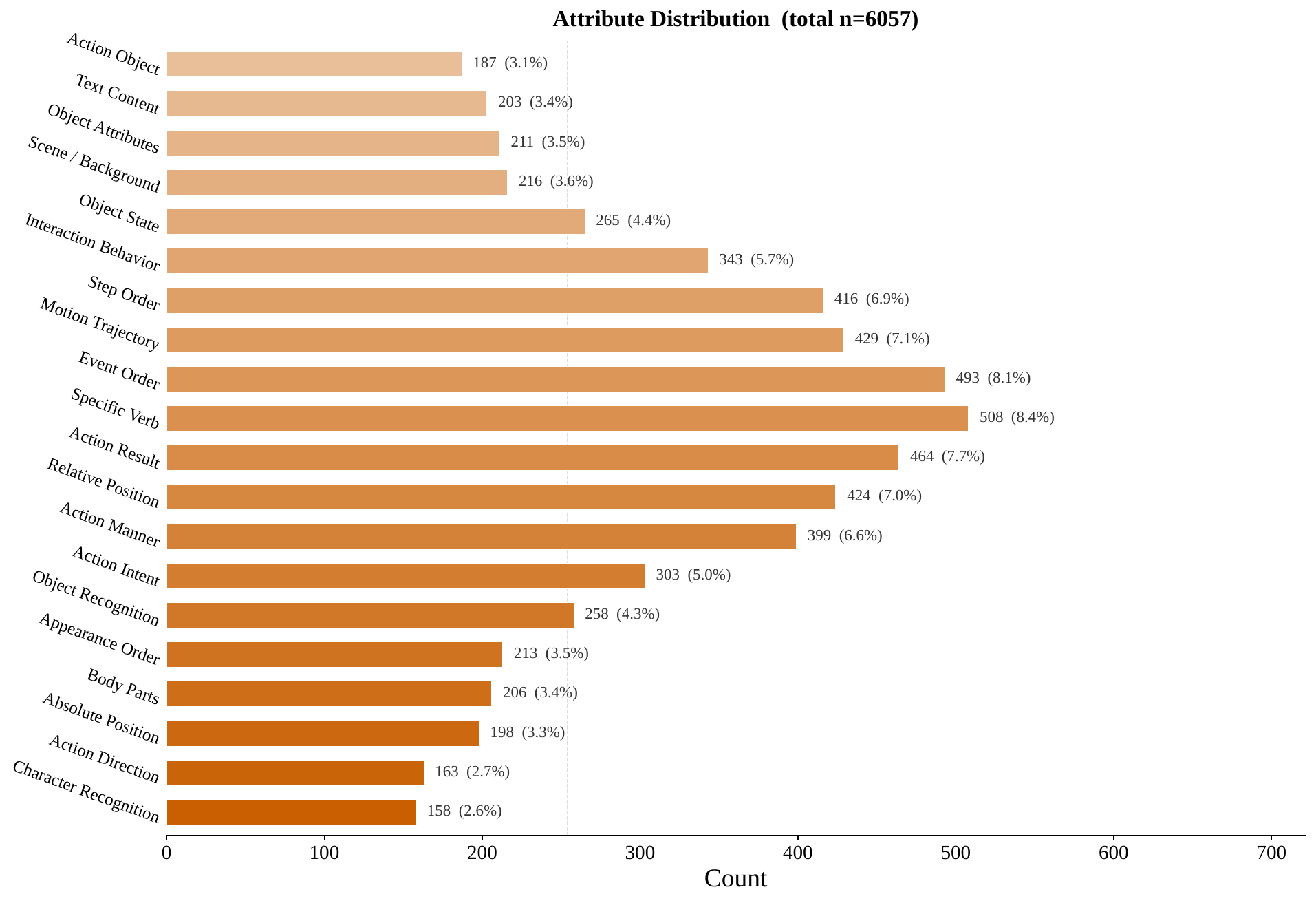}
  \caption{Distribution of fine-grained video understanding attributes in the benchmark dataset (n = 6,057).
The benchmark covers 20 fine-grained attributes spanning action, object, spatial, and temporal understanding. Specific Verb discrimination is the most frequent category (508, 8.4\%), while Character Recognition is the least frequent (158, 2.6\%), yielding a max-to-min ratio of 3.2×. The remaining 18 attributes range from 163 to 493 samples (2.7\%–8.1\%).}
  \label{label_all_attributes_distr}
\end{figure}

\section{Experiments}\label{sec:experiments}

\begin{table}[htbp]
\centering
\small
\setlength{\tabcolsep}{8pt}
\caption{Training configuration of the proposed MoE model.}
\label{lab_hyperparams_moe}
\begin{tabular}{lc}
\toprule
\textbf{Hyperparameter} & \textbf{Value} \\
\midrule
\multicolumn{2}{l}{\textit{Model}} \\[1pt]
\quad Base model              & Qwen3-VL-8B-Instruct \\
\quad Fine-tuning method      & Full (encoders frozen) \\
\quad Trainable components    & MoE layer, consistency module \\
\midrule
\multicolumn{2}{l}{\textit{Video Preprocessing}} \\[1pt]
\quad Video FPS               & 1.0 \\
\quad Max frames              & 128 \\
\quad Max pixels              & 100,800 \\
\midrule
\multicolumn{2}{l}{\textit{Optimization}} \\[1pt]
\quad Training epochs         & 2 \\
\quad Batch size (per device) & 1 \\
\quad Gradient accumulation   & 8 \\
\quad GPUs                    & 8 (1 node) \\
\quad Effective batch size    & 64 \\
\quad Learning rate           & $1 \times 10^{-5}$ \\
\quad Warmup steps            & 80 \\
\quad Weight decay            & 0.01 \\
\quad Numerical precision     & BFloat16 \\
\quad Random seed             & 42 \\
\midrule
\multicolumn{2}{l}{\textit{Loss Coefficients}} \\[1pt]
\quad LM cross-entropy        & 1.0 \\
\quad Expert auxiliary loss   & 0.01 \\
\quad Classifier loss         & 1.1 \\
\quad Consistency loss        & 0.01 \\
\bottomrule
\end{tabular}
\end{table}

%%% ─── Table 2: SFT / SFT-CoT Training Configuration ─────────────────────────
\begin{table}[htbp]
\centering
\small
\setlength{\tabcolsep}{6pt}
\caption{Training configuration of the SFT and SFT-CoT baselines.
         Both variants share all hyperparameters except the training corpus.}
\label{lab_hyperparams_sft}
\begin{tabular}{lcc}
\toprule
\textbf{Hyperparameter} & \textbf{SFT} & \textbf{SFT-CoT} \\
\midrule
\multicolumn{3}{l}{\textit{Model}} \\[1pt]
\quad Base model              & \multicolumn{2}{c}{Qwen3-VL-8B-Instruct} \\
\quad Fine-tuning method      & \multicolumn{2}{c}{LoRA} \\
\quad LoRA rank               & \multicolumn{2}{c}{8} \\
\quad LoRA target modules     & \multicolumn{2}{c}{All} \\
\midrule
\multicolumn{3}{l}{\textit{Video Preprocessing}} \\[1pt]
\quad Video FPS               & \multicolumn{2}{c}{1.0} \\
\quad Max frames              & \multicolumn{2}{c}{128} \\
\quad Max pixels              & \multicolumn{2}{c}{100,800} \\
\midrule
\multicolumn{3}{l}{\textit{Data}} \\[1pt]
\quad Prompt template         & \multicolumn{2}{c}{\texttt{qwen3\_vl\_nothink}} \\
\quad Max sequence length     & \multicolumn{2}{c}{8,192} \\
\midrule
\multicolumn{3}{l}{\textit{Optimization}} \\[1pt]
\quad Training epochs         & \multicolumn{2}{c}{6} \\
\quad Batch size (per device) & \multicolumn{2}{c}{1} \\
\quad Gradient accumulation   & \multicolumn{2}{c}{8} \\
\quad GPUs                    & \multicolumn{2}{c}{24 ($3 \times 8$)} \\
\quad Effective batch size    & \multicolumn{2}{c}{192} \\
\quad Learning rate           & \multicolumn{2}{c}{$1 \times 10^{-5}$} \\
\quad LR scheduler            & \multicolumn{2}{c}{Cosine} \\
\quad Warmup steps            & \multicolumn{2}{c}{100} \\
\quad Weight decay            & \multicolumn{2}{c}{0.01} \\
\bottomrule
\end{tabular}
\end{table}

\subsection{Experimental Paradigm}
\label{label_experimental_para}
\textbf{Video Processing.} Overlays fabricated text $\mathcal{T}^{-}$ constructed to satisfy three properties:
(i) positioned so as not to occlude visual regions relevant to answering $Q$;
(ii) contextually related to the video content, sharing semantic overlap with the
visual scene; and (iii) either affirming the distractor option $A^{\dagger}$, contradicting the ground-truth option $A^{*}$, or both. We incorporate $\mathcal{V}^{+}$ samples for the following reasons: (i) to prevent the model from over-correcting and discarding useful overlay text signals, thereby preserving general VQA capability; and (ii) to improve data efficiency by balancing the training distribution across congruent and contradictory text conditions. Not every video instance is associated with all three conditions; the composition is determined by the trade-off between hallucination suppression and task
performance retention.
For each sample under each condition, we record three metrics: \textbf{accuracy} ($\mathcal{R}_{\text{correct}}$), the proportion of correct predictions; \textbf{text-induced error rate} ($\mathcal{R}_{\text{text}}$), the proportion of incorrect predictions that align with the overlay text; and \textbf{residual error rate} ($\mathcal{R}_{\text{resid}}$), the proportion of incorrect predictions that align with neither the ground truth nor the overlay text. 

\subsection{Training Implementation Details}\label{sec:training}
%%% ─── Table 1: MoE Training Configuration ───────────────────────────────────

\subsection{Evaluation Results}\label{sec:eval_results}

\begin{table*}[t]
\centering
\fontsize{9.2pt}{11.5pt}\selectfont 
\setlength{\tabcolsep}{4pt} 
\setlength{\aboverulesep}{0pt}
\setlength{\belowrulesep}{0pt}
\renewcommand{\arraystretch}{1.1}
\caption{Performance comparison across three video understanding benchmarks organized by capability dimensions. Scores represent performance within each dimension: \textbf{Temporal} (temporal reasoning and sequencing), \textbf{Action} (action recognition and classification), \textbf{Object} (object detection and tracking), \textbf{Spatial} (spatial relationship understanding). Best scores in each benchmark are shown in \textbf{bold}. Subscripts show the gap to the best overall score in each benchmark.}
\label{label_dimension_performance}
\begin{tabular}{cl|cccc|c}
\toprule[2.0pt]
& \textbf{Model} & \textbf{Temporal} & \textbf{Action} & \textbf{Object} & \textbf{Spatial} & \textbf{Overall}${\scriptstyle\uparrow}$ \\
\midrule
% ---- LLaVA-Video Block ----
\cellcolor{white}\multirow{9}{*}{\rotatebox[origin=c]{90}{\textbf{LLaVA-Video}}} 
& Gemini-3.1-Pro & 67.2 & \textbf{77.0} & 69.6 & 76.2 & 73.7\(_{\scriptsize \color{skyblue}\downarrow 4.0}\) \\
\rowcolor{lightgray}
\cellcolor{white} & \gc Qwen3-VL-30B & \gc 44.9 & \gc 48.9 & \gc 60.0 & \gc 50.7 & \gc 50.4\(_{\scriptsize \color{skyblue}\downarrow 27.3}\) \\
\cellcolor{white} & Qwen3-VL-235B & 48.2 & 49.5 & 63.4 & 54.7 & 55.3\(_{\scriptsize \color{skyblue}\downarrow 22.4}\) \\
\rowcolor{lightgray}
\cellcolor{white} & \gc Internvl3.5-VL-241B & \gc 47.1 & \gc 48.4 & \gc 57.6 & \gc 51.2 & \gc 52.8\(_{\scriptsize \color{skyblue}\downarrow 24.9}\) \\
\cellcolor{white} & Baseline & 18.8 & 26.7 & 46.1 & 30.8 & 29.3\(_{\scriptsize \color{skyblue}\downarrow 48.4}\) \\
\rowcolor{lightgray}
\cellcolor{white} & \gc Baseline+CoT & \gc 39.5 & \gc 49.2 & \gc 64.3 & \gc 52.8 & \gc 50.7\(_{\scriptsize \color{skyblue}\downarrow 27.0}\) \\
\cellcolor{white} & Baseline+SFT & 41.3 & 51.5 & 60.8 & 50.9 & 51.4\(_{\scriptsize \color{skyblue}\downarrow 26.3}\) \\
\rowcolor{lightgray}
\cellcolor{white} & \gc Baseline+CoT+SFT & \gc 60.2 & \gc 68.8 & \gc 78.3 & \gc 73.4 & \gc 69.7\(_{\scriptsize \color{skyblue}\downarrow 8.0}\) \\
\rowcolor{green!15}
\cellcolor{white} & \textbf{VTHM-MoE} & 69.8 & 83.1 & 71.5 & 78.3 & \textbf{77.7} \\
\midrule
% ---- VideoMME Block ----
\cellcolor{white}\multirow{9}{*}{\rotatebox[origin=c]{90}{\textbf{VideoMME}}} 
& Gemini-3.1-Pro & 60.8 & 61.9 & 60.0 & 51.3 & 59.8\(_{\scriptsize \color{skyblue}\downarrow 1.6}\) \\
\rowcolor{lightgray}
\cellcolor{white} & \gc Qwen3-VL-30B & \gc 32.2 & \gc 56.4 & \gc 53.2 & \gc 45.2 & \gc 50.4\(_{\scriptsize \color{skyblue}\downarrow 11.0}\) \\
\cellcolor{white} & Qwen3-VL-235B & 36.7 & 59.6 & 55.8 & 37.5 & 53.7\(_{\scriptsize \color{skyblue}\downarrow 7.7}\) \\
\rowcolor{lightgray}
\cellcolor{white} & \gc Internvl3.5-VL-241B & \gc 34.2 & \gc 57.1 & \gc 52.1 & \gc 37.3 & \gc 51.5\(_{\scriptsize \color{skyblue}\downarrow 9.9}\) \\
\cellcolor{white} & Baseline & 23.4 & 30.5 & 37.5 & 31.2 & 31.0\(_{\scriptsize \color{skyblue}\downarrow 30.4}\) \\
\rowcolor{lightgray}
\cellcolor{white} & \gc Baseline+CoT & \gc 32.7 & \gc 47.0 & \gc 48.4 & \gc 44.7 & \gc 44.1\(_{\scriptsize \color{skyblue}\downarrow 17.3}\) \\
\cellcolor{white} & Baseline+SFT & -33.8 & 46.5 & 52.8 & 36.3 & 51.6\(_{\scriptsize \color{skyblue}\downarrow 9.8}\) \\
\rowcolor{lightgray}
\cellcolor{white} & \gc Baseline+CoT+SFT & \gc 35.8 & \gc 56.4 & \gc 63.0 & \gc 57.0 & \gc 53.9\(_{\scriptsize \color{skyblue}\downarrow 7.5}\) \\
\rowcolor{green!15}
\cellcolor{white} & \textbf{VTHM-MoE} & 52.3 & 61.1 & 61.5 & 65.0 & \textbf{61.4} \\
\midrule
% ---- TemporalBench Block ----
\cellcolor{white}\multirow{9}{*}{\rotatebox[origin=c]{90}{\textbf{TempBench}}} 
& Gemini-3.1-Pro & 70.2 & 69.0 & 71.3 & 61.0 & 62.5\(_{\scriptsize \color{orange}\uparrow 9.4}\) \\
\rowcolor{lightgray}
\cellcolor{white} & \gc Qwen3-VL-30B & \gc 38.5 & \gc 36.7 & \gc 44.7 & \gc 31.5 & \gc 38.3\(_{\scriptsize \color{skyblue}\downarrow 14.8}\) \\
\cellcolor{white} & Qwen3-VL-235B & 39.7 & 40.8 & 46.6 & 35.3 & 41.6\(_{\scriptsize \color{skyblue}\downarrow 11.5}\) \\
\rowcolor{lightgray}
\cellcolor{white} & \gc Internvl3.5-VL-241B & \gc 37.3 & \gc 38.1 & \gc 45.6 & \gc 34.2 & \gc 40.5\(_{\scriptsize \color{skyblue}\downarrow 12.6}\) \\
\cellcolor{white} & Baseline & 23.9 & 23.6 & 28.5 & 22.4 & 24.2\(_{\scriptsize \color{skyblue}\downarrow 28.9}\) \\
\rowcolor{lightgray}
\cellcolor{white} & \gc Baseline+CoT & \gc 36.0 & \gc 32.2 & \gc 37.1 & \gc 28.6 & \gc 32.7\(_{\scriptsize \color{skyblue}\downarrow 20.4}\) \\
\cellcolor{white} & Baseline+SFT & 35.1 & 33.5 & 30.1 & 28.5 & 34.8\(_{\scriptsize \color{skyblue}\downarrow 18.3}\) \\
\rowcolor{lightgray}
\cellcolor{white} & \gc Baseline+CoT+SFT & \gc 42.7 & \gc 40.0 & \gc 44.4 & \gc 33.7 & \gc 39.6\(_{\scriptsize \color{skyblue}\downarrow 13.5}\) \\
\rowcolor{green!15}
\cellcolor{white} & \textbf{VTHM-MoE} & 50.7 & 54.5 & 52.3 & 52.7 & \textbf{53.1} \\
\bottomrule[2.0pt]
\end{tabular}
\end{table*}
As shown in Figure~\ref{label_dimension_performance}, hallucination robustness varies notably across semantic
dimensions.  Two key observations emerge:

\textbf{Temporal hallucinations are the hardest to resist.}  Temporal reasoning inherently requires integrating visual cues distributed across long frame sequences, making the task itself more demanding than other dimensions.  When a misleading textual description provides a plausible shortcut to the answer, the model is further incentivized to forgo this costly cross-frame reasoning and instead rely on the textual cue.  These two factors---the intrinsic difficulty of long-range temporal grounding and the
model's tendency to exploit textual shortcuts in lieu of cross-frame understanding---compound each other, making temporal hallucinations particularly difficult to counteract.

\textbf{Action hallucinations are the easiest to resist.}  Unlike temporal reasoning, action recognition
typically involves a short temporal span with only localized visual changes; for instance, in a basketball shooting scene, the salient motion is largely confined to the ball while the background remains static.  This compact visual change gradient allows the model to ground its prediction in the actual video content with relative ease, reducing its reliance on potentially hallucinated textual descriptions.

\textbf{Dimension Conclusion.} Taken together, these two observations suggest a general
principle: hallucination resistance is inversely correlated
with both the temporal span required for reasoning and the
magnitude of the visual change gradient across frames.

As illustrated in Table 1, the correlation between hallucination resistance rate (HRR) and cognitive complexity reveals a clear dichotomy between advanced models and the remaining open-source models.

\textbf{Temporal and spatial cognitive complexity.} For the group of open-source models, HRR exhibits a
negative correlation with both temporal cognitive complexity (TLSR) and spatial cognitive complexity (SRLSR), indicating that their hallucination resistance degrades as the cognitive demand of the task increases.  In contrast, Gemini-3.1-Pro and VTHM-MoE display a positive correlation with these two metrics, maintaining or even improving their robustness under higher complexity.  We attribute this
divergence to different underlying strengths: Gemini-3.1-Pro benefits from its inherently strong visual
grounding capability, while VTHM-MoE owes its resilience to the dedicated TOIH-resistance mechanism introduced by its mixture-of-experts architecture.  Notably, the fact that VTHM-MoE matches the trend of a state-of-the-art closed-source model suggests that, in terms of resisting text-oriented image hallucinations, VTHM-MoE has achieved a level of robustness comparable to leading proprietary systems.

\textbf{Action and attribute cognitive complexity.} A different pattern emerges for action-level (AALSR) and
attribute-level (ASLSR) cognitive complexity: these two metrics are positively correlated with HRR for open-source models but show negligible or even negative correlation for Gemini-3.1-Pro and VTHM-MoE.  We interpret this as follows. Open-source models struggle more with perceptual-level
tasks---where action and object recognition rely heavily on fine-grained visual discrimination---making hallucinations in these dimensions harder to resist at lower complexity levels.  As cognitive requirements shift toward reasoning, however, the explicit reasoning process activates deeper deliberation within the model, partially compensating for its weaker perceptual grounding and leading to moderately
higher HRR on a subset of reasoning-intensive samples.  For Gemini-3.1-Pro and VTHM-MoE, since both models already achieve consistently high HRR, their performance remains stable across all cognitive complexity brackets without noticeable degradation as task difficulty increases.  This further corroborates that VTHM-MoE is sufficiently robust across varying levels of cognitive complexity, on par with the strongest closed-source baseline.

\subsection{Detailed Case Study Analysis}
\label{appendix_case_dtudy}

To intuitively demonstrate the efficacy of VTHM-MoE in mitigating
Text Overlay-Induced Hallucination (TOIH), we present a qualitative
analysis across the four fundamental conflict dimensions of video
understanding, as illustrated in the 4 figures. In all
cases, the embedded \textit{Hall Text} is deliberately designed to
semantically contradict the ground-truth visual content.

\textbf{Temporal Count Conflict (Figure a).}
The video depicts a person opening a refrigerator \textit{twice},
whereas the overlaid text falsely asserts it occurs \textit{three
times}. VTHM-MoE successfully identifies this cross-modal discrepancy,
evidenced by the conflict classifier outputting a high temporal
confidence score ($\text{cls}'_{\text{temporal}} = 0.80$). The
Adaptive Token Routing Strategy consequently allocates the largest
token proportion to the Temporal Expert
($\text{expert\_ratio}'_{\text{temporal}} = 0.46$). As answering this
query additionally requires understanding of the action
(``\textit{open}'') and the object (``\textit{refrigerator}''), the
model intelligently routes auxiliary tokens to the Action ($0.22$) and
Object ($0.21$) experts. Through this collaborative routing, VTHM-MoE
bypasses the textual trap and predicts the correct label
$y' = \text{C}$.

\textbf{Specific Action Conflict (Figure b).}
The model correctly ignores the overlay text
(``\textit{climbs back up}'') and identifies the true visual action
(``\textit{lands on a large padded mat}''), driven by accurate
conflict detection ($\text{cls}'_{\text{action}} = 0.75$) and dominant
routing to the Action Expert
($\text{expert\_ratio}'_{\text{action}} = 0.43$), with supplementary
context from Object ($0.28$) and Temporal ($0.22$) experts, yielding
the correct prediction $y' = \text{A}$.

\textbf{Object Recognition Conflict (Figure c).}
The overlay text hallucinates the person's attire as a blue shirt and
jeans while using a pressure washer, directly contradicting the visual
evidence of a red shirt and green apron. VTHM-MoE precisely localizes
this object-level conflict ($\text{cls}'_{\text{object}} = 0.82$) and
activates the Object Expert with the highest routing proportion
($\text{expert\_ratio}'_{\text{object}} = 0.51$), supplemented by
Action ($0.28$) and Temporal ($0.11$) experts for holistic contextual
grounding. This targeted routing successfully suppresses the textual
interference, yielding $y' = \text{A}$.

\textbf{Absolute Position Conflict (Figure d).}
The overlay text falsely asserts that a vehicle moves towards
mountains, contradicting the visual evidence of movement towards a
sandy shore. VTHM-MoE precisely detects this spatial conflict
($\text{cls}'_{\text{spatial}} = 0.80$) and strongly activates the
Spatial Expert ($\text{expert\_ratio}'_{\text{spatial}} = 0.46$), with
supplementary routing to Object ($0.20$) and Action ($0.18$) experts
for entity and motion context, yielding the correct prediction
$y' = \text{B}$.

Collectively, these four cases empirically validate the core design
principle of VTHM-MoE: by explicitly detecting the semantic conflict
dimension and adaptively routing tokens to specialized experts, the
model consistently grounds its reasoning in genuine visual perception
rather than being misled by hallucinated overlay text.

\section{Ablation Study}
\label{label_ablation}

We conduct ablation studies along three axes: the three-token input
representation, individual architectural components, and key
hyperparameters. All variants are evaluated on the Negative Sample
partition and trained under identical conditions, with performance
measured by overall answer accuracy on LLaVA-Video, VideoMME, and
TemporalBench.

\subsection{Three-Token Input Representation}\label{sec:three_token}

The patch representation fed to the LLM comprises three streams:
the visual feature $\hat{\mathbf{f}}_{\text{vis}}$, the OCR feature
$\hat{\mathbf{f}}_{\text{ocr}}$, and the difference token
$\Delta = \hat{\mathbf{f}}_{\text{ocr}} - \hat{\mathbf{f}}_{\text{vis}}$;
each variant in Table~\ref{tab:ablation_token} removes exactly one
stream while retaining the other two, isolating the marginal
contribution of each input signal to overall accuracy across all
three benchmarks.
Removing $\Delta$ incurs the largest accuracy drop across all three
datasets, confirming that an explicit cross-modal inconsistency signal
provides information that the LLM cannot recover by implicitly
comparing $\hat{\mathbf{f}}_{\text{vis}}$ and
$\hat{\mathbf{f}}_{\text{ocr}}$ alone. The full three-token
configuration consistently achieves the best performance, attaining
77.7\%, 61.4\%, and 53.1\% on LLaVA-Video, VideoMME, and
TemporalBench, respectively, with all single-stream ablations falling
3--9 percentage points below on each benchmark.

\begin{table}[t]
\centering
\small
\caption{Ablation on the three-token input representation
($\hat{\mathbf{f}}_{\text{vis}}$, $\hat{\mathbf{f}}_{\text{ocr}}$, $\Delta$).
Overall accuracy (\%) is reported.  Best results are \textbf{bolded}.}
\label{tab:ablation_token}
\setlength{\tabcolsep}{5pt}
\begin{tabular}{@{}lccc@{}}
\toprule
\textbf{Setting}
  & \textbf{LLaVA-Vid.}
  & \textbf{VideoMME}
  & \textbf{TemporalB.} \\
\midrule
Full
  & \textbf{77.7} & \textbf{61.4} & \textbf{53.1} \\
\quad w/o $\hat{\mathbf{f}}_{\text{vis}}$
  & 73.1            & 56.3            & 47.7            \\
\quad w/o $\hat{\mathbf{f}}_{\text{ocr}}$
  & 71.5            & 54.5            & 46.3            \\
\quad w/o $\Delta$
  & 70.3            & 53.6            & 44.2            \\
\bottomrule
\end{tabular}
\end{table}

\subsection{Model Components}\label{sec:model_comp}

\paragraph{Encoder Configuration.}
Our dual-encoder design assigns distinct roles to each
encoder: the visual encoder captures the raw visual content
of video frames, while the OCR encoder recognizes overlay
text such as subtitles, captions, and on-screen annotations.
The difference token~$\Delta$ further highlights the
distributional discrepancy of the same semantic content
under the two encoders, providing an explicit signal for
inconsistency detection.
To validate this design, we consider four encoder variants:
(i)~replacing the OCR encoder with a second visual encoder
(\emph{Both Visual-Enc.}),
(ii)~replacing the visual encoder with a second OCR encoder
(\emph{Both OCR-Enc.}),
(iii)~using only the visual encoder (\emph{Single Visual-Enc.}),
and (iv)~using only the OCR encoder (\emph{Single OCR-Enc.}).
As shown in Table~\ref{label_ablation_module}, all four
variants lead to consistent performance drops, confirming
that the complementary information from the two
heterogeneous encoders and their explicit difference signal
are all indispensable.

\paragraph{Module Ablation.}
Removing the conflict classifier eliminates the model's
ability to explicitly identify the category of inconsistency
between textual and visual content.  Without this targeted
classification, the model can no longer pinpoint which
aspect of the input is hallucinated and thus fails to
counteract hallucinations when generating answers.

\paragraph{Expert Ablation.}
We ablate each of the four specialized experts---temporal,
action, object, and spatial---individually, as well as the
entire MoE layer.  Each expert is designed to handle a
specific category of hallucination: the temporal expert
addresses errors in event ordering and duration, the action
expert focuses on misrecognized activities, the object
expert corrects entity-level hallucinations, and the spatial
expert handles errors in positional and relational
reasoning.  Removing any single expert degrades performance
on the corresponding hallucination type, while removing the
full MoE layer causes the largest overall drop, confirming
that the mixture-of-experts architecture is essential for
comprehensive hallucination mitigation.

\paragraph{Loss Ablation.}
We examine the contribution of each training objective.
The classification loss~$\mathcal{L}_{\text{cls}}$ supervises the
conflict classifier; without it, the model loses its ability
to categorize inconsistencies and cannot route hallucinated
tokens to the appropriate expert.
The expert-selection supervision
loss~$\mathcal{L}_{\text{sft}}$ ensures that tokens belonging to
each hallucination category are correctly routed to the
corresponding specialized expert, enabling both targeted
hallucination correction and more accurate video
understanding for non-hallucinated context tokens.
The load-balancing loss~$\mathcal{L}_{\text{aux}}$ prevents
expert collapse during routing; without it, the router
tends to degenerate by assigning most tokens to a single
expert, which invalidates the hallucination-category
classification mechanism and under-utilizes the parameters
of the remaining experts.

\begin{table}[t]
\centering
\small
\caption{Ablation study on model components and training objectives.
Overall accuracy (\%) is reported.  Best results are \textbf{bolded}.}
\label{label_ablation_module}
\setlength{\tabcolsep}{5pt}
\begin{tabular}{@{}lccc@{}}
\toprule
\textbf{Model Variant}
  & \textbf{LLaVA-Vid.}
  & \textbf{VideoMME}
  & \textbf{TemporalB.} \\
\midrule
Full Model
  & \textbf{77.7} & \textbf{61.4} & \textbf{53.1} \\
\midrule
\multicolumn{4}{@{}l}{\emph{Encoder Configuration}} \\
\quad Both Visual-Enc.
  & 73.2            & 57.4            & 47.3            \\
\quad Both OCR-Enc.
  & 65.7            & 50.2            & 42.6            \\
\quad Single Visual-Enc.
  & 68.7            & 52.6            & 45.5            \\
\quad Single OCR-Enc.
  & 63.8            & 48.5            & 41.2            \\
\midrule
\multicolumn{4}{@{}l}{\emph{Module Ablation}} \\
\quad w/o Conflict Classifier
  & 72.6            & 55.3            & 47.1            \\
\midrule
\multicolumn{4}{@{}l}{\emph{Expert Ablation}} \\
\quad w/o Temporal Expert
  & 62.1            & 50.3           & 38.5            \\
\quad w/o Action Expert
  & 58.5            & 43.5            & 42.9            \\
\quad w/o Object Expert
  & 66.0            & 58.07            & 46.8            \\
\quad w/o Spatial Expert
  & 64.1            & 56.70            & 44.7            \\
\quad w/o MoE Layer
  & 43.6            & 40.7            & 35.2            \\
\midrule
\multicolumn{4}{@{}l}{\emph{Loss Ablation}} \\
\quad w/o $\mathcal{L}_{\text{cls}}$
  & 73.5            & 56.6            & 47.7            \\
\quad w/o $\mathcal{L}_{\text{sft}}$
  & 75.9            & 59.3            & 51.5            \\
\quad w/o $\mathcal{L}_{\text{aux}}$
  & 76.3-            & 60.5            & 82.7           \\
\bottomrule
\end{tabular}
\end{table}

\subsection{Hyperparameter Sensitivity}\label{sec:hyperparam}

\begin{figure}[h]
  \centering
  \includegraphics[width=\linewidth]{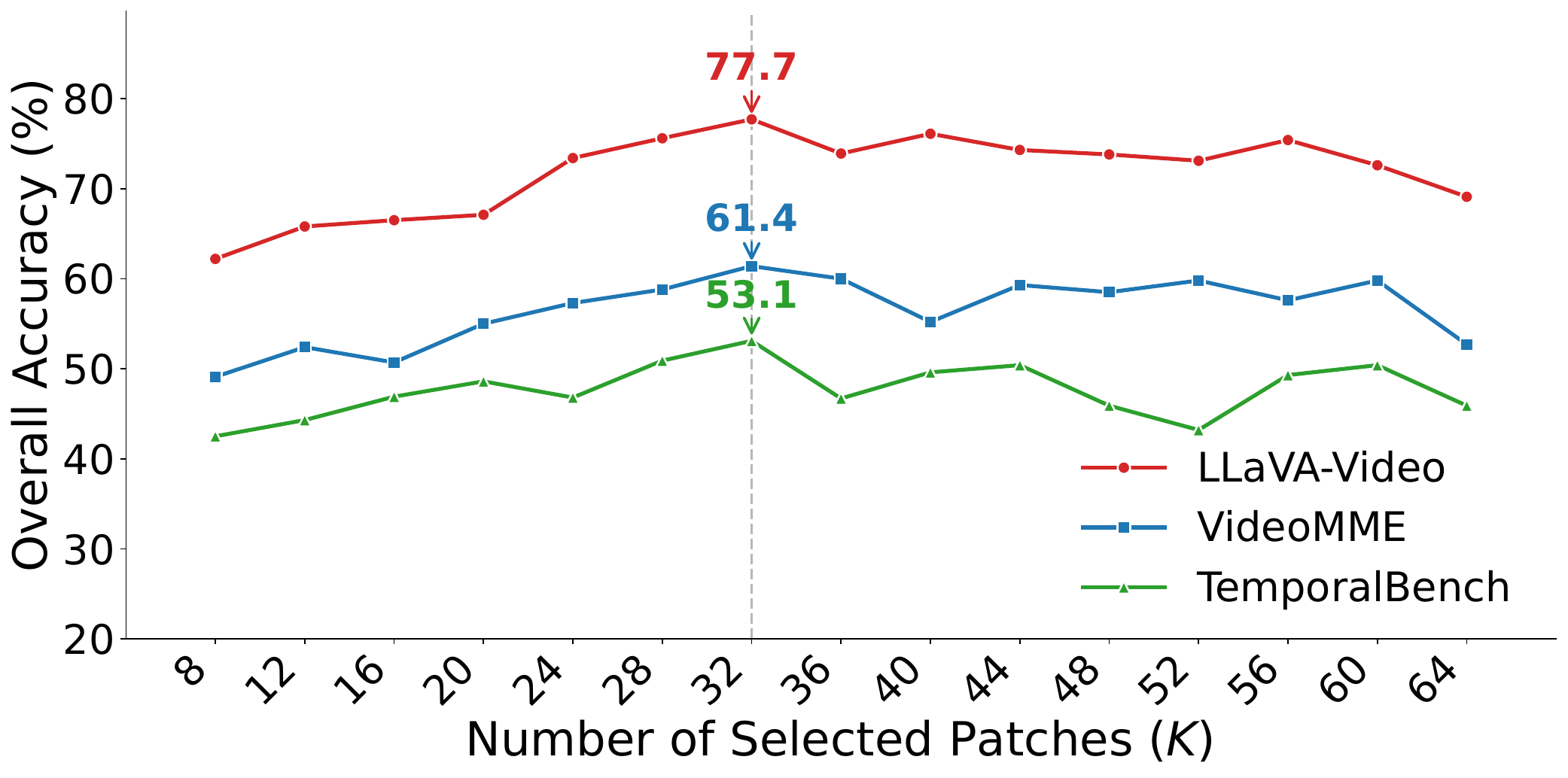}
  \caption{Ablation study on the number of selected patches $K$. We evaluate Overall accuracy on three benchmarks as $K$ varies from 8 to 64. Performance consistently peaks at $K{=}32$ across all datasets, suggesting an optimal trade-off: too few patches omit discriminative visual details, while too many introduce redundant or noisy tokens that degrade LLM reasoning.}
  \label{k_ablation}
\end{figure}

\paragraph{Number of Selected Patches $K$.}
Variants differ solely in the value of $K \in [8, 64]$,
holding all other settings fixed, so that the trend curves in
Figure~\ref{k_ablation}---plotting overall accuracy on
LLaVA-Video, VideoMME, and TemporalBench against $K$---reflect the
trade-off between patch coverage and sequence-level noise.
Performance peaks consistently at $K{=}32$ across all three
benchmarks: smaller values discard discriminative patches containing
overlay text, while larger values introduce redundant background
patches that dilute the cross-modal inconsistency signal without
further benefit. We adopt $K{=}32$ as the default.

\begin{figure}[h]
  \centering
  \includegraphics[width=\linewidth]{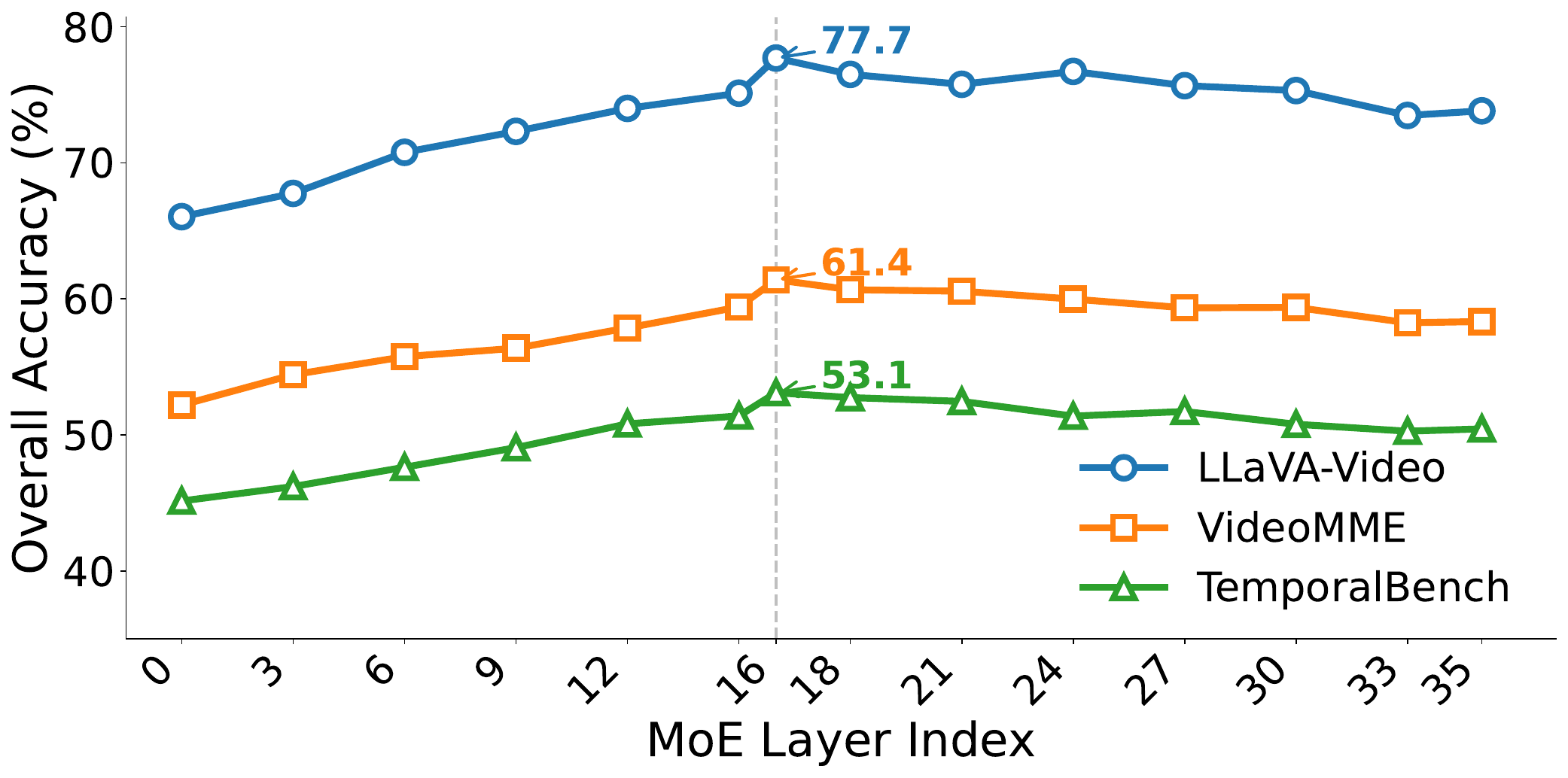}
  \caption{Overall VQA accuracy on three benchmarks (LLaVA-Video, VideoMME, and TemporalBench) as a function of the MoE insertion layer index. Performance peaks at layer 16, indicating that mid-network representations are most effective for multimodal expert routing.}
  \label{moe_layer_accuracy}
\end{figure}

\paragraph{MoE Insertion Layer.}
Variants are identical except for the transformer layer $\ell$ at
which the MoE module is inserted, covering
$\ell \in [0, 35]$, so that the resulting curves
in Figure~\ref{moe_layer_accuracy} directly reflect how the semantic
quality of hidden states available to the router at each depth
translates to overall accuracy on LLaVA-Video, VideoMME, and
TemporalBench.
Transformer layers at different depths encode qualitatively distinct
information: shallow layers capture low-level texture, while
intermediate layers encode structured semantic concepts including
temporal dynamics, actions, and spatial configurations. Layer 16
maximizes overall accuracy across all three benchmarks; shallow
insertion disproportionately degrades Temporal and Action routing,
as these concepts require multi-layer integration to emerge as
separable representations.

\paragraph{Loss Coefficients.}
\begin{figure*}[t]
    \centering
    \begin{subfigure}[b]{0.32\textwidth}
        \centering
        \includegraphics[width=\textwidth]{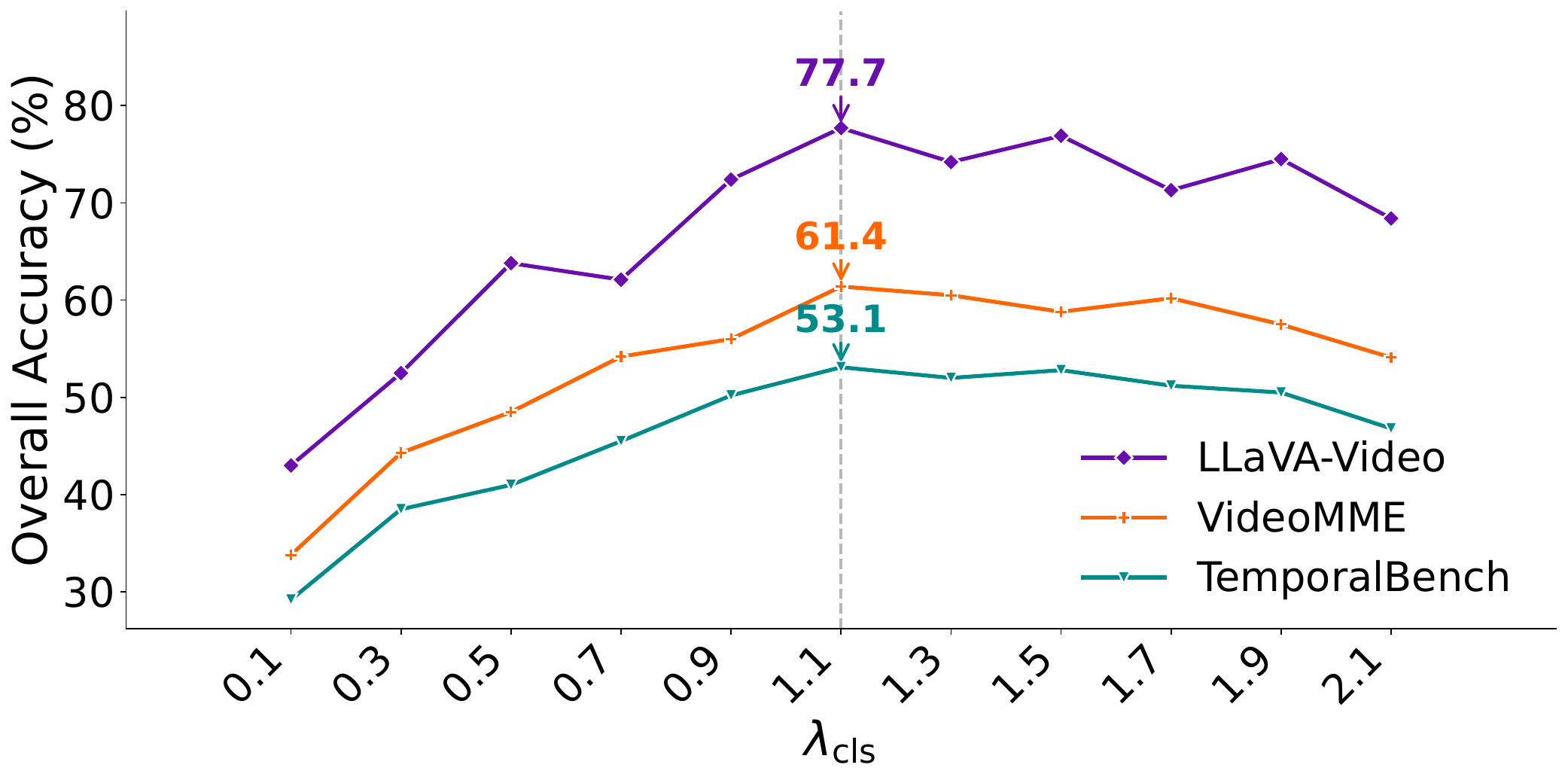}
        \caption{Effect of $\lambda_{\mathrm{cls}}$.}
        \label{fig:ablation_cls}
    \end{subfigure}
    \hfill
    \begin{subfigure}[b]{0.32\textwidth}
        \centering
        \includegraphics[width=\textwidth]{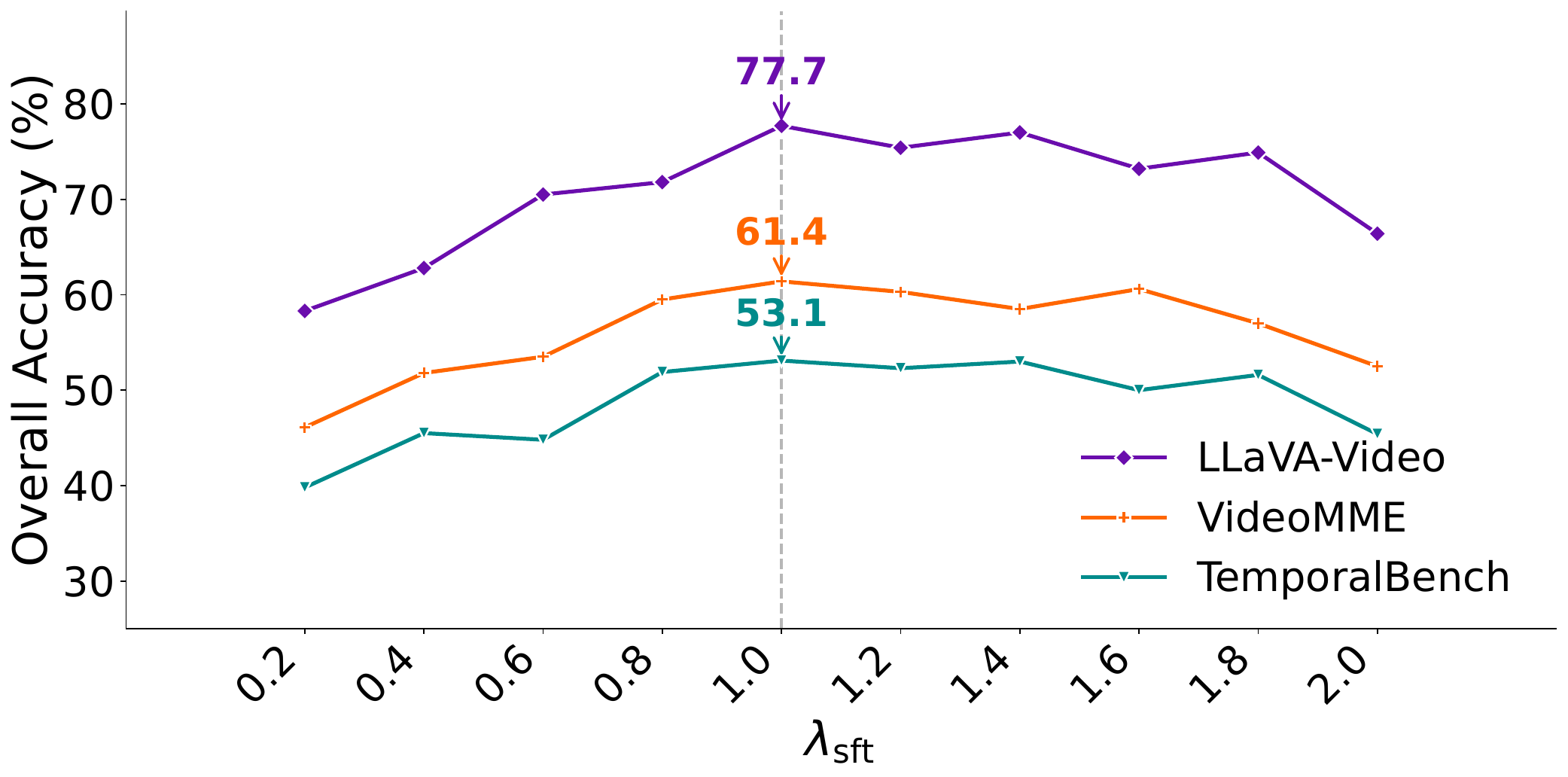}
        \caption{Effect of $\lambda_{\mathrm{sft}}$.}
        \label{fig:ablation_sft}
    \end{subfigure}
    \hfill
    \begin{subfigure}[b]{0.32\textwidth}
        \centering
        \includegraphics[width=\textwidth]{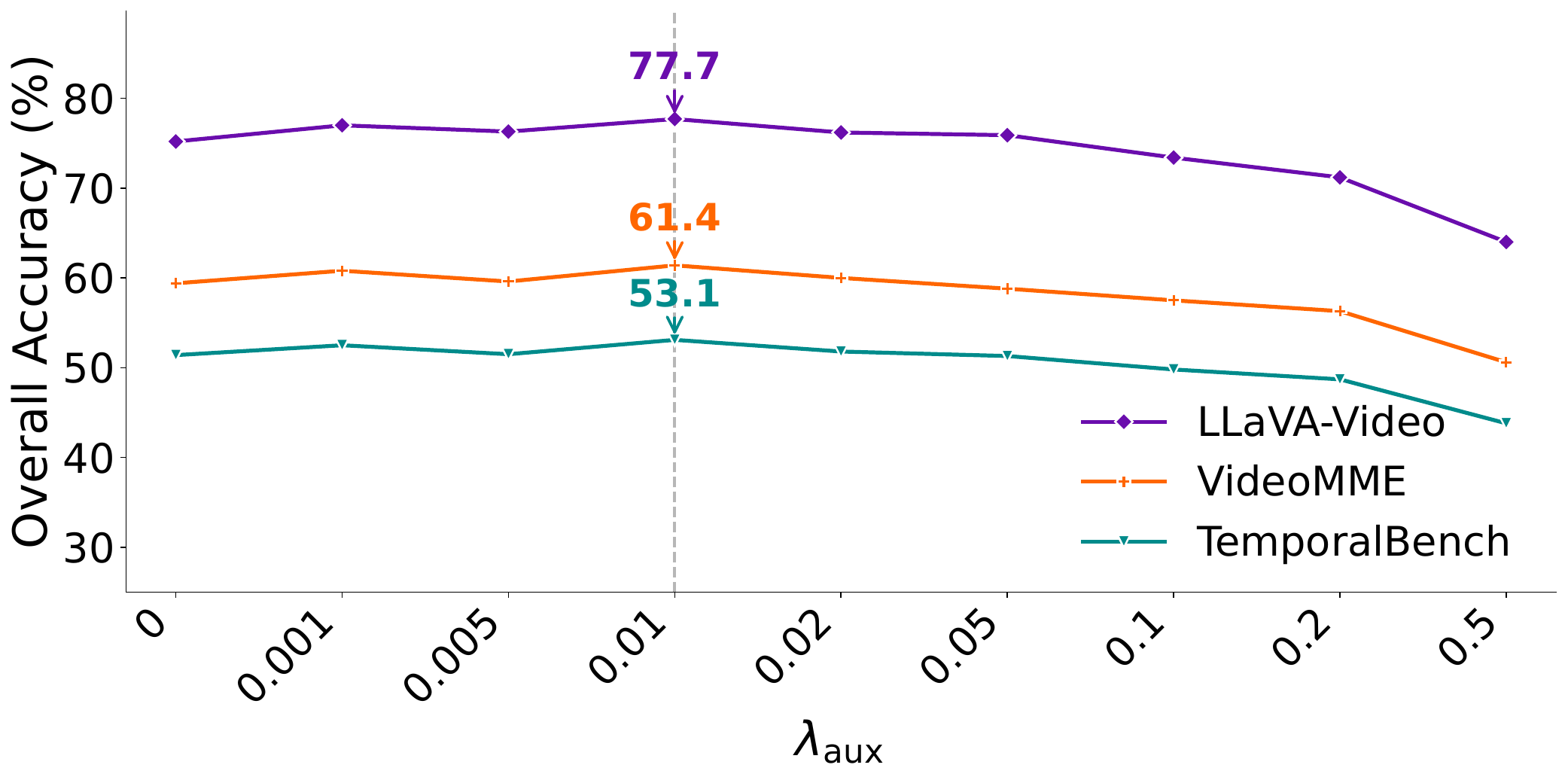}
        \caption{Effect of $\lambda_{\mathrm{aux}}$.}
        \label{fig:ablation_aux}
    \end{subfigure}
    \caption{
        Ablation study on loss coefficients.
        (a)~$\lambda_{\mathrm{cls}}$: performance peaks at $1.1$ with a steep decline for smaller values, indicating the classification loss is critical for learning discriminative temporal representations.
        (b)~$\lambda_{\mathrm{sft}}$: optimal at $1.0$ with moderate and roughly symmetric sensitivity on both sides, reflecting the need to carefully balance the supervised fine-tuning objective.
        (c)~$\lambda_{\mathrm{aux}}$: optimal at $0.01$; setting it to zero causes only a marginal accuracy drop, while large values ($\geq 0.1$) progressively degrade performance as the auxiliary objective dominates the primary training signal.
        All results are reported on LLaVA-Video, VideoMME, and TemporalBench.
    }
    \label{ablation_range_loss_hypers}
\end{figure*}

Each curve in Figure~\ref{ablation_range_loss_hypers} varies one coefficient
among $\lambda_{\text{cls}}$, $\lambda_{\text{sft}}$, and
$\lambda_{\text{aux}}$ while fixing the remaining two at their
defaults ($1.1$, $1.0$, $0.01$), with the $y$-axis reporting overall
accuracy on LLaVA-Video, VideoMME, and TemporalBench, isolating the
sensitivity of training dynamics to each supervision signal.
Downweighting $\lambda_{\text{cls}}$ produces the sharpest accuracy
degradation across all three benchmarks, as weakened classifier
supervision directly impairs dimension-accurate routing. Reducing
$\lambda_{\text{sft}}$ induces routing collapse on low-frequency
conflict categories, while setting $\lambda_{\text{aux}}{=}0$ permits
expert load imbalance without appreciable accuracy loss yet
substantially reduces routing interpretability. The adopted
configuration forms a robust optimum across all three coefficients.

\section{Prompts}\label{sec:prompts}

% ===== Action 类别 =====
\definecolor{actionhallucibg}{HTML}{D6E6FF}       % 浅蓝
\definecolor{actionhalluciframe}{HTML}{4A7EC2}
\definecolor{actioncorrectbg}{HTML}{D7F9F8}        % 淡青
\definecolor{actioncorrectframe}{HTML}{2A9D8F}

% ===== Classification（通用）=====
\definecolor{classifybg}{HTML}{E5D4EF}             % 淡紫
\definecolor{classifyframe}{HTML}{7B4FA2}

% ===== Temporal 类别 =====
\definecolor{temporalhallucibg}{HTML}{FFFFEA}      % 淡黄
\definecolor{temporalhalluciframe}{HTML}{B8A000}
\definecolor{temporalcorrectbg}{HTML}{FFF0D4}      % 暖杏
\definecolor{temporalcorrectframe}{HTML}{C47F17}

% ===== Spatial 类别 =====
\definecolor{spatialhallucibg}{HTML}{FBE0E0}       % 淡粉
\definecolor{spatialhalluciframe}{HTML}{C0504D}
\definecolor{spatialcorrectbg}{HTML}{F2EAF7}       % 浅紫
\definecolor{spatialcorrectframe}{HTML}{8B6AAE}

% ===== Object 类别 =====
\definecolor{objecthallucibg}{HTML}{F9F5FB}        % 极浅紫
\definecolor{objecthalluciframe}{HTML}{9B7DB8}
\definecolor{objectcorrectbg}{HTML}{E0F0E3}        % 浅绿（新增）
\definecolor{objectcorrectframe}{HTML}{4A8C5C}     % 深绿（新增）

% ===== Semantic Conflict Scoring =====
\definecolor{conflictscorebg}{HTML}{FFF0D4}
\definecolor{conflictscoreframe}{HTML}{C47F17}
% ===== Information Ratio =====
\definecolor{inforatiobg}{HTML}{E8F4FD}
\definecolor{inforatioframe}{HTML}{2B7A9E}

% generate video dimension

\begin{figure*}[t]
\centering
\begin{tcolorbox}[
    colback=classifybg,
    colframe=classifyframe,
    coltitle=white,
    title={\textbf{Prompt for Video Question Dimension Classification}},
    fonttitle=\small,
    fontupper=\small,
    width=\textwidth,
    boxrule=0.8pt,
    arc=3pt,
]
Given a video question and its candidate options, determine which dimension the question primarily evaluates. Each question must be assigned to \textbf{exactly one} category. Follow the priority-based decision process below:

\medskip
\noindent\textbf{Priority 1 --- Temporal:} The question focuses on the chronological order of events, temporal relationships, dynamic transitions, duration, or frequency of occurrences.

\noindent\textbf{Priority 2 --- Spatial:} The question focuses on spatial positions, relative orientations (e.g., left/right, above/below), distances between objects, or scene layouts.

\noindent\textbf{Priority 3 --- Action:} The question focuses on ongoing actions, behaviors, activities, or the manner and posture of movements.

\noindent\textbf{Priority 4 --- Object:} The question focuses on identifying specific entities such as objects, persons, animals, tools, colors, or visual attributes.

\medskip
\noindent\textbf{Decision Rules:}\\
1.~First check whether the question primarily tests \textit{Temporal}; if yes, output \texttt{temporal=YES} and set all others to \texttt{NO}.\\
2.~If not, check \textit{Spatial}; if yes, output \texttt{spatial=YES} and set all others to \texttt{NO}.\\
3.~If not, check \textit{Action}; if yes, output \texttt{action=YES} and set all others to \texttt{NO}.\\
4.~If not, check \textit{Object}; if yes, output \texttt{object=YES} and set all others to \texttt{NO}.\\
5.~If none applies, output all \texttt{NO}.

\medskip
\noindent\textbf{Question:} \{problem\}\\
\textbf{Options:} \{options\}

\medskip
\noindent\textbf{Output Format:}\\
\texttt{<temporal>YES/NO</temporal>}\\
\texttt{<spatial>YES/NO</spatial>}\\
\texttt{<action>YES/NO</action>}\\
\texttt{<object>YES/NO</object>}
\end{tcolorbox}
\caption{
    Prompt template used to classify each video question into one of four fine-grained evaluation dimensions via Gemini3.1-Pro. The priority-based design ensures that each question is assigned to exactly one category.
}
\label{label_classification_prompt}
\end{figure*}

% ============================================================
% Figure 1: Action Hallucination Generation Prompt
% ============================================================

\begin{figure*}[t]
\centering
\begin{tcolorbox}[
    colback=actionhallucibg,
    colframe=actionhalluciframe,
    coltitle=white,
    title={\textbf{Prompt for Action Hallucination Generation (Incorrect Options)}},
    fonttitle=\small,
    fontupper=\small,
    width=\textwidth,
    boxrule=0.8pt,
    arc=3pt,
]
You will see a multiple-choice question about video content, including the question and the correct option.

\medskip
\noindent\textbf{Question:} \{question\}\\
\textbf{Correct Answer:} \{correct\_option\} \{correct\_description\}\\
\textbf{Incorrect Option:} \{incorrect\_option\} \{incorrect\_description\}

\medskip
Based on the question context and the incorrect option, generate a \textbf{concise} hallucinated description that focuses on \textbf{action-related} errors, NOT temporal/sequential errors.

\medskip
\noindent\textbf{Important Requirements:}

\smallskip
\noindent 1.~\textbf{Hallucination must be action-level errors}, not detail differences:
\begin{itemize}[nosep,leftmargin=1.5em]
    \item[\checkmark] Change the action itself (e.g., ``throwing'' $\rightarrow$ ``catching'')
    \item[\checkmark] Change the action target (e.g., ``opening a door'' $\rightarrow$ ``closing a door'')
    \item[\checkmark] Change the action manner (e.g., ``gently placing'' $\rightarrow$ ``roughly throwing'')
    \item[\checkmark] Add/remove actions (e.g., add ``then jumps'' when it doesn't happen)
    \item[$\times$] Changing object colors, shapes, or sizes
    \item[$\times$] Changing temporal order (this is for temporal tasks, not action tasks)
\end{itemize}

\noindent 2.~\textbf{The hallucinated description must be based on the incorrect option}: Include the action from the incorrect option. Keep the description concise and focused on the action only.

\smallskip
\noindent 3.~\textbf{Hallucination construction strategies} (while keeping the incorrect option's action):
\begin{itemize}[nosep,leftmargin=1.5em]
    \item \textit{Keep the action, change the target}: Apply the action to the wrong object
    \item \textit{Keep the action, add details}: Keep the action but add wrong details
    \item \textit{Keep the action, change context}: Keep the action but in the wrong context
\end{itemize}

\noindent 4.~\textbf{Keep it concise and focused}: Generate short descriptions (typically 3--8 words). Focus only on the action.

\medskip
\noindent\textbf{Output Format:}\\
\texttt{<hallucination\_text>[hallucinated description]</hallucination\_text>}

\medskip
\noindent\textbf{Example:}\\
\textit{Question:} What did the person do to the blanket after opening the door?\\
\textit{Correct Answer:} D. Threw. \quad \textit{Incorrect Option:} C. Took.\\
\textit{Output:} \texttt{<hallucination\_text>Took the blanket from the couch.</hallucination\_text>}
\end{tcolorbox}
\caption{
    Prompt template used to generate action-level hallucinated descriptions for \textbf{incorrect} options via Claude-Sonnet4.6. Given a video question, the correct answer, and an incorrect option, the model produces a concise hallucinated description that retains the action verb from the incorrect option while introducing plausible but factually wrong action-level content.
}
\label{label_action_hallucination_prompt}
\end{figure*}

% ============================================================
% Figure 2: Action Correct Answer Description Prompt
% ============================================================
\begin{figure*}[t]
\centering
\begin{tcolorbox}[
    colback=actioncorrectbg,
    colframe=actioncorrectframe,
    coltitle=white,
    title={\textbf{Prompt for Action Correct Answer Description Generation}},
    fonttitle=\small,
    fontupper=\small,
    width=\textwidth,
    boxrule=0.8pt,
    arc=3pt,
]
You will see a multiple-choice question about video content and the correct answer option.

\medskip
\noindent\textbf{Question:} \{question\}\\
\textbf{Correct Answer:} \{correct\_option\} \{correct\_description\}

\medskip
Based \textbf{strictly} on the question context and the correct answer option, generate a \textbf{concise} truthful description that \textbf{combines the question context with the correct option}.

\medskip
\noindent\textbf{Critical Requirements:}

\smallskip
\noindent 1.~\textbf{Must combine question context with correct option:} Extract all relevant context from the question (objects, locations, timing, relationships, etc.) and combine this context with the action from the correct option.

\smallskip
\noindent 2.~\textbf{No additional inference or imagination:} Only use information explicitly present in the question text or the correct option description. Do not add any details, adjectives, or context that are not in the question or correct option.

\smallskip
\noindent 3.~\textbf{Construction strategy:} Identify key elements from the question (objects, subjects, locations, timing, relationships). Take the action/description from the correct option. Combine them into a concise description (typically 5--12 words).

\medskip
\noindent\textbf{Output Format:}\\
\texttt{<hallucination\_text>[truthful description]</hallucination\_text>}

\medskip
\noindent\textbf{Examples:}

\smallskip
\noindent\textit{Question:} What did the person do to the blanket after opening the door?\\
\textit{Correct Answer:} D. Threw.\\
\textit{Output:} \texttt{<hallucination\_text>Threw the blanket after opening the door.</hallucination\_text>}

\smallskip
\noindent\textit{Question:} What did the person do in the kitchen?\\
\textit{Correct Answer:} C. Cut vegetables.\\
\textit{Output:} \texttt{<hallucination\_text>Cut vegetables in the kitchen.</hallucination\_text>}

\smallskip
\noindent\textit{Question:} What specific action do the children perform while following the instructor's movements?\\
\textit{Correct Answer:} B. Raising their arms and moving rhythmically.\\
\textit{Output:} \texttt{<hallucination\_text>Raising their arms and moving rhythmically while following the instructor's movements.</hallucination\_text>}
\end{tcolorbox}
\caption{
    Prompt template used to generate truthful descriptions for \textbf{correct} options via Claude-Sonnet4.6. The model combines contextual information from the question (e.g., objects, locations, timing) with the correct option to produce a concise and faithful description, without introducing any additional inference.
}
\label{label_action_correct_answer_prompt}
\end{figure*}

% ============================================================
% Figure: Object Hallucination Generation Prompt
% ============================================================
\begin{figure*}[t]
\centering
\begin{tcolorbox}[
    colback=objecthallucibg,
    colframe=objecthalluciframe,
    coltitle=white,
    title={\textbf{Prompt for Object Hallucination Generation (Incorrect Options)}},
    fonttitle=\small,
    fontupper=\small,
    width=\textwidth,
    boxrule=0.8pt,
    arc=3pt,
]
You will see a multiple-choice question about video content, including the question and the correct option.

\medskip
\noindent\textbf{Question:} \{question\}\\
\textbf{Correct Answer:} \{correct\_option\} \{correct\_description\}\\
\textbf{Incorrect Option:} \{incorrect\_option\} \{incorrect\_description\}

\medskip
Based on the question context and the incorrect option, generate a \textbf{concise} hallucinated description that focuses on \textbf{object-related} errors, specifically about object identity, attributes, appearance, or state.

\medskip
\noindent\textbf{Core Strategy:} Use the incorrect option's object/attribute content to replace the correct answer's object/attribute, while keeping the question's context.

\medskip
\noindent\textbf{Important Requirements:}

\smallskip
\noindent 1.~\textbf{Hallucination must be object-level errors}, not temporal or spatial differences:
\begin{itemize}[nosep,leftmargin=1.5em]
    \item[\checkmark] Wrong object identity (e.g., ``a cat'' $\rightarrow$ ``a dog'')
    \item[\checkmark] Wrong color/appearance (e.g., ``red sauce'' $\rightarrow$ ``yellow sauce'')
    \item[\checkmark] Wrong material/texture (e.g., ``wooden table'' $\rightarrow$ ``metal table'')
    \item[\checkmark] Wrong count/quantity (e.g., ``two dogs'' $\rightarrow$ ``three dogs'')
    \item[\checkmark] Wrong state/condition (e.g., ``well-cooked beans'' $\rightarrow$ ``hard and uncooked beans'')
    \item[\checkmark] Wrong text/label/brand (e.g., ``AMF logo'' $\rightarrow$ ``Brunswick logo'')
    \item[$\times$] Changing temporal order or sequence of events
    \item[$\times$] Changing spatial positions or locations
    \item[$\times$] Changing the action performed on the object
\end{itemize}

\noindent 2.~\textbf{The hallucinated description must be based on the incorrect option}: Extract the object/attribute content from the incorrect option and place it into the question's context.

\smallskip
\noindent 3.~\textbf{Hallucination construction strategies}:
\begin{itemize}[nosep,leftmargin=1.5em]
    \item \textit{Object substitution}: Replace the correct object with the incorrect option's object
    \item \textit{Attribute substitution}: Keep the object but change its attribute to the incorrect option's
    \item \textit{State substitution}: Keep the object but change its state/condition
    \item \textit{Label/text substitution}: Replace the correct text/label with the incorrect option's
\end{itemize}

\noindent 4.~\textbf{Keep it concise and focused}: Generate short descriptions (typically 5--15 words). Focus only on the object attribute/identity. Include the question's object context.

\medskip
\noindent\textbf{Output Format:}\\
\texttt{<hallucination\_text>[hallucinated description]</hallucination\_text>}

\medskip
\noindent\textbf{Example 1} (Wrong color attribute)\textbf{:}\\
\textit{Question:} What is the color of the sauce applied to the meat?\\
\textit{Correct Answer:} A. Reddish-brown. \quad \textit{Incorrect Option:} B. Yellow.\\
\textit{Output:} \texttt{<hallucination\_text>The sauce applied to the meat is yellow.</hallucination\_text>}

\smallskip
\noindent\textbf{Example 2} (Wrong object state)\textbf{:}\\
\textit{Question:} How are the beans described at the end of the video?\\
\textit{Correct Answer:} A. Soft and well-cooked. \quad \textit{Incorrect Option:} B. Hard and uncooked.\\
\textit{Output:} \texttt{<hallucination\_text>The beans appear hard and uncooked at the end of the video.</hallucination\_text>}

\smallskip
\noindent\textbf{Example 3} (Wrong brand/label)\textbf{:}\\
\textit{Question:} What logo is visible on the back wall of the bowling alley throughout the video?\\
\textit{Correct Answer:} A. AMF. \quad \textit{Incorrect Option:} B. Brunswick.\\
\textit{Output:} \texttt{<hallucination\_text>A Brunswick logo is visible on the back wall of the bowling alley.</hallucination\_text>}
\end{tcolorbox}
\caption{
    Prompt template used to generate object-level hallucinated descriptions for \textbf{incorrect} options via Claude-Sonnet4.6. Given a video question, the correct answer, and an incorrect option, the model produces a concise hallucinated description that substitutes the correct object identity, attribute, state, or label with the incorrect option's content, while preserving the question's object context.
}
\label{label_object_hallucination_prompt}
\end{figure*}

% ============================================================
% Figure: Object Correct Answer Description Prompt
% ============================================================
\begin{figure*}[t]
\centering
\begin{tcolorbox}[
    colback=objectcorrectbg,
    colframe=objectcorrectframe,
    coltitle=white,
    title={\textbf{Prompt for Object Correct Answer Description Generation}},
    fonttitle=\small,
    fontupper=\small,
    width=\textwidth,
    boxrule=0.8pt,
    arc=3pt,
]
You will see a multiple-choice question about video content and the correct answer option.

\medskip
\noindent\textbf{Question:} \{question\}\\
\textbf{Correct Answer:} \{correct\_option\} \{correct\_description\}

\medskip
Based \textbf{strictly} on the question context and the correct answer option, generate a \textbf{concise} truthful description that \textbf{combines the question's object context with the correct option's object information}.

\medskip
\noindent\textbf{Critical Requirements:}

\smallskip
\noindent 1.~\textbf{Must combine question object context with correct option:} Extract the subject/object being asked about from the question (e.g., ``the sauce'', ``the dog's collar'', ``the logo'', ``the beans'') and combine this object context with the attribute/description from the correct option. The goal is to accurately describe what the object is, looks like, or its state.

\smallskip
\noindent 2.~\textbf{No additional inference or imagination:} Only use information explicitly present in the question text or correct option description. Do not add any object details, attributes, or context not mentioned in the question or correct option. Do not infer or imagine properties that are not explicitly stated.

\smallskip
\noindent 3.~\textbf{Construction strategy:} Identify the object/subject being asked about from the question. Take the attribute/description/state from the correct option. Combine them into a concise description (typically 5--15 words). Keep natural language flow while staying faithful to the source.

\medskip
\noindent\textbf{Output Format:}\\
\texttt{<hallucination\_text>[truthful description]</hallucination\_text>}

\medskip
\noindent\textbf{Example 1} (Color attribute)\textbf{:}\\
\textit{Question:} What is the color of the sauce applied to the meat?\\
\textit{Correct Answer:} A. Reddish-brown.\\
\textit{Output:} \texttt{<hallucination\_text>The sauce applied to the meat is reddish-brown.</hallucination\_text>}

\smallskip
\noindent\textbf{Example 2} (Object state)\textbf{:}\\
\textit{Question:} How are the beans described at the end of the video?\\
\textit{Correct Answer:} A. Soft and well-cooked.\\
\textit{Output:} \texttt{<hallucination\_text>The beans are soft and well-cooked at the end of the video.</hallucination\_text>}

\smallskip
\noindent\textbf{Example 3} (Brand/label)\textbf{:}\\
\textit{Question:} What logo is visible on the back wall of the bowling alley throughout the video?\\
\textit{Correct Answer:} A. AMF.\\
\textit{Output:} \texttt{<hallucination\_text>An AMF logo is visible on the back wall of the bowling alley.</hallucination\_text>}

\smallskip
\noindent\textbf{Example 4} (Color of accessory)\textbf{:}\\
\textit{Question:} What color is the hair tie the person uses?\\
\textit{Correct Answer:} A. Black.\\
\textit{Output:} \texttt{<hallucination\_text>The person uses a black hair tie.</hallucination\_text>}
\end{tcolorbox}
\caption{
    Prompt template used to generate truthful object descriptions for \textbf{correct} options via Claude-Sonnet4.6. The model combines the object/subject context from the question (e.g., object identity, appearance, state) with the correct option's attribute to produce a concise and faithful description, without introducing any additional inference.
}
\label{label_object_correct_answer_prompt}
\end{figure*}

% ============================================================
% Figure: Spatial Hallucination Generation Prompt
% ============================================================
\begin{figure*}[t]
\centering
\begin{tcolorbox}[
    colback=spatialhallucibg,
    colframe=spatialhalluciframe,
    coltitle=white,
    title={\textbf{Prompt for Spatial Hallucination Generation (Incorrect Options)}},
    fonttitle=\small,
    fontupper=\small,
    width=\textwidth,
    boxrule=0.8pt,
    arc=3pt,
]
You will see a multiple-choice question about video content, including the question and the correct option.

\medskip
\noindent\textbf{Question:} \{question\}\\
\textbf{Correct Answer:} \{correct\_option\} \{correct\_description\}\\
\textbf{Incorrect Option:} \{incorrect\_option\} \{incorrect\_description\}

\medskip
Based on the question context and the incorrect option, generate a \textbf{concise} hallucinated description that focuses on \textbf{spatial/positional} errors, NOT action-type or temporal errors.

\medskip
\noindent\textbf{Core Strategy:} Place the incorrect option's spatial content into the spatial context asked about in the question.

\medskip
\noindent\textbf{Important Requirements:}

\smallskip
\noindent 1.~\textbf{Hallucination must be spatial-level errors}, covering position, location, direction, or spatial relationship:
\begin{itemize}[nosep,leftmargin=1.5em]
    \item[\checkmark] Wrong position (e.g., ``on the left'' $\rightarrow$ ``on the right'', ``above'' $\rightarrow$ ``below'')
    \item[\checkmark] Wrong location (e.g., ``on the ceiling'' $\rightarrow$ ``on the floor'')
    \item[\checkmark] Wrong spatial relationship (e.g., ``in front of'' $\rightarrow$ ``behind'', ``next to'' $\rightarrow$ ``on top of'')
    \item[\checkmark] Wrong region/area (e.g., ``in the background'' $\rightarrow$ ``in the foreground'')
    \item[\checkmark] Wrong distance/proximity (e.g., ``close to'' $\rightarrow$ ``far from'')
    \item[\checkmark] Wrong orientation/direction (e.g., ``facing left'' $\rightarrow$ ``facing right'')
    \item[$\times$] Changing the action type (e.g., ``walking'' $\rightarrow$ ``running'')
    \item[$\times$] Changing object colors, sizes, or appearance
    \item[$\times$] Changing temporal order or sequence of events
\end{itemize}

\noindent 2.~\textbf{The hallucinated description must be based on the incorrect option}: Extract the spatial framework from the question and place the content of the incorrect option into that spatial context.

\smallskip
\noindent 3.~\textbf{Hallucination construction strategies}:
\begin{itemize}[nosep,leftmargin=1.5em]
    \item \textit{Location substitution}: Keep the spatial query from the question, substitute with incorrect option's location
    \item \textit{Spatial relationship swap}: Use the wrong spatial relationship implied by the incorrect option
    \item \textit{Region/direction substitution}: Place the subject in the wrong region or direction
\end{itemize}

\noindent 4.~\textbf{Keep it concise and focused}: Generate short descriptions (typically 5--15 words). Focus only on the spatial aspect. Include the subject from the question if explicitly present.

\medskip
\noindent\textbf{Output Format:}\\
\texttt{<hallucination\_text>[hallucinated description]</hallucination\_text>}

\medskip
\noindent\textbf{Example 1} (Location substitution)\textbf{:}\\
\textit{Question:} Where is the neon `OPEN' sign located in the tattoo parlor?\\
\textit{Correct Answer:} B. Behind the person in the striped shirt. \quad \textit{Incorrect Option:} A. On the ceiling.\\
\textit{Output:} \texttt{<hallucination\_text>The neon `OPEN' sign is located on the ceiling of the tattoo parlor.</hallucination\_text>}

\smallskip
\noindent\textbf{Example 2} (Spatial relationship swap)\textbf{:}\\
\textit{Question:} Where does the person stand relative to the door?\\
\textit{Correct Answer:} C. Directly in front of the door. \quad \textit{Incorrect Option:} D. Behind the door.\\
\textit{Output:} \texttt{<hallucination\_text>The person stands behind the door.</hallucination\_text>}

\smallskip
\noindent\textbf{Example 3} (Direction substitution)\textbf{:}\\
\textit{Question:} On which side of the screen does the logo appear?\\
\textit{Correct Answer:} B. The left side. \quad \textit{Incorrect Option:} D. The right side.\\
\textit{Output:} \texttt{<hallucination\_text>The logo appears on the right side of the screen.</hallucination\_text>}
\end{tcolorbox}
\caption{
    Prompt template used to generate spatial-level hallucinated descriptions for \textbf{incorrect} options via Claude-Sonnet4.6. Given a video question, the correct answer, and an incorrect option, the model produces a concise hallucinated description that substitutes the correct spatial position, location, direction, or spatial relationship with the incorrect option's content, while preserving the question's spatial context and subject.
}
\label{label_spatial_hallucination_prompt}
\end{figure*}

% ============================================================
% Figure: Spatial Correct Answer Description Prompt
% ============================================================
\begin{figure*}[t]
\centering
\begin{tcolorbox}[
    colback=spatialcorrectbg,
    colframe=spatialcorrectframe,
    coltitle=white,
    title={\textbf{Prompt for Spatial Correct Answer Description Generation}},
    fonttitle=\small,
    fontupper=\small,
    width=\textwidth,
    boxrule=0.8pt,
    arc=3pt,
]
You will see a multiple-choice question about video content and the correct answer option.

\medskip
\noindent\textbf{Question:} \{question\}\\
\textbf{Correct Answer:} \{correct\_option\} \{correct\_description\}

\medskip
Based \textbf{strictly} on the question context and the correct answer option, generate a \textbf{concise} truthful description that \textbf{combines the question's spatial context with the correct option's spatial information}.

\medskip
\noindent\textbf{Critical Requirements:}

\smallskip
\noindent 1.~\textbf{Must combine question spatial context with correct option:} Extract all relevant spatial context from the question --- location keywords (where, position, location, side, area, region, direction), spatial relationships (in front of, behind, above, below, next to, left, right), and spatial references --- and combine this with the content from the correct option. The goal is to accurately describe where things are located or how they are spatially arranged.

\smallskip
\noindent 2.~\textbf{No additional inference or imagination:} Only use information explicitly present in the question text or correct option description. Do not add any spatial details, positions, or location information not mentioned in the question or correct option. Do not infer spatial relationships beyond what is explicitly stated.

\smallskip
\noindent 3.~\textbf{Construction strategy:} Identify spatial keywords from the question (where/location/position/which side/background/foreground/left/right/above/below/near/far/in front of/behind). Take the spatial content from the correct option. Combine them into a concise description (typically 5--15 words). If the correct option already contains the full spatial description, use it directly. Include the subject from the question if explicitly present.

\medskip
\noindent\textbf{Output Format:}\\
\texttt{<hallucination\_text>[truthful description]</hallucination\_text>}

\medskip
\noindent\textbf{Example 1} (Where question)\textbf{:}\\
\textit{Question:} Where is the neon `OPEN' sign located in the tattoo parlor?\\
\textit{Correct Answer:} B. Behind the person in the striped shirt.\\
\textit{Output:} \texttt{<hallucination\_text>The neon `OPEN' sign is located behind the person in the striped shirt.</hallucination\_text>}

\smallskip
\noindent\textbf{Example 2} (Position question)\textbf{:}\\
\textit{Question:} What logo is visible on the back wall of the bowling alley throughout the video?\\
\textit{Correct Answer:} B. Brunswick.\\
\textit{Output:} \texttt{<hallucination\_text>The Brunswick logo is visible on the back wall of the bowling alley.</hallucination\_text>}

\smallskip
\noindent\textbf{Example 3} (Backdrop/location question)\textbf{:}\\
\textit{Question:} What is the backdrop for the breakdancing demonstration?\\
\textit{Correct Answer:} D. A black backdrop with a `HIP HOP' banner.\\
\textit{Output:} \texttt{<hallucination\_text>The breakdancing demonstration takes place against a black backdrop with a `HIP HOP' banner.</hallucination\_text>}

\smallskip
\noindent\textbf{Example 4} (Support/location question)\textbf{:}\\
\textit{Question:} What does the person rely on for support while descending the cliffside?\\
\textit{Correct Answer:} C. A rope.\\
\textit{Output:} \texttt{<hallucination\_text>The person relies on a rope for support while descending the cliffside.</hallucination\_text>}
\end{tcolorbox}
\caption{
    Prompt template used to generate truthful spatial descriptions for \textbf{correct} options via Claude-Sonnet4.6. The model combines the spatial context from the question (e.g., location, position, direction, spatial relationship) with the correct option's content to produce a concise and faithful description, without introducing any additional spatial inference.
}
\label{label_spatial_correct_answer_prompt}
\end{figure*}

% ============================================================
% Figure: Temporal Hallucination Generation Prompt
% ============================================================
\begin{figure*}[t]
\centering
\begin{tcolorbox}[
    colback=temporalhallucibg,
    colframe=temporalhalluciframe,
    coltitle=white,
    title={\textbf{Prompt for Temporal Hallucination Generation (Incorrect Options)}},
    fonttitle=\small,
    fontupper=\small,
    width=\textwidth,
    boxrule=0.8pt,
    arc=3pt,
]
You will see a multiple-choice question about video content, including the question and the correct option.

\medskip
\noindent\textbf{Question:} \{question\}\\
\textbf{Correct Answer:} \{correct\_option\} \{correct\_description\}\\
\textbf{Incorrect Option:} \{incorrect\_option\} \{incorrect\_description\}

\medskip
Based on the question context and the incorrect option, generate a \textbf{concise} hallucinated description that focuses on \textbf{temporal/sequential} errors, NOT action-type or appearance errors.

\medskip
\noindent\textbf{Core Strategy:} Place the incorrect option's content into the temporal position/relationship asked about in the question.

\medskip
\noindent\textbf{Important Requirements:}

\smallskip
\noindent 1.~\textbf{Hallucination must be temporal-level errors}, not action differences:
\begin{itemize}[nosep,leftmargin=1.5em]
    \item[\checkmark] Change temporal positions (e.g., ``first action'' $\rightarrow$ ``second action'')
    \item[\checkmark] Swap sequence order (e.g., ``A then B'' $\rightarrow$ ``B then A'')
    \item[\checkmark] Change before/after relationships (e.g., ``after X, did Y'' $\rightarrow$ ``after X, did Z'' using incorrect option)
    \item[\checkmark] Change concurrent/simultaneous timing (e.g., ``while X, did Y'' $\rightarrow$ ``while X, did Z'')
    \item[\checkmark] Change transition points (e.g., ``when X finished, did Y'' $\rightarrow$ ``when X finished, did Z'')
    \item[$\times$] Changing the action type itself when action is not in incorrect option
    \item[$\times$] Changing object colors, shapes, or sizes
    \item[$\times$] Adding temporal steps not implied by question or options
\end{itemize}

\noindent 2.~\textbf{The hallucinated description must be based on the incorrect option}: Extract the temporal framework from the question (e.g., ``first'', ``after X'', ``second action'', ``before Y'') and place the content of the incorrect option into that temporal position.

\smallskip
\noindent 3.~\textbf{Hallucination construction strategies}:
\begin{itemize}[nosep,leftmargin=1.5em]
    \item \textit{Temporal position substitution}: Keep the temporal position from the question, substitute with incorrect option's content
    \item \textit{Sequence reversal/reorder}: Use the reordered sequence implied by the incorrect option
    \item \textit{Before/after swap}: Switch the temporal connector while keeping the context
\end{itemize}

\noindent 4.~\textbf{Keep it concise and focused}: Generate short descriptions (typically 5--15 words). Focus only on the temporal aspect. Include the subject from the question if explicitly present.

\medskip
\noindent\textbf{Output Format:}\\
\texttt{<hallucination\_text>[hallucinated description]</hallucination\_text>}

\medskip
\noindent\textbf{Example 1} (Temporal position substitution)\textbf{:}\\
\textit{Question:} What did the person do after opening the door?\\
\textit{Correct Answer:} A. Sat down on the chair. \quad \textit{Incorrect Option:} C. Picked up the phone.\\
\textit{Output:} \texttt{<hallucination\_text>The person picked up the phone after opening the door.</hallucination\_text>}

\smallskip
\noindent\textbf{Example 2} (Sequence order error)\textbf{:}\\
\textit{Question:} What was the correct order of actions in the video?\\
\textit{Correct Answer:} B. First washed hands, then dried them. \quad \textit{Incorrect Option:} D. First dried hands, then washed them.\\
\textit{Output:} \texttt{<hallucination\_text>First dried hands, then washed them.</hallucination\_text>}

\smallskip
\noindent\textbf{Example 3} (First/second position error)\textbf{:}\\
\textit{Question:} What was the first action the chef performed?\\
\textit{Correct Answer:} C. Chopped the vegetables. \quad \textit{Incorrect Option:} A. Boiled the water.\\
\textit{Output:} \texttt{<hallucination\_text>The chef boiled the water first.</hallucination\_text>}
\end{tcolorbox}
\caption{
    Prompt template used to generate temporal-level hallucinated descriptions for \textbf{incorrect} options via Claude-Sonnet4.6. Given a video question, the correct answer, and an incorrect option, the model produces a concise hallucinated description that places the incorrect option's content into the temporal position or sequential relationship asked about in the question, introducing plausible but temporally wrong content.
}
\label{label_temporal_hallucination_prompt}
\end{figure*}

% ============================================================
% Figure: Temporal Correct Answer Description Prompt
% ============================================================
\begin{figure*}[t]
\centering
\begin{tcolorbox}[
    colback=temporalcorrectbg,
    colframe=temporalcorrectframe,
    coltitle=white,
    title={\textbf{Prompt for Temporal Correct Answer Description Generation}},
    fonttitle=\small,
    fontupper=\small,
    width=\textwidth,
    boxrule=0.8pt,
    arc=3pt,
]
You will see a multiple-choice question about video content and the correct answer option.

\medskip
\noindent\textbf{Question:} \{question\}\\
\textbf{Correct Answer:} \{correct\_option\} \{correct\_description\}

\medskip
Based \textbf{strictly} on the question context and the correct answer option, generate a \textbf{concise} truthful description that \textbf{combines the question's temporal context with the correct option's temporal information}.

\medskip
\noindent\textbf{Critical Requirements:}

\smallskip
\noindent 1.~\textbf{Must combine question temporal context with correct option:} Extract all relevant temporal context from the question --- sequence keywords (first, second, last, before, after, then, finally, when, while, during, next), temporal positions, and event ordering --- and combine this temporal context with the action/event from the correct option. The goal is to accurately describe when and in what order things happened.

\smallskip
\noindent 2.~\textbf{No additional inference or imagination:} Only use information explicitly present in the question text or correct option description. Do not add any temporal details, intermediate steps, or timing information not mentioned in the question or correct option. Do not infer what happened before or after beyond what is explicitly stated.

\smallskip
\noindent 3.~\textbf{Construction strategy:} Identify temporal keywords from the question (first/second/last, before/after, then/when/while/during). Take the action/event from the correct option. Combine them into a concise description that preserves the temporal relationship (typically 5--15 words). If the correct option already contains the full temporal sequence, use it directly.

\smallskip
\noindent 4.~\textbf{Include the subject from the question if explicitly present:} If the question mentions a subject (e.g., ``the person'', ``the chef''), include it naturally before the action verb. Do not invent a subject not in the question.

\medskip
\noindent\textbf{Output Format:}\\
\texttt{<hallucination\_text>[truthful description]</hallucination\_text>}

\medskip
\noindent\textbf{Example 1} (After/before context)\textbf{:}\\
\textit{Question:} What did the person do after opening the door?\\
\textit{Correct Answer:} A. Sat down on the chair.\\
\textit{Output:} \texttt{<hallucination\_text>The person sat down on the chair after opening the door.</hallucination\_text>}

\smallskip
\noindent\textbf{Example 2} (First position)\textbf{:}\\
\textit{Question:} What was the first action the chef performed?\\
\textit{Correct Answer:} C. Chopped the vegetables.\\
\textit{Output:} \texttt{<hallucination\_text>The chef chopped the vegetables first.</hallucination\_text>}

\smallskip
\noindent\textbf{Example 3} (Before context)\textbf{:}\\
\textit{Question:} What did the person do before sitting down?\\
\textit{Correct Answer:} B. Removed their coat.\\
\textit{Output:} \texttt{<hallucination\_text>The person removed their coat before sitting down.</hallucination\_text>}

\smallskip
\noindent\textbf{Example 4} (No explicit subject)\textbf{:}\\
\textit{Question:} What happened after the alarm went off?\\
\textit{Correct Answer:} B. The blanket was folded and put away.\\
\textit{Output:} \texttt{<hallucination\_text>The blanket was folded and put away after the alarm went off.</hallucination\_text>}
\end{tcolorbox}
\caption{
    Prompt template used to generate truthful temporal descriptions for \textbf{correct} options via Claude-Sonnet4.6. The model combines temporal context from the question (e.g., sequence keywords, before/after relationships, event ordering) with the correct option's content to produce a concise and faithful description, without introducing any additional temporal inference.
}
\label{label_temporal_correct_answer_prompt}
\end{figure*}

% ============================================================
% Figure: Semantic Conflict Scoring Prompt
% ============================================================
\begin{figure*}[t]
\centering
\begin{tcolorbox}[
    colback=conflictscorebg,
    colframe=conflictscoreframe,
    coltitle=white,
    title={\textbf{Prompt for Semantic Conflict Scoring Between Correct Answer and Hallucinated Text}},
    fonttitle=\small,
    fontupper=\small,
    width=\textwidth,
    boxrule=0.8pt,
    arc=3pt,
]
You are a professional semantic analysis expert. Please analyze the degree of semantic conflict between the ``Correct Answer Content'' and the ``Hallucinated Text'' in the following video question-answer pair.

\medskip
\noindent\textbf{Scoring Rubric --- Semantic Conflict Score (SCS):}

\smallskip
\noindent\textbf{1} \textit{[Weakly Related / Irrelevant]:} The hallucinated text describes unrelated details and does not directly contradict the correct answer.

\noindent\textbf{2} \textit{[Entity Displacement]:} The hallucinated text assigns the action or state to the wrong object or entity.

\noindent\textbf{3} \textit{[Attribute / State Conflict]:} Both texts describe the same object, but conflict in attributes such as color, material, posture, or state.

\noindent\textbf{4} \textit{[Direct Semantic Opposition]:} The two descriptions are logically mutually exclusive and cannot physically co-exist (e.g., ``on the table'' vs.\ ``underground'').

\noindent\textbf{5} \textit{[Polarity Reversal / Mirror Misdirection]:} The hallucinated text is a direct antonym or mirror-direction of the correct answer (e.g., left vs.\ right, up vs.\ down, increase vs.\ decrease).

\medskip
\noindent\textbf{Input:}\\
\textbf{Question:} \{problem\}\\
\textbf{Correct Answer Content:} \{correct\_answer\_content\}\\
\textbf{Hallucinated Text:} \{hallucination\_text\}

\medskip
\noindent\textbf{Output Requirement:}\\
Output \textbf{only} a JSON object (no additional explanation text) with the following fields:

\smallskip
\begin{verbatim}
{
    "conflict_score": <integer 1-5>,
    "conflict_reason": "<brief justification>",
    "is_mutually_exclusive": <true/false>
}
\end{verbatim}
\end{tcolorbox}
\caption{
    Prompt template used to evaluate the semantic conflict between the correct answer and a hallucinated description via Claude-Sonnet4.6. The model assigns a Semantic Conflict Score (SCS) on a 1--5 scale, ranging from weakly related content (1) to direct polarity reversal (5), along with a brief justification and a binary mutual-exclusivity judgment.
}
\label{label_conflict_scoring_prompt}
\end{figure*}

% ============================================================
% Figure: Information Ratio Prompt
% ============================================================
\begin{figure*}[t]
\centering
\begin{tcolorbox}[
    colback=inforatiobg,
    colframe=inforatioframe,
    coltitle=white,
    title={\textbf{Prompt for Multi-Dimensional Information Ratio Estimation}},
    fonttitle=\small,
    fontupper=\small,
    width=\textwidth,
    boxrule=0.8pt,
    arc=3pt,
]
You will see a video-related question and an option description.

\smallskip
\noindent\textbf{Question:} \{problem\} \quad \textbf{Option Description:} \{option\_description\}

\smallskip
\noindent\textbf{Task:} Analyze what types of information from the video are needed to answer this question with this option. There are 4 types of information:

\smallskip
\noindent
\textbf{1.\ Temporal:} Information about time sequence, order of events, temporal relationships, before/after, duration, etc.\\
\textbf{2.\ Action:} Information about actions, movements, activities, what someone is doing, how actions are performed, etc.\\
\textbf{3.\ Object:} Information about objects, people, items, entities that appear in the video, their identification, etc.\\
\textbf{4.\ Spatial:} Information about spatial relationships, positions, locations, directions, layout, distance, etc.

\smallskip
\noindent Output a \(1\!\times\!4\) array where each element is a float in \([0,1]\), representing the proportion/importance of each information type: \texttt{[Temporal, Action, Object, Spatial]}.

\medskip
\noindent\textbf{Critical Rules:}

\smallskip
\noindent\textbf{Rule 1 --- Answer Category Must Have the Highest Proportion.}
First, identify the answer category from the question's semantic meaning: ``What object'' \(\!\to\!\) Object; ``What action / What did'' \(\!\to\!\) Action; ``Where'' \(\!\to\!\) Spatial; ``When / Before / After'' \(\!\to\!\) Temporal. The answer category's proportion must be the largest among all four.

\smallskip
\noindent\textbf{Rule 2 --- Other Categories Are Conditions.}
Condition categories represent prerequisites for locating the answer. Their proportions must \textbf{never} exceed the answer category's proportion. Assign \(0.0\) to unused conditions.

\smallskip
\noindent\textbf{Rule 3 --- General Constraints.}
Each value must lie in \([0.0, 1.0]\). The sum of all four values must \textbf{not} exceed \(1.0\).

\medskip
\noindent\textbf{Output Format (mandatory):}\\
\texttt{<ratios>[float1, float2, float3, float4]</ratios>}

\medskip
\noindent\textbf{Examples:}

\smallskip
\noindent\textit{Q: What did the person do to the blanket after opening the door? \; A: Threw the blanket.}\\
Answer category: Action (highest). Temporal ``after opening door'': 0.3; Action: 0.5; Object ``blanket'': 0.2; Spatial: 0.0. Sum\,=\,1.0\,\(\leq\)\,1.0\;\checkmark\\
\texttt{<ratios>[0.3, 0.5, 0.2, 0.0]</ratios>}

\smallskip
\noindent\textit{Q: Which object is on the left side of the table? \; A: A red cup.}\\
Answer category: Object (highest). Spatial ``left side'': 0.4; Object: 0.6; others: 0.0. Sum\,=\,1.0\;\checkmark\\
\texttt{<ratios>[0.0, 0.0, 0.6, 0.4]</ratios>}

\smallskip
\noindent\textit{Q: Where did the person place the book? \; A: On the shelf.}\\
Answer category: Spatial (highest). Action ``place'': 0.2; Object ``book, shelf'': 0.3; Spatial: 0.5. Sum\,=\,1.0\;\checkmark\\
\texttt{<ratios>[0.0, 0.2, 0.3, 0.5]</ratios>}

\smallskip
\noindent\textit{Q: What happened before the person sat down? \; A: Put the book on the shelf.}\\
Answer category: Action (highest). Temporal ``before sitting'': 0.35; Action: 0.4; Object: 0.15; Spatial: 0.1. Sum\,=\,1.0\;\checkmark\\
\texttt{<ratios>[0.35, 0.4, 0.15, 0.1]</ratios>}

\end{tcolorbox}
\caption{
    Prompt template for estimating the multi-dimensional information ratio needed to answer a video question with a given option. Gemini-3.1-Pro outputs a four-element vector \texttt{[Temporal, Action, Object, Spatial]} whose values reflect the relative importance of each information dimension. The answer category (determined by question semantics) is constrained to hold the highest proportion, while remaining categories serve as conditional prerequisites.
}
\label{fig:information_ratio_prompt}
\end{figure*}

\end{document}